\documentclass{article}

\PassOptionsToPackage{numbers, sort&compress} {natbib}
\usepackage[eandd, final]{neurips_2026}
\usepackage[utf8]{inputenc} % allow utf-8 input
\usepackage[T1]{fontenc}    % use 8-bit T1 fonts
\usepackage{hyperref}       % hyperlinks
\usepackage{url}            % simple URL typesetting
\usepackage{booktabs}       % professional-quality tables
\usepackage{amsfonts}       % blackboard math symbols
\usepackage{nicefrac}       % compact symbols for 1/2, etc.
\usepackage{microtype}      % microtypography
\usepackage{xcolor}         % colors
\usepackage{graphicx}
\usepackage{subcaption}
\usepackage{multirow}
\usepackage{enumitem}
\usepackage[autostyle]{csquotes}
\usepackage{wrapfig2}

\usepackage{amsmath}
\usepackage{amssymb}
\usepackage{mathtools}
\usepackage{amsthm}

\usepackage[capitalize,noabbrev]{cleveref}
\usepackage[table]{xcolor}
\usepackage{tabularx}
\usepackage{array}
\usepackage{makecell}
\newcommand{\withinstd}{\cellcolor{blue!12}}   % within 1 std (non-best)
\newcolumntype{Y}{>{\centering\arraybackslash}X}

\title{OmicsGBench: A Framework for Evaluating Graph Neural Networks on Omics Data}

\author{%
  Louisa Cornelis \\
  UC Santa Barbara\\
  \texttt{louisacornelis@ucsb.edu} \\
  \And
  Johan Mathe \\
  Atmo, Inc.\\
  \And
  Louis Van Langendonck \\
  Universitat Politècnica de Catalunya\\
  \AND
  Guillermo Bernárdez\thanks{Joint last authors.} \\
  UC Santa Barbara\\
  \And
  Nina Miolane\footnotemark[1] \\
  UC Santa Barbara\\
}

\begin{document}

\maketitle
% \footnotetext[1]{$^{\dagger}$Equal contribution.}

\begin{abstract}
Graph Neural Networks (GNNs) have become the dominant framework for inductive graph-level learning. Yet most benchmarks focus on the regime $n \gg p$, where the number of graphs $n$ greatly exceeds the number of nodes per graph $p$. 
This overlooks biological domains such as omics, which operate in the opposite $n \ll p$ regime, characterized by large graphs of genes, transcripts, or proteins across few patient samples.
This raises the question: \textit{how do GNNs perform in this low-sample, high-node omics setting?} 
We introduce \texttt{OmicsGBench} (Omics-Graph Bench), the first benchmarking platform for graph-level prediction on omics expression data, a domain whose defining characteristic is the $n \ll p$ regime. We provide a standardized, end-to-end modular infrastructure from raw omics data to families of featured graphs with varied structural properties.
We benchmark classical GNNs, as well as GNNs designed for large graphs and omics applications, alongside MLPs and machine learning baselines to establish reference performances. Our results show that widely used GNNs often do not outperform simple MLPs and classical baselines. Ablation studies show that performance relies primarily on feature selection and whether the readout preserves node identity. Furthermore, graphs built from known biological interactions offer no consistent advantage over co-expression graphs. These findings challenge the prevailing assumption that graph structure inherently adds value in this domain, fostering a critical reassessment of current learning paradigms. Ultimately, by exposing these limitations, OmicsGBench provides the open-source ecosystem necessary for the community to develop and validate novel architectures explicitly tailored for biological graphs. The code is available at \url{https://github.com/geometric-intelligence/OmicsGBench}.
\end{abstract}

\section{Introduction}

Many real-world systems—such as biological \citep{barabasi2004network} or social networks \citep{easley_kleinberg_2010_networks}—can be naturally represented as graphs, where entities serve as nodes and their interactions as edges. Graph Neural Networks (GNNs) \citep{scarselli2008graph} are designed to extract knowledge from these complex structures \citep{graphsage, kipf2017semisupervised, gilmer2017neural}. This structural approach has found application in omics~\citep{Li25}, where the challenge is to understand how genes, transcripts, and proteins interact to drive
biological function. The field is increasingly adopting graph representations because traditional \textquote{bag-of-features} analyses miss these interactions. In genomics, nodes can represent genes linked by regulatory or spatial relationships; in transcriptomics, nodes usually correspond to RNA transcripts with edges reflecting co-expression patterns; and in proteomics, the connectivity between proteins can be formulated using known or predicted protein-protein interactions (PPI). These biological graphs provide a mechanism-aware framework for using GNNs to study disease complexity.
%Many real-world systems---such as biological \citep{barabasi2004network} or social networks \citep{easley_kleinberg_2010_networks}---can be naturally represented as graphs, where entities are nodes and their interactions are edges. Graph Neural Networks (GNNs)~\citep{scarselli2008graph} are designed to process and extract knowledge from such structures \citep{10.5555/3294771.3294869, DBLP:conf/iclr/KipfW17, 10.5555/3305381.3305512}. In biomedicine, this perspective is especially useful for omics~\citep{Li25}. In genomics, nodes represent genes and edges can indicate regulatory or spatial relationships. In transcriptomics, nodes correspond to RNA transcripts and edges can reflect co-expression. In proteomics, nodes are proteins and edges can be formulated using known or predicted protein-protein interactions. These biological graphs provide a natural framework for applying GNNs to better understand underlying biology across health and disease.

However, a significant disconnect exists between standard GNN practices and the unique characteristics of omics data. On the one hand, the regime $n \ll p$, where the number of nodes per graph $p$ far exceeds the number of graphs $n$, is a defining characteristic of modern omics. Over the past two decades, advances in high-throughput measurement technologies have enabled genome-, transcriptome-, and proteome-wide profiling at an unprecedented scale \citep{soon_hariharan_snyder_2013_high_throughput_sequencing, Dai2022}. In proteomics, for example, it is now feasible to measure the expression of tens of thousands of proteins \textit{in vivo}, a dramatic expansion over the few hundred previously accessible \citep{Kuster2024}---with comparable orders of magnitude observed in genomics \citep{satam_etal_2024_erratum_ngs_trends} and transcriptomics \cite{molla_desta_birhanu_2025_advancements_scrna_spatial}. This surge in dimensionality naturally leads to graph-based representations with a large number of nodes $p$, such as networks representing gene co-expressions, protein–protein interactions, or metabolic pathways \citep{vanDam2018Gene, PPI, agamah2022_networkbased_integrative_multiomics}. Yet, while node counts have exploded, the number of labeled samples $n$ (e.g., subjects in a study) remains constrained by costs, privacy regulations, and clinical availability.

%---and this is most acute for rare diseases that remain understudied, where limited biological and clinical understanding makes research efforts essential.
%As a result, graph-based representations with a large number of nodes $p$, such as gene co-expression \citep{vandam2018_gene_coexpression}, protein–protein interaction \citep{sharan2007networkbased_protein_function}, or pathway networks, are increasingly used in downstream modeling \citep{agamah2022_networkbased_integrative_multiomics}. Yet, due to cost, privacy, and clinical constraints, the number of labeled samples $n$, which typically correspond to the number of subjects, often remains small. This is especially true for rare diseases that remain understudied, where limited biological and clinical understanding make research efforts essential. %This low-sample, large-graph ($n \ll p$) regime of omics data contrasts with the common GNN benchmarks, which operate in $n \gg p$ settings.

\begin{wrapfigure}{r}{0.5\textwidth}
    \centering
    \vspace{-15pt}
    \includegraphics[width=0.48\textwidth]{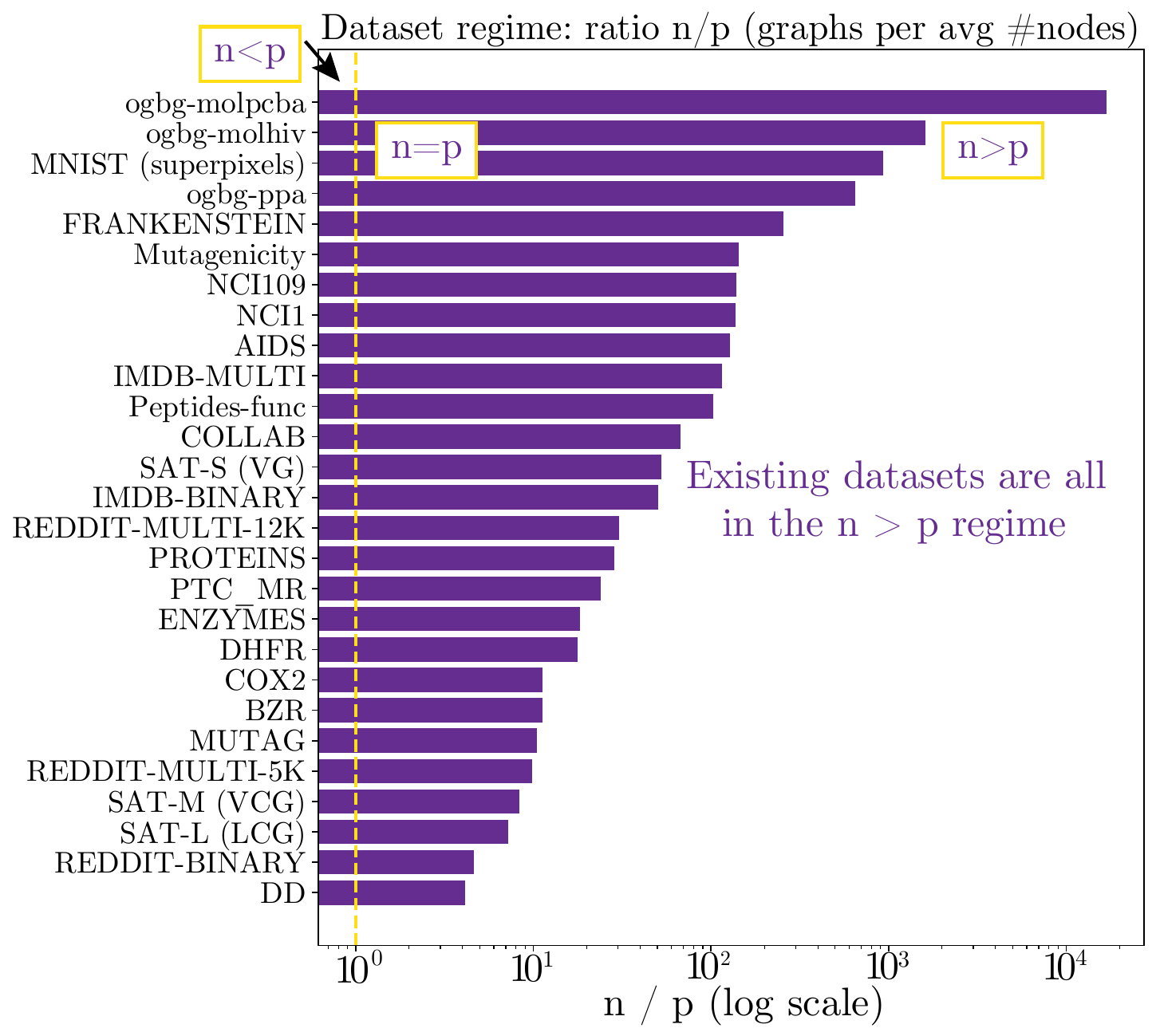}
    \caption{\textbf{Existing graph benchmarks operate in the $n \gg p$ regime}, where the number of graphs $n$ far exceeds the average number of nodes per graph $p$. Bar plot of $n / p$ for benchmark graph classification datasets from \cite{Morris+2020,hu2020ogb,stoll2025graphbenchnextgenerationgraphlearning, dwivedi2022long,dwivedi2020benchmarking}.}
    \label{fig:combined_ratio_histogram}
\end{wrapfigure}

On the other hand, existing GNN inductive benchmarks—ranging from the recent GraphBench~\citep{stoll2025graphbenchnextgenerationgraphlearning} to established ones like OGB~\citep{hu2020ogb}, TUDataset~\citep{Morris+2020}, and LRGB~\citep{dwivedi2022long}—predominantly operate in the opposite regime. As quantified in Figure \ref{fig:combined_ratio_histogram}, which visualizes major classification dataset statistics across these suites, the prevailing landscape is almost exclusively defined by the $n \gg p$ ratio regime. Consequently, while these benchmarks have driven consistent progress in \textquote{many-graph} learning, they fail to reflect the $n \ll p$ structure inherent of omics data: massive graphs with scarce labeled samples. %massive graphs observed over a scarce number of labeled samples.

Notably, in statistical learning the $n \ll p$ regime is notoriously prone to overfitting and instability \citep{tibshirani1996regression, hastie2009elements}. Operating here requires regularization and strong inductive biases to extract a meaningful signal from the noise. Given their ability to leverage structural priors, GNNs could be ideal candidates for this task, as demonstrated by their success in molecular property prediction \citep{gilmer2017neural} and drug discovery \citep{stokes2020deep}. However, that success has largely been confined to settings where training data is abundant. % and the statistical regime favors deep learning. 
In the data-scarce omics setting, the advantages of GNNs are less clear. For instance, recent work shows that ChebNet-based architectures rarely outperform traditional machine learning baselines on transcriptomic datasets \cite{Brouard24}. Such results suggest that the structural benefits of GNNs may be surpassed by the challenges of the $n \ll p$ regime, prompting the question: do current GNN architectures consistently underperform in these high-node, low-sample environments?

Answering this question is further complicated by the current pipeline heterogeneity in the omics field \cite{gaudilliere2023harnessing}. %Beyond these statistical hurdles, a pervasive lack of reproducibility prevents the rigorous evaluation of omics models. 
Unlike computer vision or natural language processing benchmarks where the input data is often static and standardized, biological datasets require extensive upstream processing before a graph can even be constructed. Studies often use different datasets and if they use the same ones, critical steps—such as probe-to-gene aggregation, normalization, and covariate adjustment—%, and specific split definitions
are frequently undocumented or implemented inconsistently \cite{yang2025motgnn}. These discrepancies mean reported performance gains may stem from preprocessing choices rather than architectural innovations. Consequently, the field currently lacks a consistent benchmarking infrastructure that separates model performance from pipeline artifacts, obscuring whether GNNs are truly advancing the state of the art.

\begin{figure*}[!t]
  \centering
  \includegraphics[width=\linewidth]{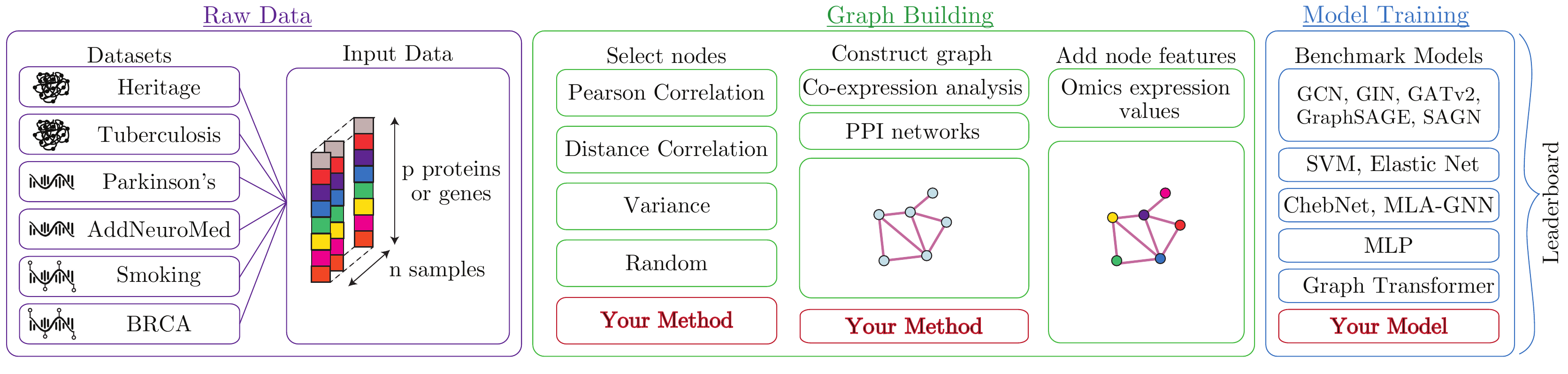}
  \caption{\textbf{Overview of \texttt{OmicsGBench}}: First GNN benchmark platform for omics graph datasets. Left: Transcriptomics, proteomics, or epigenomics expression data across $p$ genes/proteins and $n$ samples. Icons correspond to modality. Middle: 1) Co-expression or PPI graphs are constructed using classical omics approaches; each sample becomes a graph with nodes representing genes/proteins and normalized expression values as node features. 2) A model is trained on a graph-level classification task. Right: Our reference benchmark trains 128k models across hyperparameters, dataset statistics, graph structures, and classification tasks.}
  \label{fig:overview}
\end{figure*}

\textbf{Contributions.} To address these challenges, % and lay the groundwork for analyzing GNNs in the $n \ll p$ regime, 
we introduce \texttt{OmicsGBench}, the first benchmark framework explicitly designed to evaluate inductive graph learning in low-sample, large-graph omics settings (see Figure~\ref{fig:overview} for a pipeline overview). Our contributions are:%A practical barrier to progress in omics deep learning is that comparable datasets are often not actually comparable: the most consequential steps live upstream of modeling (probe-to-gene aggregation, normalization and covariate adjustment, filtering, and split definitions), and are frequently undocumented or re-implemented differently across papers. OmicsGBench treats this upstream pipeline as part of the benchmark contribution by releasing cleaned, model-ready datasets (on Hugging Face) and preprocessing code (on Github) alongside the evaluation protocol. 

\begin{itemize}[leftmargin=*,nosep,topsep=1pt,itemsep=1pt,parsep=0pt,partopsep=0pt]

    \item \textbf{Standardized omics datasets for the $n \ll p$ regime.}  
    We integrate critical upstream processing—e.g. probe-to-gene aggregation, normalization, covariate adjustment—directly into the benchmark pipeline. Alongside this open-source preprocessing code, we release a suite of cleaned, model-ready graph classification tasks spanning neurology, oncology, infectious disease, and cardiorespiratory fitness. These datasets provide a rigorous, reproducible testbed with standardized splits, ensuring performance differences reflect model capability rather than preprocessing artifacts.
    
    \item \textbf{An extensible, open benchmarking platform.}
    \texttt{OmicsGBench} provides a complete, end-to-end infrastructure to facilitate immediate adoption—including cleaned datasets hosted on Hugging Face, automated preprocessing scripts, and standardized training and evaluation pipelines. Its modular architecture is explicitly designed for extensibility: omics practitioners can seamlessly integrate custom datasets, while AI researchers can rapidly evaluate novel architectures. To foster progress, we maintain a companion website at \url{https://OmicsGbench.github.io} featuring interactive graph statistics and a public leaderboard for tracking community progress.
    %\item \textbf{Open ecosystem for future benchmarking.}  
    %\texttt{OmicsGBench} provides an end-to-end infrastructure—cleaned datasets on Hugging Face, preprocessing scripts, configurations, models, and evaluation pipelines—to facilitate immediate adoption. Our modular is designed for extensibility, allowing omics practitioners to easily plug in new datasets and AI researchers to benchmark novel architectures. Additionally, we provide a companion website at \url{https://OmicsGBench.org} for visualizing graph statistics across varied settings and tracking current benchmarking results on a shared leaderboard.
    
    \item \textbf{Enabling controlled inductive bias studies.}  
    \texttt{OmicsGBench} decouples node selection and graph construction from the learning pipeline, allowing researchers to swap strategies (e.g., varying the $n/p$ ratio, comparing edge construction strategies) without altering the whole workflow. We use this design to benchmark feature selection methods and evaluate how graph structure influences performance in high-dimensional settings.

    \item \textbf{A reference evaluation of GNNs vs. classical baselines.}  
    To demonstrate the platform's capabilities, we benchmark a diverse array of models: classical GNNs and graph transformers (GCN, GATv2, GIN, GraphSAGE, GPS), scalable models (SAGN), and omics-specific architectures (MLA-GNN, ChebNet), comparing these against MLPs and traditional machine learning techniques (SVM, Elastic Net) to determine whether structural learning provides a tangible advantage over simpler methods. The framework's modular design ensures new models can be rapidly integrated and evaluated against these baselines.

\end{itemize}

Through this reference evaluation, our experimental results reveal a critical insight: widely used GNNs often struggle to convincingly outperform simple MLPs and classical baselines in the omics domain. %\color{green} Our ablations also investigate why: GNN performance tracks which molecular features are selected (RQ3) and whether the readout preserves node identity (RQ2), rather than the graph itself, as curated interaction networks offer no consistent advantage over co-expression graphs (RQ4). \color{black} %This lack of predictive gain is consistent across varying graph structures and regularization techniques. 
These findings echo recent warnings that \textquote{graph learning will lose relevance due to poor benchmarks} \citep{bechlerspeicher2025positiongraphlearninglose} if the field blindly applies graph methods to data without questioning the validity of the underlying structure. By facilitating objective evaluation, \texttt{OmicsGBench} sheds light on the true utility of GNNs in high-stakes scenarios, redirecting attention from incremental performance gains to a critical reassessment of the learning paradigm itself: does the path forward lie in specialized architectures, or in rethinking how, and if, biological priors should be encoded as graphs?

\section{Related Work}

%We review prior work on GNNs related to benchmarking and its application to omics (an extension to low-sample learning can be found in Appendix \ref{sec:extended_related_works}). %We review prior work on GNNs related to benchmarking, application to omics, and evaluation in low-sample regimes.

\textbf{Inductive GNN Benchmarking.} The reliability of GNN research has recently been called into question, with warnings that the field risks stagnation due to poor benchmarking practices and a lack of real-world relevance \citep{bechlerspeicher2025positiongraphlearninglose}. While foundational efforts—such as TUDataset \citep{Morris+2020}, OGB \citep{hu2020ogb}, or LRGB \citep{dwivedi2022long}—have successfully driven architectural innovation, they have been criticized for inconsistent evaluation protocols and a reliance on datasets that do not reflect the complexity of modern scientific problems. Crucially, another fundamental limitation shared by these suites is their focus on the $n \gg p$ regime, where the number of graphs $n$ far exceeds the number of nodes per graph $p$ (see Figure \ref{fig:combined_ratio_histogram}). Even the most recent GraphBench \cite{stoll2025graphbenchnextgenerationgraphlearning}, which explicitly aims to address the aforementioned methodological concerns by introducing diverse domains, fails to cover the $n \ll p$ setting characteristic of omics. %Consequently, the community currently lacks a rigorous evaluation framework to evaluate whether GNNs can generalize in high-dimensional, data-scarce environments. %, or if their perceived utility is confined to sample-rich settings.

\textbf{Challenges in Applying GNNs to Omics.}
Driven by the hypothesis that biological function is governed by complex interactomes, GNNs have seen wide adoption across bioinformatics \citep{Zhang21}, with applications across transcriptomics, proteomics and epigenomics \citep{Li25}, spanning drug discovery \citep{jiang2025network}, cancer classification \citep{wang2020_cancer_survival_gcn}, and biomarker discovery \citep{10.1093/bib/bbae658}. However, despite this widespread adoption, the actual predictive utility of GNNs in these settings remains contentious. Recent independent evaluations suggest that structural models often fail to outperform simpler baselines. Notably, \cite{Brouard24} found that ChebNet-based GNNs \citep{defferrard2016convolutional} rarely yield performance gains over traditional machine learning methods, despite incurring significantly higher computational costs. Yet, their study examined only a single dataset in the $n \ll p$ regime and did not investigate how performance is influenced by upstream choices like node selection, graph construction, or regularization. By contrast, \texttt{OmicsGBench} systematically varies these factors under standardized protocols across many models and datasets.

% \paragraph{Challenges in Applying GNNs to Omics.}
% GNNs have been applied to omics data \citep{jiang2025network,Zhang21}, motivated by their ability to model underlying biological networks. Applications span single-cell transcriptomics \citep{Li25}, cancer classification \citep{wang2020_cancer_survival_gcn}, and biomarker discovery \citep{10.1093/bib/bbae658}. Despite growing interest, recent evaluations have raised questions about whether GNNs offer predictive advantages over simpler models in omics. Notably, \citet{Brouard24} found that a ChebNet \citep{10.5555/3157382.3157527} based GNN rarely improves performance over simpler baselines, despite significantly higher computational costs. Importantly, their study only included one dataset in the $n \ll p$ regime and did not analyze how node selection, graph construction, regularization, and sample-to-node $n/p$ ratio affect performance. By contrast, we design \texttt{OmicsGBench} to systematically vary these factors using standardized protocols across a wide range of models and datasets.

\textbf{Lack of GNN Benchmarks for Omics Expression Data.}
While biological graphs are widely used, no standard benchmark exists for graph-level tasks constructed from omics expression data. Existing biological graph datasets fall into two main categories: interaction graphs and protein structure graphs. Interaction graphs from \cite{hu2020ogb}—such as \texttt{ogbg-ppa}, \texttt{ogbl-ppa}, and \texttt{ogbl-biokg}—are built from known biological relationships but do not include sample-level omics expression data; likewise, \texttt{ogbn-proteins} encodes interaction evidence rather than per-sample expression profiles. Protein structure datasets such as D\&D \citep{dobson2003_distinguishing_enzymes} and ENZYMES \citep{schomburg2004brenda} classify protein function from 3D molecular graphs but do not reflect omics measurements, and static databases such as STRING \citep{Szklarczyk2023STRING} and HINT \citep{Patil2005HINT} describe protein--protein associations but are disease-agnostic. Additionally, despite new releases of omics datasets for machine learning applications \citep{mlomics}, they are not tailored to GNNs with graph building pipelines--so no public benchmark exists where each graph is a biological sample, nodes correspond to genes or proteins, and node features encode expression. %. Thus, despite the growth of omics data and their importance in biomedical research, there is no public benchmark where each graph is a biological sample, nodes correspond to genes or proteins, and node features encode expression. 
\texttt{OmicsGBench} fills this gap by introducing omics graph datasets, derived from sample-specific expression profiles and tailored for evaluating models in the $n \sim p$ and $n \ll p$ regimes.

% \paragraph{Benchmarks for Low-Sample Learning.}
% Outside of graph learning, low-sample benchmarking is gaining traction. FLIP \citep{mollon2025exploringlargeproteinlanguage} targets protein language models under data constraints, XTREME-UP \citep{ruder-etal-2023-xtreme} supports multilingual NLP in low-resource settings, and PMLBmini \citep{knauer2024tabmini} provides small-scale tabular datasets. However, none of these efforts address the unique structural and statistical challenges of graph learning in the low-sample regime. \texttt{OmicsGBench} is the first benchmark for graph-level tasks with many nodes and few graphs.

% \begin{table}[ht]
% \centering
% \scriptsize
% \begin{tabular}{|l|l|c|c|c|c|}
% \hline
% \textbf{AuthorYEAR} & \textbf{Dataset Proposal} & \textbf{Graph-Level Tasks} & \textbf{K-Fold Validation} & \textbf{Seeds per Model} & \textbf{\# GNNs Evaluated} \\
% \hline
% Dwivedi et al., 2020 & New datasets & Yes & Yes & Not specified & 5 \\
% Errica et al., 2020  & Existing datasets & Yes & Yes & Not specified & 4 \\
% Hu et al., 2020      & New + existing & Yes & No  & 10             & 11 \\
% Lv et al., 2021      & Existing datasets & Yes & No  & Not specified & 12 \\
% Zhao et al., 2020    & Existing datasets & No (node-level) & Yes & Not specified & 7 \\
% \hline
% \end{tabular}
% \caption{Comparison of representative graph benchmark papers.}
% \label{tab:graph_benchmark_papers}
% \end{table}

\section{Overview of \texttt{OmicsGBench}} \label{sec:benchmark-design}

We design \texttt{OmicsGBench} as a modular framework (Figure~\ref{fig:overview}) that adapts the flexible, configuration-driven architecture of \texttt{TopoBench}~\citep{telyatnikov2025topobench} to the specific requirements of the omics domain. By building upon this robust backbone, which is naturally compatible with GNNs, \texttt{OmicsGBench} provides a standardized pipeline comprising:
\begin{enumerate}[leftmargin=*,nosep,label=(\roman*)]
    \item a dataset selection module,
    \item a graph building module, which includes critical upstream choices of node-selection and %that maps data to a node index set, 
    edge-construction methods, 
    \item a model selection module, encompassing classical baselines, MLPs, and a diverse suite of GNNs (scalable, omics-specific, and general-purpose) with customizable encoders and readouts,
    \item a training procedure with rigorous regularization controls for the $n \ll p$ regime, and
    \item a performance evaluation module. 
\end{enumerate}
This structure enables controlled ablations and rapid iteration. New node-selectors and edge-constructors can be tested by swapping one component while keeping splits, preprocessing, and training fixed. We describe each module next:

% We design \texttt{OmicsGBench} as a modular framework (Fig.~\ref{fig:overview}) with: 
% \begin{enumerate}[leftmargin=*,nosep,label=(\roman*)]
%     \item a dataset selection module,
%     \item a graph building module, which includes choices of node-selection and %that maps data to a node index set, 
%     edge-construction methods, 
%     \item a model selection module-- machine learning baselines, MLPs, and classical, large-scale and omics-specific GNNs for which practitioners can additionally vary encoder (of omics expressions into higher dimensional features on nodes) and readout,
%     \item a training procedure with hyperparameters controlling overfitting such as dropout and regularization, and,
%     \item a performance evaluation module. 
% \end{enumerate}
% This modularity enables controlled ablations and rapid iteration. New node-selectors and edge-constructors can be evaluated by swapping a single component while keeping splits, preprocessing, and training fixed. We detail each module in this section.
%\textit{(i)} a node-selection module that maps data to a node index set, \textit{(ii)} a graph-construction module that constructs edges based on correlations computed from the selected nodes' feature data, \textit{(iii)} model selection, where the practitioner can select individual encoder, backbone, and readout models, \textit{(iv)} a training procedure where parameters such as dropout and regularization can be set, and \textit{(v)} a model evaluation module. 
%Details on the experiments can be found in App. \ref{sec:experiment_config}.

\subsection{Datasets}
To demonstrate the platform, we release a suite of cleaned, model-ready graph classification tasks built from six publicly available omics datasets spanning neurology, oncology, infectious disease, and cardiorespiratory fitness \citep{addneuromed, shamir2017parkinsons, robbins2021human, Creighton2013, tsaprouni2014cigarette, kunnath2010dynamic}. The datasets were chosen for their biomedical relevance and cover transcriptomic, proteomic, and epigenomic modalities. Each is deposited on Hugging Face and transformed at runtime into featured graphs whose nodes are genes or proteins, with node features given by expression, abundance, or promoter methylation. Graphs are converted to PyTorch Geometric objects \citep{fey2019fast} and evaluated with stratified 5-fold cross-validation, with all data-dependent preprocessing fitted on training samples only (Appendix~\ref{sec:Preprocessing}). Splits are random by default and group-disjoint where grouping metadata exists: Parkinson's is grouped by hybridization-date batch and Tuberculosis by country of collection. AddNeuroMed's two batches (microarray platforms) are too few for group-disjoint 5-fold splits, so we split randomly and remove batch effects with ComBat fit on each fold's training samples. Heritage, Smoking and BRCA provide no family, batch or plate identifiers and are split randomly (for BRCA, as in MLOmics \cite{mlomics}). We flag further batch-aware splitting as a direction for a future release, and note the absence or non-use of grouping metadata as a limitation of the public data and current pipeline rather than the splitting framework itself. Table~\ref{tab:datasets} summarizes dataset characteristics and class distributions. 

\textbf{Heritage Plasma Proteomics: exercise response.}
Aptamer-based plasma proteomic profiles of $n=654$ sedentary adults from the HERITAGE Family Study \citep{robbins2021human}. The binary task predicts exercise responders, defined as a relative VO$_2$max improvement above 15\% after a 20-week exercise intervention \citep{jones2008safety,warburton2006health,warburton2006prescribing}.

\textbf{Tuberculosis Proteomics: culture status.}
Serum antibody responses against the \textit{Mycobacterium tuberculosis} proteome, measured on a protein microarray for $n=561$ suspected TB patients \citep{kunnath2010dynamic}. The binary task predicts culture result (negative versus positive).

\textbf{Parkinson's Transcriptomics: cognitive status.}
Whole-blood microarray gene expression from $n=535$ individuals in the GENEPARK consortium \citep{shamir2017parkinsons}, including idiopathic Parkinson's disease patients, healthy controls, and individuals with other neurodegenerative diseases. The binary task separates MCI/Normal from Dementia (Montreal Cognitive Assessment, MoCA < 21 \citep{nasreddine2005montreal, dalrymple2010MoCA}).

\textbf{AddNeuroMed Transcriptomics: clinical diagnosis.}
Illumina microarray gene expression from $n=711$ subjects in the AddNeuroMed study \citep{addneuromed}, measured on two platforms. The three-class task distinguishes controls (CTL), mild cognitive impairment (MCI), and Alzheimer's disease (AD).

\textbf{Smoking Epigenomics: smoking status.}
Whole-blood DNA methylation profiles from $n=464$ individuals \citep{tsaprouni2014cigarette}, summarized to gene-level promoter methylation following the MLOmics protocol. The binary task separates never-smokers from ever-smokers.

\textbf{BRCA Epigenomics: breast cancer subtype.}
DNA methylation profiles of Breast Invasive Carcinoma from TCGA \citep{Creighton2013} for $n=640$ samples, in the gene-level form provided by MLOmics \citep{mlomics}. The task is four-class PAM50 subtype classification.

% .. and thus providing the hardest test of whether graph topology adds value.
% -> Remove per rebuttal request. 

%The task is four-class classification of breast cancer subtypes: Luminal A, Luminal B, HER2-enriched, and Basal-like.

%\begin{table*}[h]
%\centering
%\small{
%\caption{Omics graph classification datasets in \texttt{OmicsGBench} across proteomics and transcriptomics modalities. The number of graphs is per dataset is denoted $n$ and the number of nodes per graph is $p$. \label{tab:datasets}}
%\begin{tabular}{lcccccc}
%\hline
%\textbf{Dataset} & \textbf{Modality} & \textbf{n} & \textbf{p} & \textbf{Target} & \textbf{Class distribution} \\
%\hline
%HERITAGE & Proteomics & 654 & 4977 & $\Delta$VO$_2$max $> 15\%$ 
%& $\boldsymbol{\leq 15\%}$: 279; $\boldsymbol{> 15\%}$: 375 \\
% \hline
%Parkinson’s & Transcriptomics & 535 & 21755 & MoCA score $\geq 21$
%& $\boldsymbol{\geq 21}$: 332; $\boldsymbol{< 21}$: 203 \\
% \hline
%AddNeuroMed & Transcriptomics & 711 & 17198 & Diagnosis
%& \textbf{AD}: 284; \textbf{CTL}: 238; \textbf{MCI}: 189 \\
%BRCA & Epigenomics & 640 & 19049 & Cancer Subtype
%& \textbf{LumA}: 353; \textbf{LumB}: 132; \textbf{Her2}: 42; \textbf{Basal}: 113
%\hline
%\end{tabular}
%}
%\end{table*}

% New table with extra line at bottom because not enough space for distribution col in Neurips format
\begin{table*}[h]
\centering
\small{
\caption{Omics graph classification datasets in \texttt{OmicsGBench} across proteomics, transcriptomics, and epigenomics modalities. The number of graphs per dataset is denoted $n$ and the number of nodes per graph (before node selection) is $p$. \label{tab:datasets}}
\begin{tabular}{lccccc}
\hline
\textbf{Dataset} & \textbf{Modality} & \textbf{n} & \textbf{p} & \textbf{Target} & \textbf{Class distribution} \\
\hline
Heritage & Proteomics & 654 & 4976 & $\Delta$VO$_2$max $> 15\%$
& $\boldsymbol{\leq 15\%}$: 279; $\boldsymbol{> 15\%}$: 375 \\
Tuberculosis & Proteomics & 561 & 3814 & Culture result
& \textbf{Neg}: 307; \textbf{Pos}: 254 \\
Parkinson's & Transcriptomics & 535 & 21755 & MoCA score $\geq 21$
& $\boldsymbol{\geq 21}$: 332; $\boldsymbol{< 21}$: 203 \\
AddNeuroMed & Transcriptomics & 711 & 17197 & Diagnosis
& \textbf{AD}: 284; \textbf{CTL}: 238; \textbf{MCI}: 189 \\
Smoking & Epigenomics & 464 & 20763 & Smoking status
& \textbf{Never}: 179; \textbf{Ever}: 285 \\
BRCA & Epigenomics & 640 & 19049 & Cancer subtype
& \begin{tabular}[t]{@{}c@{}}
\textbf{LumA}: 353; \textbf{LumB}: 132; \\
\textbf{Her2}: 42; \textbf{Basal}: 113
\end{tabular} \\
\hline
\end{tabular}
}
\end{table*}

%Select Nodes: Subsampling and Feature Selection
\subsection{Node Selection: Subsampling and Feature Selection}\label{subsec:subsampling}

To address the $n \ll p$ challenge, a common strategy is to reduce $p$ via feature selection, improving the $n/p$ ratio and mitigating overfitting. To support this step within the platform, we include four reference strategies: two standard univariate filters commonly used by the omics community (variance and Pearson correlation), a non-linear filter (distance correlation), and random selection (as a control). These strategies are filter methods, intentionally selected because they are independent of the subsequent classification model, unlike wrapper or embedded techniques. Importantly, node selection is implemented as an interchangeable submodule with a standardized interface (inputs: training features/labels; output: selected node indices). To illustrate the modularity of our pipeline, we provide distance correlation as an example, showing how practitioners can easily plug in custom or complex selectors while leaving edge construction and model training fixed. As with all \texttt{OmicsGBench} preprocessing, node selection only considers the training split to prevent data leakage.

% To address the $n \ll p$ challenge of omics data, one may decide to downsample nodes to regain the advantages of the $n \gg p$ regime. To assess whether selecting the most informative nodes \textit{a priori} improves performance, we include three options: random selection (as a control) and two univariate feature selection strategies that are commonly used by the omics community. Both selection strategies are filter methods, intentionally selected because they are independent of the subsequent classification model applied, unlike wrapper or embedded methods. Importantly, node selection is implemented as an interchangeable submodule with a standardized interface (inputs: training features/labels; output: selected node indices). This design allows users to plug in alternative selectors (e.g., embedded or learned feature selection) while leaving edge construction and model training unchanged. Node selection is done using only the training dataset to avoid data leakage and is applied to the validation and test sets. 

\textbf{(1) Variance-based filtering:} Features are ranked by standard deviation across samples and the top $p$ retained. Common in transcriptomic preprocessing \citep{bommert21, LABORY20241274, wang2021scgnn, tan2025amogel}, this serves as a simple proxy for information: highly variable genes/proteins are more likely to separate groups, while low-variance features are discarded as uninformative or noisy.

\textbf{(2) Correlation-based filtering:} Features are ranked by absolute Pearson correlation with the target and the top $p$ retained. This method is widely used in omics settings \citep{Mitic25,bommert21, li2022benchmark, PerezRiverol2017Accurate}.

\textbf{(3) Distance Correlation filtering:} To capture complex dependencies that linear metrics might miss, we compute the distance correlation \citep{szekely2007measuring} between each feature and the target. Unlike Pearson correlation, distance correlation is zero if and only if the variables are statistically independent, allowing it to detect non-linear relationships. Including this strategy allows us to test whether non-linear filtering yields better graph inputs for downstream GNNs.

\textbf{(4) Random subsampling:} We randomly select $p$ nodes from the full set, without regard to variance or association with the target. This provides a control for evaluating the benefit of informed feature selection relative to uninformed dimensionality reduction.

%These strategies provide a principled way to probe the impact of node selection—specifically the trade-offs between linear, non-linear, and variance-based priors—on model performance in $n \ll p$ regimes.
These strategies provide a principled way to probe the impact of node selection priors on model performance in $n \ll p$ regimes. Our framework also controls the number of selected nodes ($p$) as a function of the training set size ($n_{train}$), letting practitioners study how performance changes with feature count separately from the choice of selection method (as we do in Section~\ref{sec:experiments}). %We iterate over both of these parameters in Section \ref{sec:experiments}.
%These strategies provide a principled way to probe the impact of node selection, i.e., feature selection, on model performance in challenging $n \ll p$ regimes. 
%In classification tasks, analogous univariate filters include t-tests or ANOVA F-tests to rank features by their class-discriminative power.
%\subsection{Construct Edges: Using Co-Expressions Patterns}
\subsection{Edge Construction: Data-driven vs. Biologically Motivated} \label{subsec:edge_generation}
\textbf{Data-driven Approach: Co-expression.} %We adopt a data-driven strategy for graph construction in omics-based learning: define edges based on pairwise similarity between features. In particular, expression correlation is frequently used to reflect functional or regulatory relationships among genes or proteins, and has been employed across numerous biomedical graph learning works \citep{RAO2021102393, Liu22, 10879447}. Following \citet{Xing22}, we compute a co-expression similarity matrix by calculating the Pearson correlation between each pair of features in the preprocessed training data. The underlying motivation is that features with similar expression patterns often participate in shared biological processes.
As a purely data-driven approach to edge construction, we adopt the gene expression correlation method, widely used in biomedical graph learning \citep{RAO2021102393, Liu22, 10879447}. Following \cite{Xing22}, we compute a co-expression similarity matrix via pairwise Pearson correlation between features in the preprocessed training data, motivated by the observation that co-expressed features tend to share biological function. We transform this matrix into a graph using soft-thresholding, selecting a power that best approximates scale-free topology \citep{vanDam2018Gene}, then binarize the result with a hard threshold retaining the top 10\% of edges (density = 0.1). This heuristic balances sparsity and connectivity: enforcing a single connected component typically yields an overly dense, computationally intractable graph at this dimensionality.%We transform this matrix into a graph using soft-thresholding with the \texttt{PyWGCNA} package \citep{pywgcna}, selecting a power that best approximates scale-free topology \citep{vanDam2018Gene}, then binarize the result with a hard threshold retaining the top 10\% of edges (density = 0.1). This heuristic balances sparsity and connectivity: enforcing a single connected component typically yields an overly dense, computationally intractable graph at this dimensionality.

%To transform this similarity matrix into a graph, we apply soft-thresholding using the \texttt{PyWGCNA} package \citep{pywgcna}, which selects a power that best approximates scale-free topology. This package is a Python version of an R package \citep{WGCNA} widely employed in omics for correlation network analysis of large, high-dimensional data sets \citep{vanDam2018Gene}. The resulting adjacency matrix is then binarized using a hard threshold to obtain an unweighted graph. This threshold is empirically calibrated to to retain the top 10\% of all possible edges (i.e., a graph density of 0.1). We adopt this heuristic as a trade-off between graph sparsity and connectivity; strictly enforcing a single connected component typically yields an overly dense graph that is computationally intractable given the high dimensionality of the data.

\textbf{Using a Biological Prior: Protein-Protein Interactions.} To incorporate functional biological knowledge, we construct edges using the STRING database \citep{Szklarczyk2023STRING}, which aggregates physical and functional protein-protein interactions (PPI) from high-throughput experiments, co-expression, and literature mining, a widely used approach \cite{chereda2021explaining, ramirez, ijcai2018p490, SAADAT2025112812}. We map dataset features (genes or proteins) to STRING identifiers via UniProt or Entrez IDs and consider interactions between any constituent members and retain the maximum combined score. As recommended by STRING \citep{Szklarczyk2023STRING}, we retain \textit{medium-confidence} interactions (\textit{combined score} $\geq 0.4$) and treat them as unweighted edges.

% Crucially, this construction yields a \textit{single global graph topology} shared across all samples.
Crucially, both approaches yield a \textit{single global graph topology} shared across all samples. While the node features (omics expression) are sample-specific, the adjacency matrix is fixed throughout training and inference. However, edge construction is modular so users can swap in alternative priors (e.g., Reactome pathways) or learned structure and compare them under identical preprocessing and training protocols. Note that in the co-expression case the two are not independent: node selection determines which correlations are computed, and hence the edges of the co-expression graph.

\subsection{Models}\label{sec:models}
We implement and provide a range of GNNs within \texttt{OmicsGBench} spanning classical, scalable, and omics specific architectures. Additionally, we include non-relational baselines to serve as reference points to evaluate whether graph structure yields predictive benefits beyond standard machine-learning and deep learning approaches. 
%We implement and evaluate a range GNNs on \texttt{OmicsGBench} spanning classical, scalable, and omics specific architectures. Additionally, we include non-relational baselines to evaluate whether graph structure yields predictive benefits beyond standard machine-learning and deep learning approaches. 
All methods follow their original papers and code.

\textbf{Machine Learning Baselines.}
In the $n \ll p$ regime, linear models with strong regularization are often the most competitive baseline and hence default choice. We include Support Vector Machines (SVM) \citep{cortes1995support} and Elastic Net \citep{zou2005regularization} as our primary sanity checks. These models are widely used in bioinformatics because their objective functions (margin maximization and L1/L2 regularization, resp.) are tailored to handle high-dimensional, sparse data without overfitting. %We do not force these models into the deep learning pipeline used for GNNs and instead implement them using a distinct, optimized workflow that mirrors standard bioinformatics protocols already applied to our selected datasets~\citep{robbins2021human,shamir2017parkinsons,sood2015novel} (details in App. \ref{sec:experiment_config}). While this introduces a pipeline divergence, it keeps the comparison realistic: we compare GNNs against the gold standard implementation of classical methods rather than a version constrained by deep learning architectures. %If a GNN cannot outperform Elastic Net, it suggests that the non-linear graph propagation is adding unnecessary complexity rather than extracting signal.

\textbf{Deep Learning Baseline.} To isolate the contribution of graph topology from non-linear feature transformation, we evaluate a standard Multi-Layer Perceptron (MLP) as a graph-free baseline. Comparing GNNs to it directly tests whether graph structure adds value in omics.

\textbf{Classical GNNs and Graph Transformers}.
We include the most widely adopted GNN and Graph Transformer architectures: GCN \citep{kipf2017semisupervised}, GraphSAGE (SAGE) \citep{graphsage}, GATv2 \citep{brody2022how}, GIN \citep{xu2018how} and GPS \citep{rampavsek2022recipe}. These models represent the default toolkit for graph representation learning. However, they are often applied to biological data without verifying whether their inductive biases %(e.g., isotropic smoothing in GCN vs. attention mechanisms in GAT) 
are suitable for them. %molecular networks.

\textbf{Scalable GNNs.}
Omics graphs are often massive, containing tens of thousands of nodes per sample, which can cause standard GNNs to run out of memory. To address this, we include SAGN \citep{sagn}, a representative scalable architecture selected from the large-scale benchmarking work of \cite{duan2022a}. Scalable GNNs typically decouple feature propagation from training to handle size constraints. Our goal is to assess whether these efficiency-focused design choices, originally optimized for node classification on single giant graphs (e.g., %\texttt{Reddit} \citep{graphsage}, 
\texttt{ogbn-products} \citep{hu2020ogb}, \texttt{Flickr} \citep{graphsaint-iclr20}), remain effective. % when transferred to the graph-level classification of biological networks.

\textbf{Omics-specific GNNs.}
Finally, we evaluate models tailored for biological tasks. We include MLA-GNN \citep{Xing22}, a multi-level attention architecture designed for transcriptomic biomarker discovery, and ChebNet \citep{defferrard2016convolutional}, which utilizes spectral convolutions and has been applied to cancer classification \citep{ramirez}. These models are justified by their ability to encode biological priors or modularity. However, they are often validated on datasets where preprocessing ensures $n \gg p$. %We test the robustness of these domain-specific designs to determine if they truly offer an advantage in the raw $n \ll p$ regime, or if their previous success was dependent on specific feature selection pipelines.

\textbf{Optional Node Identity.} Message passing applies the same function at every node and mixes each node's signal with its neighbors', so marker-specific information can be diluted before it reaches the readout, even a position-aware one. To support controlled tests of this effect, OmicsGBench provides an optional module (\texttt{gene\_identity=learnable}) that concatenates a learnable embedding $e_v \in \mathbb{R}^{d}$ (default $d=32$), shared across samples, to each node's expression before the input encoder, $h_v^{(0)} = \phi([\,x_v \,\Vert\, e_v\,])$, so that message passing can condition on gene identity. The module is compatible with any GNN backbone, and we leave its systematic evaluation, including with pretrained gene embeddings, to future work.

 \textbf{Readouts.} Because all graphs share the same node ordering, GNN node embeddings can be aggregated either by mean pooling (``vanilla readout'', \texttt{no\_readout} in our configuration) or by an MLP readout that flattens node embeddings and applies position-specific weights (\texttt{omics\_readout}). \color{black}

\subsection{Evaluation Metrics}

To ensure standardized comparisons, our evaluation pipeline measures performance according to predictive capability (Accuracy, Macro F1, Weighted F1, and Macro AUROC, ensuring robust assessment across both balanced and imbalanced classes) and computational efficiency. Appendix \ref{sec:details_metrics} details the evaluation module and ~\ref{sec:best_results} reports computational costs.
\section{Experiments}
\label{sec:experiments}

To demonstrate the platform's evaluation capabilities, we benchmark the built-in models across the six provided omics datasets, systematically varying node selection strategies (Section~\ref{subsec:subsampling}), edge construction methods (Section~\ref{subsec:edge_generation}), and sample-to-node ratios ($n_{train}/p \in \{0.3, 0.5, 0.8, 1.0\}$) to establish a comprehensive reference baseline. Detailed pipeline configurations are in Appendix~\ref{sec:setup_details}. We organize results around five research questions: %whether GNNs outperform simple baselines (RQ1), what drives their performance (RQ2–RQ4), and whether these conclusions are robust to model selection (RQ5). 
\color{black}

\begin{figure}[t]
\centering
\includegraphics[width=0.8\linewidth]{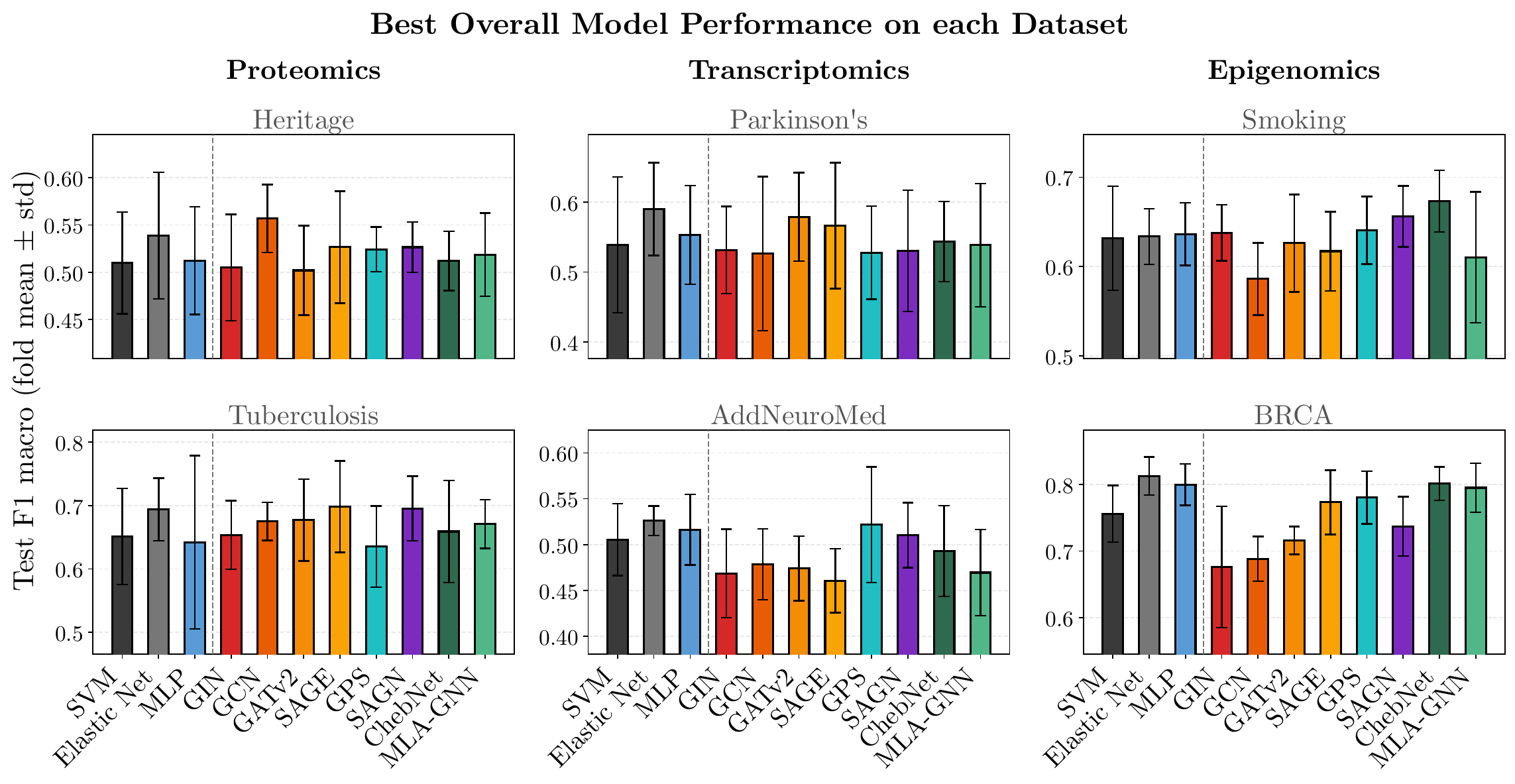}
\caption{\textbf{Best model performance per dataset} (selected by validation $F_{\mathrm{macro}}$, error bars = std across 5 folds). No model type consistently outperforms the others, and non-graph baselines (Elastic Net, MLP) are within one standard deviation of the best GNN on five out of six datasets.}
\label{fig:best_overall}
\end{figure}

\textbf{RQ1: Do Complex Models Consistently Outperform Simple Baselines?}
Figure~\ref{fig:best_overall} shows test performance across model types (full metrics in Tables~\ref{tab:best-configs-with-gpu} and \ref{tab:best-configs-with-gpu-2}), revealing that no single architecture consistently dominates. Notably, Elastic Net (a linear model with no access to graph structure) achieves the highest mean $F_{\mathrm{macro}}$ on Parkinson's, AddNeuroMed, and BRCA, and remains a top performer on all datasets except Smoking. While GNNs lead on Heritage and Smoking and essentially tie Elastic Net on Tuberculosis, these margins are notably small. On every dataset except Smoking, the best non-graph baseline falls within one standard deviation of the top GNN. Furthermore, architectural complexity fails to predict success. The best-performing GNN varies widely by dataset, and the most complex models are not consistently the strongest. For example, GPS excels on AddNeuroMed but performs worst on Tuberculosis. Notably, simple baselines reach this comparable performance at a fraction of the computational cost, with the MLP training fastest on every dataset while using far fewer parameters than most GNNs. To understand why GNNs offer no consistent structural advantage, RQ2 through RQ4 isolate the specific contributions of the readout, node selection, and edge construction mechanisms.

\textbf{RQ2: Do GNNs Exploit Graph Structure or Node Identity?}
Comparing the two readouts introduced in Section~\ref{sec:models} (Figure~\ref{fig:readout_effect}, Appendix~\ref{app:readout_ablation}) reveals the core signals driving GNN performance. Specifically, on Tuberculosis, AddNeuroMed, and BRCA, replacing the MLP readout with mean pooling substantially reduces macro F1 for most backbones, by 0.04 to 0.07 on average and by more than 0.1 in five backbone–dataset pairs. The effect on Heritage, Parkinson's, and Smoking is smaller and inconsistent. We attribute this divergence to the preservation of node identity: while an MLP readout can learn which specific molecular features predict the label, mean pooling largely discards their origin. Consequently, if a task's discriminating signal relies on a specific marker's expression, abundance, or methylation level, pooling inherently destroys this critical information, regardless of whether the graph edges carry useful structural signal. The resulting readout gap therefore measures a task's dependence on node-specific features rather than the informativeness of the graph structure itself. Supporting this view, GNNs perform no better than position-aware baselines—such as Elastic Net, which directly accesses node-specific features—on the three datasets with the largest readout gaps. This is further highlighted by MLA-GNN, which flattens node features from every layer into a single vector before the readout. By retaining node identity regardless of the readout method, it is the backbone least affected by mean pooling on these datasets. To explore this further, OmicsGBench's optional learnable node-identity module (Section~\ref{sec:models}) provides a direct mechanism to test whether restoring identity before message passing can effectively close this performance gap.

\textbf{RQ3: How Does Node Selection Interact with Graph-Based Learning?}
\begin{figure*}[t]
    \centering
    \includegraphics[width=\linewidth]{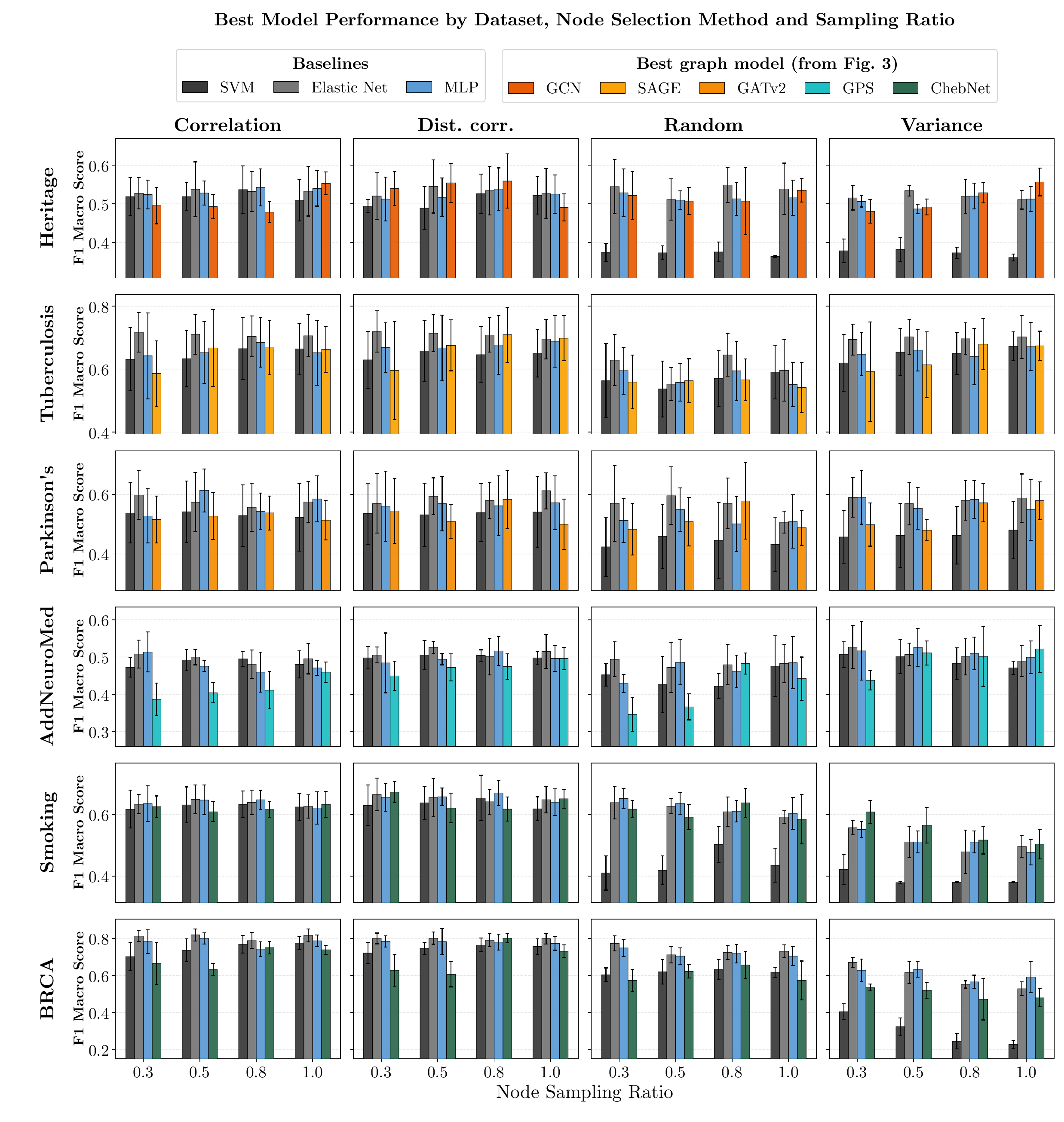}
    \caption{\textbf{Test $F_{\mathrm{macro}}$ by node selection method and sampling ratio.} Baselines versus the best GNN per dataset from Figure~\ref{fig:best_overall}, across selection methods (columns) and sampling ratios ($x$-axis: samples per selected node, so higher ratios retain fewer nodes). Each bar is the best validation configuration for that setting (mean $\pm$ std over five folds). Target-informed selection (correlation, distance correlation) generally matches or outperforms random and variance-based selection across model families. All GNN backbones are shown in Appendix~\ref{sec:ratiovsablation}.}
    \label{fig:ratio_method}
\end{figure*}
Of the design choices we examine, node selection has the most consistent effect on performance (Figure~\ref{fig:ratio_method}). Target-informed selection (correlation and distance correlation, computed only on training folds) generally matches or outperforms unsupervised methods like random or variance-based selection. This advantage is most pronounced on Tuberculosis and BRCA, where random selection degrades every model family. Because this preference holds for graph and non-graph models alike, it reinforces our RQ2 findings: performance depends primarily on which molecular features a model can access, and node selection determines which features those are. The weakness of variance-based selection is clearest on the two epigenomic datasets, where it performs poorly and worsens as fewer nodes are retained (Figures~\ref{fig:full_pt2} and~\ref{fig:ratio_regression}). This aligns with known properties of DNA methylation arrays: some probes reflect sex or genotype rather than methylation \citep{chen2013discovery}, cell-type composition adds further variation in heterogeneous tissue \citep{jaffe2014accounting}, and phenotype-associated differences, such as those linked to smoking, involve only a small fraction of measured sites \citep{joehanes2016epigenetic}. Hence, selecting the most variable features inadvertently favors confounded probes, increasingly so as fewer are kept. Interestingly, under target-informed selection, the sampling ratio (samples per selected node) affects GNNs and baselines in opposite directions. While Elastic Net is flat or declines as fewer nodes are retained, most GNN backbones improve on Tuberculosis and AddNeuroMed (Figures~\ref{fig:full_pt1} and~\ref{fig:ratio_regression}). We hypothesize this reflects internal feature selection: Elastic Net's sparsity penalty benefits from a larger candidate set, whereas GNNs lack this built-in filtering. Instead, each additional node adds readout parameters and introduces noise that message passing propagates to its neighbors. GNNs therefore rely heavily on upstream node selection to remove uninformative nodes, making it a first-class design parameter for omics GNNs.

\textbf{RQ4: Do Biological Priors Improve Graph Construction?}
Curated biological priors do not consistently improve over graphs estimated from the data. We compare STRING-derived protein-protein interaction (PPI) networks against co-expression graphs estimated from each dataset's training split (Figure~\ref{fig:edge_construct_effect}, Appendix~\ref{app:edge_con}). Most differences between the two fall within fold-to-fold variance, and their direction reverses across backbones within every dataset, in contrast to the more consistent readout effects in RQ2. Where a pattern does emerge, it favors co-expression: the best-performing GNN on each dataset uses co-expression edges and outperforms its PPI counterpart on five of six datasets. One hypothesis is that co-expression edges connect features that vary together in the data, so that message passing averages over correlated measurements and reduces independent noise. Tuberculosis is the exception, with most backbones performing better with PPI edges. It is also the only dataset built on pathogen rather than human proteins, with edges drawn from the \textit{M. tuberculosis} STRING network. We hypothesize that this prior may be better matched to the measured features, although a single dataset cannot confirm this. Taken together with RQ2 and RQ3, these results suggest that the choice of edges matters less, and less consistently, than which molecular features are selected and whether the readout preserves their identity. Further supporting this, structural properties of the STRING graphs show at most a weak relationship with the gap between GNNs and the MLP: only feature homophily is weakly associated with it, in the direction expected if edges between co-varying features help, and not after correction for multiple comparisons (Appendix~\ref{sec:struct_propt}). Biological priors therefore do not consistently improve graph construction in our setting.

\textbf{RQ5: How Reliable Is Validation-Based Model Selection?}
In the $n \ll p$ regime, small validation sets make single-best-validation selection noisy. Among the ten highest-ranked configurations by validation $F_{\mathrm{macro}}$, test performance varies considerably more for GNNs than for the MLP (Figure~\ref{fig:val_rank_test}), likely because GNN configurations also vary in readout and edge source, which can shift performance substantially (RQ2, RQ4). To test whether this noise affects our conclusions, we ensemble the top-10 validation configurations by averaging their predicted class probabilities (Appendix~\ref{app:ensemble_selection}). Ensembling improves most models, including the MLP on every dataset, and makes a GNN the top model on five of six datasets rather than three (Figure~\ref{fig:ensemble_comparison}). However, these margins over the best baseline are at most 0.03, within fold-to-fold variance, and the linear baselines were not ensembled. The conclusion of RQ1 therefore holds: no model family consistently outperforms simple baselines. For small validation sets, top-$K$ ensembling is a simple way to reduce sensitivity to the selected configuration, at $K$ times the inference cost. \color{black}

\section{Conclusion}
\texttt{OmicsGBench} establishes the first standardized benchmark framework for graph-level learning on omics expression data, a domain whose defining characteristic is the $n \ll p$ regime. It provides a modular infrastructure for reproducible and easily extensible research, and we invite the community to contribute new datasets, graph constructions and models. Our results show that standard GNNs do not consistently outperform classical ML and simple MLPs, questioning the assumed benefit of structural inductive biases in this setting. These results concern message passing over a single shared topology, a standard construction in omics GNN work \citep{pfeifer2022gnn, pfeifer2023ensemble, tripathy2025,10879447, ijcai2018p490, ramirez, chereda2021explaining}, rather than graph learning on omics in general. They argue not for abandoning graphs but for testing, rather than assuming, the value of a given graph construction, and for more careful graph construction and architectures tailored to large-scale biological networks. The suite's public release aims to foster innovation at the intersection of geometric deep learning and biomedicine. %As a future direction, systematically exploring the adjacency threshold (which directly controls graph sparsity and structure) may yield important insights into when and how GNNs can be optimized for omics.  The suite's public release aims to foster innovation at the intersection of geometric deep learning and biomedicine.

% \input{Sections/6.Old_Experiment}
% \section*{Software and Data}

% If a paper is accepted, we strongly encourage the publication of software and
% data with the camera-ready version of the paper whenever appropriate. This can
% be done by including a URL in the camera-ready copy. However, \textbf{do not}
% include URLs that reveal your institution or identity in your submission for
% review. Instead, provide an anonymous URL or upload the material as
% ``Supplementary Material'' into the OpenReview reviewing system. Note that
% reviewers are not required to look at this material when writing their review.

% \textbf{Do not} include acknowledgements in the initial version of the paper
% submitted for blind review.

% If a paper is accepted, the final camera-ready version can (and usually should)
% include acknowledgements.  Such acknowledgements should be placed at the end of
% the section, in an unnumbered section that does not count towards the paper
% page limit. Typically, this will include thanks to reviewers who gave useful
% comments, to colleagues who contributed to the ideas, and to funding agencies
% and corporate sponsors that provided financial support.

\paragraph{Limitations} \label{sec:Limitations}
OmicsGBench currently provides six reference datasets across three omics modalities (transcriptomics, proteomics, epigenomics), so conclusions may not generalize to other modalities (e.g., metabolomics, single-cell, spatial transcriptomics), disease areas, or cohort sizes. All samples share a single fixed topology. This shared topology is a deliberate first step: by holding the adjacency matrix constant across samples, we isolate the question of whether message-passing over a shared biological graph adds value beyond node features, decoupling structural inductive biases from sample-adaptive graph learning. This design choice may however underrepresent biological heterogeneity across individuals or disease subtypes, and sample-adaptive topology construction is an important direction for future work. For tractability, reference experiments use $n_{train}/p \in [0.3, 1]$. The released datasets support the full-$p$ regime ($n/p \approx 0.02$ to $0.15$).  \newpage %Finally, the full reference benchmark evaluation presented in this work requires substantial compute (8 NVIDIA A100 GPUs), which may be prohibitive for some practitioners.

% Acknowledgements should only appear in the accepted version.
\section*{Acknowledgements}

L. C., G. B., and N. M. acknowledge partial support from NSF Career grant 2240158, NSF grant 2602079, and from the Noyce Foundation. 
L. C. also acknowledges support from an NSF GRFP Fellowship and from the Chan Zuckerberg Foundation. 
G. B. also acknowledges partial support from Arlequin AI.
L. V. L. acknowledges partial support from the predoctoral program AGAUR-FI ajuts (2025 FI-3 00065) Joan Oró, of the Department of Research and Universities of the Generalitat of Catalonia, as well as the European Social Plus Fund. Additionally, we thank Rowan Saloner for helpful discussions and valuable feedback.

\bibliography{references}
\bibliographystyle{plain}

\newpage
\appendix

%\section{Extended Related Works} %\label{sec:extended_related_works}

%\paragraph{Benchmarks for Low-Sample Learning.}
%Outside of graph learning, low-sample benchmarking is gaining traction. FLIP \citep{mollon2025exploringlargeproteinlanguage} targets protein language models under data constraints, XTREME-UP \citep{ruder-etal-2023-xtreme} supports multilingual NLP in low-resource settings, and PMLBmini \citep{knauer2024tabmini} provides small-scale tabular datasets. However, none of these efforts address the unique structural and statistical challenges of graph learning in the low-sample regime. \texttt{OmicsGBench} is the first benchmark for graph-level tasks with many nodes and few graphs.

\section{Dataset Preprocessing}\label{sec:Preprocessing}

Detailed preprocessing steps for each dataset are outlined below.

\subsection{Shared Pipeline and Leakage Control}

All datasets are published to Hugging Face as parquet files in an uncorrected format. To prevent data leakage, any step that estimates parameters from the data is fitted strictly within the training split of each fold. We employ stratified $5$-fold cross-validation, with fold assignment using a fixed random seed for reproducibility. For a given fold index $i$, fold $i$ serves as the test set, fold $i+1 \bmod 5$ is the validation set, and the remaining three folds comprise the training set, resulting in an approximate $60/20/20$ partition that rotates across all five folds. Within each fold, we apply the following sequence: dataset-specific corrections that precede imputation (described below), mean imputation of missing values, any remaining corrections (smoking median-centering), node selection, graph construction, and per-feature $z$-score standardization. Imputation means, correction coefficients, selected nodes, and standardization statistics are all calculated using only the training samples of that fold, and are subsequently applied to the validation and test samples without modification. Co-expression (WGCNA) adjacency is likewise estimated from the training samples only. When STRING adjacency is used, database edges are restricted to that training-selected node set. Node selection retains $p$ features such that the node sampling ratio $n_{\text{train}}/p$ takes a fixed value in $\{0.3, 0.5, 0.8, 1.0\}$, so the number of retained features scales with the training set size.

\subsection{Heritage}
We downloaded the MoTrPAC HERITAGE SomaLogic proteomics matrix and the associated analyte annotation file, utilized the analyte table to select and identify valid protein columns, and filtered the cohort to participants with complete baseline and post-training $\mathrm{VO}_2\max$ values. This resulted in $n=654$ participants and $4976$ proteins. We calculated the relative $\mathrm{VO}_2\max$ change as $\Delta_{\mathrm{rel}} = (\mathrm{VO}_2\max_{\mathrm{post}} - \mathrm{VO}_2\max_{\mathrm{base}})/\mathrm{VO}_2\max_{\mathrm{base}}$ and established a binary target label defined by $\Delta_{\mathrm{rel}} > 0.15$ \cite{jones2008safety, warburton2006health, warburton2006prescribing}. Baseline covariates (age, sex, BMI, race) were extracted, and analytes missing more than $10\%$ of their values were removed. This initial filter is independent of the target labels and is applied prior to data splitting. Any remaining missing values are handled later via train-only imputation. Protein abundances were log$_2$-transformed, a step that operates on each measurement independently and therefore prevents information leakage across samples. Covariate adjustment is \emph{not} performed prior to cross-validation. Instead, within each fold, we fit a linear model ($\text{protein} \sim \text{covariates}$) for each protein using only the training samples and one-hot encoding for categorical variables. We then subtract the resulting covariate-driven component (centered at the training mean covariate profile) from all three splits using those frozen coefficients.

\subsection{AddNeuroMed}
Expression data from two microarray platforms (GPL6947 and GPL10558) were downloaded and processed separately. Probe-level expression values were aggregated to the gene level by averaging all probes mapping to the same gene identifier, utilizing a gene-probe mapping derived from the platform annotation files. Only genes common to both platforms were retained, yielding $17197$ genes. The two platform-specific datasets were then merged, and clinical status labels were extracted from the GEO metadata. Samples with ambiguous or transitional status labels (CTL to AD, MCI to CTL, OTHER, and borderline MCI) were excluded, leaving $n=711$ samples across three classes. The platforms are concatenated \emph{without} batch correction prior to data splitting, and the platform of origin is recorded as a batch label. Within each fold, parametric ComBat (batch location and scale) \cite{yu2024assessing} is fitted exclusively on the training samples, and the resulting parameters are then applied to the validation and test samples. Class labels are not included as covariates in the ComBat adjustment.

\subsection{Parkinson's}
Gene expression data and Montreal Cognitive Assessment (MoCA) scores were extracted from the GEO series matrix file. Affymetrix probe IDs were mapped to gene symbols using the GPL570 platform annotation file. When multiple probes mapped to a single gene, their expression values were averaged to produce one gene-level measurement, resulting in $21755$ genes. Samples missing MoCA scores were excluded, leaving $n=535$ samples. The continuous MoCA scores were dichotomized based on standard clinical interpretations: scores of $21$ or higher were classified as MCI/Normal (class $0$), and scores below $21$ as Dementia (class $1$), aligning with established diagnostic thresholds for cognitive impairment \cite{dalrymple2010MoCA, ismail2025sensitivity, dautzenberg2021clinical}. This dataset exhibits a strong hybridization-date batch structure. Rather than attempting statistical correction, we account for this by design: the GEO characteristics include a hybridization-date batch field, which we use as the grouping key for stratified group $k$-fold partitioning. This ensures every batch falls entirely within a single fold, meaning no batch is ever shared across the training, validation, and test splits. Consequently, no explicit batch-correction algorithm is applied to these data.

\subsection{Tuberculosis}
This dataset measures antibody responses against the \textit{Mycobacterium tuberculosis} proteome using patient sera on a full-proteome protein microarray \citep{kunnath2010dynamic}. The initial features in the GEO series matrix consist of microarray spot coordinates. We map these spots to H37Rv locus tags using the GPL9790 platform table, and subsequently map the locus tags to UniProt accessions via the H37Rv reference proteome (UP000001584). Any spots failing to resolve to a UniProt accession are discarded. Because the array prints multiple segments for the same protein, measurements from spots sharing a locus tag are averaged together to create a single value. This yields $3814$ protein-level features. The classification target is the \textit{M. tuberculosis} culture result, representing the bacteriological confirmation of active disease rather than a direct serological readout (thus, a negative culture does not guarantee the absence of antibodies). Samples recorded as ``not applicable'' are excluded, leaving $n=561$ samples for a binary task distinguishing culture-negative (class $0$) from culture-positive (class $1$). Samples were collected in multiple countries, and culture status is confounded with country of collection. We therefore use stratified group $k$-fold partitioning grouped by country, so that no country appears in more than one split. Default graphs are constructed with WGCNA on the training split. When STRING adjacency is selected, edges are retrieved for the \textit{M. tuberculosis} taxonomy ID $83332$ using the UniProt accessions.

\subsection{Smoking}
This dataset contains whole-blood DNA methylation profiles from $n=464$ individuals \citep{tsaprouni2014cigarette}. We follow an analogous gene-level construction to the BRCA dataset \citep{mlomics}, selecting one promoter probe per gene and applying per-gene median-centering, except that we perform these steps within each training split rather than relying on a preprocessed matrix. We retain Illumina 450k probes annotated as TSS1500 or TSS200, leaving $139125$ candidate probes \citep{jiao2014systems}. Within each fold, the never-smokers in the training split serve as the normal tissue reference used in the MLOmics protocol: each gene retains the candidate probe with the lowest mean $\beta$ value in that reference group, any remaining missing values are imputed using training column means, and each gene is median-centered based on its training median. This process yields approximately $20763$ genes, with slight variations across folds when all candidate probes for a specific gene are missing among the training never-smokers. The original GEO classifications (never, former, current) are consolidated into a binary task of never-smokers versus ever-smokers (former and current combined).

\subsection{BRCA}
This dataset contains DNA methylation profiles of Breast Invasive Carcinoma from TCGA \citep{Creighton2013}, utilizing the preprocessed version provided by MLOmics \citep{mlomics}. MLOmics identifies methylation regions from the metadata and maps them to genes using metrics such as the average methylation ($\beta$-values) of promoters, defined as $500$\,bp upstream and $50$\,bp downstream of the transcription start site. Only regions with a coverage of $\geq 20$ in $70\%$ of tumor samples are retained. The data are normalized by median-centering using the R package \texttt{limma}, and for genes with multiple promoters, the one with the lowest methylation level in normal tissues is chosen as the representative. We employ the \texttt{Original} feature scale, which preserves the full set of genes extracted from the processed methylation files. We discarded the normal-like PAM50 subtype, leaving $n=640$ samples, $19049$ genes, and a four-class subtype classification task. We include this epigenomic modality because it reflects stable chromatin state rather than gene activity, which makes both edge sources biologically uncertain. Co-expression captures regulatory co-variation rather than functional relationships, and PPI edges require an approximate gene-to-protein mapping.

\clearpage

\section{Experimental Setup Details} \label{sec:setup_details}

This appendix provides supplementary details regarding the experimental setup used in this work to ensure reproducibility.

\subsection{Hardware}\label{sec:hardware}

All experiments were run on a server with 128 CPU cores, 1 TB RAM, 4.5 TB Solid State Drive storage, and 8 NVIDIA A100 GPUs (80 GB each).

\subsection{Model Configurations}\label{sec:experiment_config}

\paragraph{Training Setup.} Models are trained in PyTorch Lightning for up to 200 epochs (minimum 50), with early stopping (patience $=50$) based on validation F$_{macro}$ to reduce overfitting. Adam optimizer~\citep{kingma2015adam} is used with a Step Learning Rate schedule (step\_size $=50$, $\gamma=0.5$), and the best checkpoint is selected via validation check-pointing each epoch. 

\paragraph{Hyperparameter Optimization of Deep Learning Models.} For each model--dataset configuration, and for each combination of readout, adjacency source, node-sampling ratio, and node-selection method, we search architecture-specific hyperparameters (Table~\ref{tab:grid-search-space}) with a TPE sampler \cite{bergstra2011algorithms}. The search evaluates seven trials, of which the first three are random startup trials. Every trial is trained once on each of five stratified folds with a shared training seed. For a fold index $i$, fold $i$ is the test set, fold $i+1 \bmod 5$ is the validation set, and the remaining three folds are the training set. Parkinson’s uses stratified group k-fold on hybridization-date batch, and Tuberculosis on country of collection, so that no batch or country appears in more than one split. We select the trial with the highest mean validation macro-$F_1$ across the five folds, then report the mean and standard deviation of the five corresponding test scores. Full deep learning hyperparameters are outlined in Table~\ref{tab:grid-search-space}. 

\paragraph{SVM and Elastic Net Baselines.}
To ensure a fair comparison, the baselines use the same folds, train-only corrections, imputation, and node-selected features as the GNN pipeline. They never use the graph. For every dataset, fold, node-sampling ratio, and node-selection method we fit two models on those features: \texttt{StandardScaler} followed by either a sigmoid-calibrated linear SVM (\texttt{CalibratedClassifierCV} wrapping \texttt{LinearSVC}) or elastic-net logistic regression (\texttt{LogisticRegression} with \texttt{penalty=elasticnet}, \texttt{solver=saga}). Both use balanced class weights. Hyperparameters are chosen with \texttt{GridSearchCV} on that fold's training/validation split, scoring by macro-$F_1$ (SVM: $C \in \{0.001, 0.01, 0.1\}$; elastic net: $C \in \{0.001, 0.01, 0.1, 1.0\}$ and $\texttt{l1\_ratio} \in \{0.0, 0.1, 0.3, 0.5, 0.7\}$). The selected pipeline is then refit on the training split only and evaluated on the held-out validation and test sets of the same fold.

\begin{table}[!h]
\centering
\caption{Hyperparameter search space for the multi-dataset Optuna sweep. Each ablation cell is one dataset, model, graph construction, node sampling setting, and readout configuration, and these axes are enumerated fully. Within a cell, 7 trials are run: the first 3 are drawn at random and the other 4 are proposed by TPE. A trial is scored by the mean validation $F_{macro}$ across five folds. Each fold trains for at most one hour and is then evaluated on its best checkpoint. In \texttt{omics\_readout}, every graph backbone uses a readout MLP with fixed widths $[16, 8, 4]$, and readout dropout is set equal to the sampled backbone dropout. MLP is run only with \texttt{no\_readout}.}
\label{tab:grid-search-space}
\renewcommand{\arraystretch}{1.2}
\setlength{\tabcolsep}{6pt}
\begin{tabular}{@{}l l p{0.52\linewidth}@{}}
\toprule
\textbf{Category} & \textbf{Hyperparameter} & \textbf{Candidates} \\
\midrule
\multirow{5}{*}{\makecell[l]{Data \&\\Graph}}
& Datasets & Heritage, Tuberculosis, Parkinson's, AddNeuroMed, Smoking, BRCA \\
& Edge construction & Co-Expression (wgcna, 10\% target connectivity), PPI (string, score $\geq$ 0.4) \\
& Node sampling ratio & 1.0, 0.8, 0.5, 0.3 \\
& Node selection method & variance, random, correlation, distance correlation \\
& Cross-validation & 5-fold, training seed 42 \\
\midrule
\multirow{3}{*}{\makecell[l]{Optimization \&\\Regularization}}
& Learning rate & log-uniform $[10^{-5}, 10^{-2}]$ \\
& Weight decay & $0$, $10^{-6}$, $10^{-5}$, $10^{-4}$ \\
& Backbone dropout & $0.0, 0.1, \ldots, 0.5$ \\
\midrule
\multirow{2}{*}{Experiment}
& Readout & no\_readout, omics\_readout \\
& Readout dropout & matched to backbone dropout (omics\_readout only) \\
\midrule
\multicolumn{3}{@{}l}{\textbf{Model-Specific Hyperparameters}} \\
\midrule
\multirow{3}{*}{GCN, GIN}
& Feature encoder out channels & 16, 32, 64, 128 \\
& \# layers & 2, 3, 4 \\
\cmidrule(lr){2-3}
\multirow{4}{*}{GATv2}
& Feature encoder out channels & 16, 32, 64, 128 \\
& \# layers & 2, 3, 4 \\
& Attention heads & 2, 4, 8 \\
\cmidrule(lr){2-3}
\multirow{4}{*}{MLA-GNN}
& Hidden channels & [16, 32], [32, 64], [64, 128] \\
& Attention heads & [4, 4], [8, 8] \\
& Layer normalization & true, false \\
\cmidrule(lr){2-3}
\multirow{3}{*}{GraphSAGE}
& Feature encoder out channels & 16, 32, 64, 128 \\
& \# layers & 2, 3, 4 \\
\cmidrule(lr){2-3}
\multirow{3}{*}{ChebNet}
& Feature encoder out channels & 16, 32, 64, 128 \\
& \# layers & 2, 3, 4 \\
\cmidrule(lr){2-3}
\multirow{4}{*}{SAGN}
& Hidden channels & 16, 32, 64, 128 \\
& \# layers & 2, 3, 4 \\
& Attention heads & 2, 4, 8 \\
\cmidrule(lr){2-3}
\multirow{4}{*}{GPS}
& Feature encoder out channels & 16, 32, 64, 128 \\
& \# layers & 2, 3, 4 \\
& Positional encodings & [LapPE], [RWSE] \\
\cmidrule(lr){2-3}
\multirow{3}{*}{MLP}
& Hidden channels & [16, 32, 8], [32, 64, 16], [64, 128, 32] \\
& Normalization & null, batch \\
\bottomrule
\end{tabular}
\end{table}

\subsection{Evaluation Metrics} \label{sec:details_metrics}

In evaluating model performance in \texttt{OmicsGBench}, we consider metrics that provide a comprehensive view of predictive accuracy, performance across classes, and computational efficiency.

We rely on a variety of metrics to capture both overall accuracy and class-specific performance:
\begin{itemize}[leftmargin=*,nosep,topsep=1pt,itemsep=1pt,parsep=0pt,partopsep=0pt]
    \item \textbf{Accuracy:} The overall proportion of correct predictions across all classes.
    \item \textbf{Macro F1 Score:} The unweighted average of F1 scores computed independently for each class, treating all classes equally regardless of size.
    \item \textbf{Weighted F1 Score:} The average of F1 scores weighted by the number of true instances per class, accounting for class imbalance.
    \item \textbf{Macro AUROC:} The average Area Under the Receiver Operating Characteristic curve computed in a one-vs-rest manner for each class.
    % \item \textbf{Confusion Matrix:} A tabular summary showing the counts of correct and incorrect predictions broken down by class, useful for detailed diagnostic analysis.
\end{itemize}

Finally, we report measures of efficiency and scalability:
\begin{itemize}[leftmargin=*,nosep,topsep=1pt,itemsep=1pt,parsep=0pt,partopsep=0pt]
    \item \textbf{Number of parameters of the model:} Total trainable parameters of the model, providing a measure of its size and capacity.
    \item \textbf{Memory:} Peak GPU memory usage during inference, reflecting the model’s scalability and hardware efficiency.
    \item \textbf{Training time per epoch:} Wall-clock time per epoch on the training set. 
\end{itemize}

Together, these metrics capture not only predictive accuracy but also resource requirements and efficiency, enabling a balanced evaluation of model performance.

\subsection{Best Model Results, Timing and Performance Details}
\label{sec:best_results}

Tables~\ref{tab:best-configs-with-gpu} and \ref{tab:best-configs-with-gpu-2} report all test metrics and computational costs for the best configuration of each model and dataset shown in Figure~\ref{fig:best_overall}.

% Selection: rows of best_per_model_dataset.csv

% Fold runs per server: placeholder 230, unknown 270.
% Selection: validation F1, best trial per model and dataset over all studies
% PLACEHOLDER resources (params, GPU memory, time per epoch) borrowed from the earlier configuration of the same model and dataset until rerun: addneuromed/chebnet, addneuromed/gatv2, addneuromed/gatv4, addneuromed/gcn, addneuromed/gin, addneuromed/gps, addneuromed/mlp, addneuromed/sage, brca/gatv2, brca/gatv4, brca/gcn, brca/gin, brca/gps, brca/sagn, motrpac/chebnet, motrpac/gatv2, motrpac/gatv4, motrpac/gcn, motrpac/gin, motrpac/gps, motrpac/mlp, motrpac/sage, motrpac/sagn, parkinsons/chebnet, parkinsons/gatv2, parkinsons/gatv4, parkinsons/gcn, parkinsons/gin, parkinsons/gps, parkinsons/mlp, parkinsons/sagn, smoking/gatv2, smoking/gatv4, smoking/gcn, smoking/gps, smoking/mlp, smoking/sage, smoking/sagn, tuberculosis/chebnet, tuberculosis/gatv2, tuberculosis/gatv4, tuberculosis/gcn, tuberculosis/gps, tuberculosis/mlp, tuberculosis/sage, tuberculosis/sagn.
\begin{table*}[!h]
\centering
\footnotesize
\setlength{\tabcolsep}{1.2pt}
\caption{Best configuration per model and dataset, selected by validation $F_{\mathrm{macro}}$. Performance metrics are test scores (mean $\pm$ std over the five folds). Best value per dataset and column is bold; other entries within one std of that best are shaded blue. Parameters and GPU memory (peak allocated) are means over folds. Time per epoch is the mean training-epoch time of each fold (first ten epochs excluded), reported as mean $\pm$ std over folds. SVM and Elastic Net do not have a comparable trained parameter count and do not require GPU. \textbf{Three out of six datasets shown} (Heritage, Tuberculosis and Parkinson's).}
\label{tab:best-configs-with-gpu}
\begin{tabular*}{\textwidth}{@{\extracolsep{\fill}}l c c c c c c c c@{}}
\toprule
 & \makecell{$F_{\mathrm{macro}}$ \\ $(\uparrow)$} & \makecell{$F_{\mathrm{weighted}}$ \\ $(\uparrow)$} & \makecell{Accuracy \\ $(\uparrow)$} & \makecell{AUROC \\ $(\uparrow)$} & \makecell{$\#$ Params. \\ $(\downarrow)$} & \makecell{GPU \\ memory \\ (GB) \\ $(\downarrow)$} & \makecell{Time / \\ epoch \\ (s) \\ $(\downarrow)$} \\
\midrule
\multicolumn{8}{l}{\textbf{Heritage}}\\
\hspace{0.8em}SVM & 0.510 $\pm$ 0.048 & 0.528 $\pm$ 0.047 & \cellcolor{blue!12}0.543 $\pm$ 0.049 & 0.547 $\pm$ 0.067 & -- & -- & -- \\
\hspace{0.8em}Elastic Net & \cellcolor{blue!12}0.539 $\pm$ 0.060 & \cellcolor{blue!12}0.547 $\pm$ 0.061 & \cellcolor{blue!12}0.546 $\pm$ 0.061 & 0.538 $\pm$ 0.046 & -- & -- & -- \\
\cmidrule(lr){1-8}
\hspace{0.8em}MLP & 0.512 $\pm$ 0.057 & 0.517 $\pm$ 0.057 & 0.515 $\pm$ 0.058 & 0.547 $\pm$ 0.054 & 19K & 0.30 & \textbf{0.34 $\pm$ 0.10} \\
\hspace{0.8em}GIN & 0.505 $\pm$ 0.056 & 0.520 $\pm$ 0.054 & \cellcolor{blue!12}0.535 $\pm$ 0.061 & 0.524 $\pm$ 0.065 & \textbf{2K} & \textbf{0.03} & 0.72 $\pm$ 0.03 \\
\hspace{0.8em}GCN & \textbf{0.557 $\pm$ 0.036} & \textbf{0.566 $\pm$ 0.037} & \textbf{0.567 $\pm$ 0.038} & \textbf{0.580 $\pm$ 0.021} & 649K & 0.32 & 1.83 $\pm$ 0.12 \\
\hspace{0.8em}GATv2 & 0.502 $\pm$ 0.047 & 0.516 $\pm$ 0.049 & 0.523 $\pm$ 0.052 & 0.536 $\pm$ 0.065 & 259K & 0.36 & 1.12 $\pm$ 0.06 \\
\hspace{0.8em}SAGE & \cellcolor{blue!12}0.527 $\pm$ 0.059 & \cellcolor{blue!12}0.535 $\pm$ 0.055 & \cellcolor{blue!12}0.543 $\pm$ 0.049 & 0.540 $\pm$ 0.063 & 666K & 0.10 & 2.87 $\pm$ 0.28 \\
\hspace{0.8em}GPS & \cellcolor{blue!12}0.524 $\pm$ 0.024 & \cellcolor{blue!12}0.531 $\pm$ 0.023 & \cellcolor{blue!12}0.531 $\pm$ 0.023 & 0.548 $\pm$ 0.058 & 348K & 0.07 & 1.59 $\pm$ 0.10 \\
\hspace{0.8em}SAGN & \cellcolor{blue!12}0.527 $\pm$ 0.027 & \cellcolor{blue!12}0.535 $\pm$ 0.024 & \cellcolor{blue!12}0.535 $\pm$ 0.022 & 0.525 $\pm$ 0.035 & 1.2M & 0.53 & 1.60 $\pm$ 0.08 \\
\hspace{0.8em}ChebNet & 0.512 $\pm$ 0.031 & 0.521 $\pm$ 0.032 & 0.526 $\pm$ 0.036 & 0.533 $\pm$ 0.047 & 1.7M & 0.31 & 1.72 $\pm$ 0.02 \\
\hspace{0.8em}MLA-GNN & 0.519 $\pm$ 0.044 & \cellcolor{blue!12}0.531 $\pm$ 0.039 & \cellcolor{blue!12}0.535 $\pm$ 0.033 & 0.531 $\pm$ 0.058 & 291K & 2.86 & 2.36 $\pm$ 0.17 \\
\midrule
\multicolumn{8}{l}{\textbf{Tuberculosis}}\\
\hspace{0.8em}SVM & \cellcolor{blue!12}0.651 $\pm$ 0.068 & \cellcolor{blue!12}0.672 $\pm$ 0.061 & \cellcolor{blue!12}0.668 $\pm$ 0.070 & \cellcolor{blue!12}0.765 $\pm$ 0.059 & -- & -- & -- \\
\hspace{0.8em}Elastic Net & \cellcolor{blue!12}0.694 $\pm$ 0.044 & \cellcolor{blue!12}0.717 $\pm$ 0.048 & \cellcolor{blue!12}0.713 $\pm$ 0.043 & \textbf{0.793 $\pm$ 0.072} & -- & -- & -- \\
\cmidrule(lr){1-8}
\hspace{0.8em}MLP & \cellcolor{blue!12}0.642 $\pm$ 0.137 & \cellcolor{blue!12}0.660 $\pm$ 0.150 & \cellcolor{blue!12}0.667 $\pm$ 0.137 & \cellcolor{blue!12}0.729 $\pm$ 0.119 & \textbf{14K} & 0.04 & \textbf{0.24 $\pm$ 0.09} \\
\hspace{0.8em}GIN & \cellcolor{blue!12}0.654 $\pm$ 0.054 & \cellcolor{blue!12}0.677 $\pm$ 0.051 & \cellcolor{blue!12}0.670 $\pm$ 0.053 & \cellcolor{blue!12}0.731 $\pm$ 0.053 & 110K & \textbf{0.02} & 0.43 $\pm$ 0.04 \\
\hspace{0.8em}GCN & \cellcolor{blue!12}0.675 $\pm$ 0.030 & \cellcolor{blue!12}0.705 $\pm$ 0.028 & \cellcolor{blue!12}0.709 $\pm$ 0.040 & \cellcolor{blue!12}0.731 $\pm$ 0.048 & 136K & 0.08 & 0.60 $\pm$ 0.04 \\
\hspace{0.8em}GATv2 & \cellcolor{blue!12}0.677 $\pm$ 0.065 & \cellcolor{blue!12}0.702 $\pm$ 0.078 & \cellcolor{blue!12}0.703 $\pm$ 0.078 & \cellcolor{blue!12}0.741 $\pm$ 0.053 & 275K & 0.08 & 0.52 $\pm$ 0.07 \\
\hspace{0.8em}SAGE & \textbf{0.698 $\pm$ 0.072} & \textbf{0.724 $\pm$ 0.080} & \textbf{0.724 $\pm$ 0.078} & \cellcolor{blue!12}0.756 $\pm$ 0.076 & 221K & 0.03 & 0.36 $\pm$ 0.06 \\
\hspace{0.8em}GPS & \cellcolor{blue!12}0.635 $\pm$ 0.064 & \cellcolor{blue!12}0.668 $\pm$ 0.071 & \cellcolor{blue!12}0.670 $\pm$ 0.073 & 0.702 $\pm$ 0.055 & 1.8M & 0.22 & 2.74 $\pm$ 0.38 \\
\hspace{0.8em}SAGN & \cellcolor{blue!12}0.695 $\pm$ 0.051 & \cellcolor{blue!12}0.723 $\pm$ 0.046 & \cellcolor{blue!12}0.724 $\pm$ 0.052 & \cellcolor{blue!12}0.734 $\pm$ 0.057 & 623K & 0.29 & 1.00 $\pm$ 0.04 \\
\hspace{0.8em}ChebNet & \cellcolor{blue!12}0.659 $\pm$ 0.080 & \cellcolor{blue!12}0.679 $\pm$ 0.086 & \cellcolor{blue!12}0.678 $\pm$ 0.086 & \cellcolor{blue!12}0.738 $\pm$ 0.062 & 218K & 0.05 & 0.64 $\pm$ 0.12 \\
\hspace{0.8em}MLA-GNN & \cellcolor{blue!12}0.671 $\pm$ 0.039 & \cellcolor{blue!12}0.699 $\pm$ 0.027 & \cellcolor{blue!12}0.696 $\pm$ 0.029 & 0.720 $\pm$ 0.019 & 267K & 0.40 & 0.53 $\pm$ 0.07 \\
\midrule
\multicolumn{8}{l}{\textbf{Parkinson's}}\\
\hspace{0.8em}SVM & \cellcolor{blue!12}0.539 $\pm$ 0.087 & \cellcolor{blue!12}0.589 $\pm$ 0.079 & \textbf{0.637 $\pm$ 0.048} & \cellcolor{blue!12}0.620 $\pm$ 0.138 & -- & -- & -- \\
\hspace{0.8em}Elastic Net & \textbf{0.590 $\pm$ 0.059} & \textbf{0.616 $\pm$ 0.063} & \cellcolor{blue!12}0.623 $\pm$ 0.052 & \textbf{0.633 $\pm$ 0.065} & -- & -- & -- \\
\cmidrule(lr){1-8}
\hspace{0.8em}MLP & \cellcolor{blue!12}0.553 $\pm$ 0.071 & \cellcolor{blue!12}0.579 $\pm$ 0.071 & 0.587 $\pm$ 0.058 & \cellcolor{blue!12}0.600 $\pm$ 0.075 & \textbf{54K} & 0.05 & \textbf{0.31 $\pm$ 0.07} \\
\hspace{0.8em}GIN & \cellcolor{blue!12}0.531 $\pm$ 0.062 & \cellcolor{blue!12}0.559 $\pm$ 0.055 & 0.562 $\pm$ 0.051 & 0.536 $\pm$ 0.067 & 551K & 0.08 & 1.60 $\pm$ 0.15 \\
\hspace{0.8em}GCN & 0.526 $\pm$ 0.110 & 0.551 $\pm$ 0.108 & 0.575 $\pm$ 0.059 & \cellcolor{blue!12}0.580 $\pm$ 0.083 & 260K & \textbf{0.03} & \cellcolor{blue!12}0.37 $\pm$ 0.06 \\
\hspace{0.8em}GATv2 & \cellcolor{blue!12}0.579 $\pm$ 0.063 & \cellcolor{blue!12}0.603 $\pm$ 0.046 & \cellcolor{blue!12}0.606 $\pm$ 0.047 & \cellcolor{blue!12}0.606 $\pm$ 0.098 & 262K & 0.30 & 0.80 $\pm$ 0.05 \\
\hspace{0.8em}SAGE & \cellcolor{blue!12}0.566 $\pm$ 0.090 & \cellcolor{blue!12}0.594 $\pm$ 0.081 & \cellcolor{blue!12}0.602 $\pm$ 0.063 & \cellcolor{blue!12}0.596 $\pm$ 0.115 & 1.2M & 0.08 & 0.48 $\pm$ 0.05 \\
\hspace{0.8em}GPS & 0.528 $\pm$ 0.067 & \cellcolor{blue!12}0.581 $\pm$ 0.038 & \cellcolor{blue!12}0.623 $\pm$ 0.037 & \cellcolor{blue!12}0.596 $\pm$ 0.046 & 686K & 0.12 & 1.67 $\pm$ 0.16 \\
\hspace{0.8em}SAGN & 0.530 $\pm$ 0.087 & 0.551 $\pm$ 0.081 & 0.564 $\pm$ 0.060 & \cellcolor{blue!12}0.570 $\pm$ 0.113 & 358K & 0.33 & 0.88 $\pm$ 0.08 \\
\hspace{0.8em}ChebNet & \cellcolor{blue!12}0.543 $\pm$ 0.057 & \cellcolor{blue!12}0.577 $\pm$ 0.030 & \cellcolor{blue!12}0.598 $\pm$ 0.021 & \cellcolor{blue!12}0.612 $\pm$ 0.069 & 262K & 0.04 & 0.46 $\pm$ 0.10 \\
\hspace{0.8em}MLA-GNN & \cellcolor{blue!12}0.538 $\pm$ 0.088 & \cellcolor{blue!12}0.563 $\pm$ 0.078 & 0.583 $\pm$ 0.041 & \cellcolor{blue!12}0.593 $\pm$ 0.102 & 292K & 2.99 & 2.36 $\pm$ 0.12 \\
\bottomrule
\end{tabular*}
\end{table*}

\begin{table*}[!h]
\centering
\footnotesize
\setlength{\tabcolsep}{1.2pt}
\caption{Best configuration per model and dataset, selected by validation $F_{\mathrm{macro}}$. Performance metrics are test scores (mean $\pm$ std over the five folds). Best value per dataset and column is bold; other entries within one std of that best are shaded blue. Parameters and GPU memory (peak allocated) are means over folds. Time per epoch is the mean training-epoch time of each fold (first ten epochs excluded), reported as mean $\pm$ std over folds. SVM and Elastic Net do not have a comparable trained parameter count and do not require GPU. \textbf{Three out of six datasets shown} (AddNeuroMed, Smoking and BRCA).}
\label{tab:best-configs-with-gpu-2}
\begin{tabular*}{\textwidth}{@{\extracolsep{\fill}}l c c c c c c c c@{}}
\toprule
 & \makecell{$F_{\mathrm{macro}}$ \\ $(\uparrow)$} & \makecell{$F_{\mathrm{weighted}}$ \\ $(\uparrow)$} & \makecell{Accuracy \\ $(\uparrow)$} & \makecell{AUROC \\ $(\uparrow)$} & \makecell{$\#$ Params. \\ $(\downarrow)$} & \makecell{GPU \\ memory \\ (GB) \\ $(\downarrow)$} & \makecell{Time / \\ epoch \\ (s) \\ $(\downarrow)$} \\
\midrule
\multicolumn{8}{l}{\textbf{AddNeuroMed}}\\
\hspace{0.8em}SVM & 0.505 $\pm$ 0.035 & \cellcolor{blue!12}0.523 $\pm$ 0.034 & \textbf{0.539 $\pm$ 0.032} & \textbf{0.719 $\pm$ 0.029} & -- & -- & -- \\
\hspace{0.8em}Elastic Net & \textbf{0.526 $\pm$ 0.015} & \textbf{0.537 $\pm$ 0.016} & \cellcolor{blue!12}0.537 $\pm$ 0.009 & \cellcolor{blue!12}0.717 $\pm$ 0.026 & -- & -- & -- \\
\cmidrule(lr){1-8}
\hspace{0.8em}MLP & \cellcolor{blue!12}0.516 $\pm$ 0.039 & \cellcolor{blue!12}0.527 $\pm$ 0.042 & \cellcolor{blue!12}0.532 $\pm$ 0.042 & \cellcolor{blue!12}0.716 $\pm$ 0.025 & \textbf{67K} & 0.20 & \textbf{0.30 $\pm$ 0.07} \\
\hspace{0.8em}GIN & 0.468 $\pm$ 0.048 & 0.482 $\pm$ 0.047 & 0.488 $\pm$ 0.040 & 0.658 $\pm$ 0.035 & 1.5M & 0.10 & 1.05 $\pm$ 0.15 \\
\hspace{0.8em}GCN & 0.478 $\pm$ 0.039 & 0.499 $\pm$ 0.038 & \cellcolor{blue!12}0.508 $\pm$ 0.036 & 0.683 $\pm$ 0.042 & 595K & 0.25 & 1.69 $\pm$ 0.11 \\
\hspace{0.8em}GATv2 & 0.474 $\pm$ 0.035 & 0.492 $\pm$ 0.031 & 0.502 $\pm$ 0.031 & 0.680 $\pm$ 0.017 & 612K & 0.28 & 1.10 $\pm$ 0.23 \\
\hspace{0.8em}SAGE & 0.460 $\pm$ 0.035 & 0.484 $\pm$ 0.034 & \cellcolor{blue!12}0.511 $\pm$ 0.022 & 0.687 $\pm$ 0.013 & 1.2M & 0.21 & 4.09 $\pm$ 0.05 \\
\hspace{0.8em}GPS & \cellcolor{blue!12}0.522 $\pm$ 0.063 & \cellcolor{blue!12}0.533 $\pm$ 0.064 & \cellcolor{blue!12}0.533 $\pm$ 0.066 & \cellcolor{blue!12}0.709 $\pm$ 0.035 & 1.8M & 0.13 & 1.21 $\pm$ 0.14 \\
\hspace{0.8em}SAGN & 0.510 $\pm$ 0.035 & \cellcolor{blue!12}0.525 $\pm$ 0.035 & \cellcolor{blue!12}0.530 $\pm$ 0.032 & 0.684 $\pm$ 0.042 & 5.3M & 4.22 & 2.97 $\pm$ 0.26 \\
\hspace{0.8em}ChebNet & 0.493 $\pm$ 0.050 & 0.509 $\pm$ 0.046 & \cellcolor{blue!12}0.511 $\pm$ 0.042 & 0.687 $\pm$ 0.051 & 181K & \textbf{0.05} & 0.62 $\pm$ 0.09 \\
\hspace{0.8em}MLA-GNN & 0.469 $\pm$ 0.047 & 0.484 $\pm$ 0.040 & 0.492 $\pm$ 0.030 & 0.658 $\pm$ 0.038 & 301K & 5.21 & 6.13 $\pm$ 0.37 \\
\midrule
\multicolumn{8}{l}{\textbf{Smoking}}\\
\hspace{0.8em}SVM & 0.632 $\pm$ 0.052 & \cellcolor{blue!12}0.656 $\pm$ 0.047 & \cellcolor{blue!12}0.664 $\pm$ 0.045 & \cellcolor{blue!12}0.723 $\pm$ 0.045 & -- & -- & -- \\
\hspace{0.8em}Elastic Net & 0.634 $\pm$ 0.028 & 0.649 $\pm$ 0.027 & 0.647 $\pm$ 0.028 & \cellcolor{blue!12}0.712 $\pm$ 0.043 & -- & -- & -- \\
\cmidrule(lr){1-8}
\hspace{0.8em}MLP & 0.636 $\pm$ 0.035 & 0.655 $\pm$ 0.031 & \cellcolor{blue!12}0.655 $\pm$ 0.030 & 0.680 $\pm$ 0.034 & \textbf{14K} & \textbf{0.03} & \textbf{0.15 $\pm$ 0.04} \\
\hspace{0.8em}GIN & 0.638 $\pm$ 0.031 & 0.649 $\pm$ 0.028 & 0.644 $\pm$ 0.029 & 0.659 $\pm$ 0.032 & 1.5M & 0.10 & 0.54 $\pm$ 0.03 \\
\hspace{0.8em}GCN & 0.586 $\pm$ 0.040 & 0.614 $\pm$ 0.037 & 0.623 $\pm$ 0.034 & 0.663 $\pm$ 0.070 & 1.2M & 0.11 & 0.71 $\pm$ 0.04 \\
\hspace{0.8em}GATv2 & 0.626 $\pm$ 0.055 & 0.650 $\pm$ 0.049 & \cellcolor{blue!12}0.655 $\pm$ 0.045 & 0.701 $\pm$ 0.029 & 1.5M & 0.17 & 0.55 $\pm$ 0.06 \\
\hspace{0.8em}SAGE & 0.617 $\pm$ 0.044 & 0.639 $\pm$ 0.038 & 0.642 $\pm$ 0.033 & \cellcolor{blue!12}0.719 $\pm$ 0.049 & 1.0M & 0.07 & 0.37 $\pm$ 0.02 \\
\hspace{0.8em}GPS & \cellcolor{blue!12}0.640 $\pm$ 0.038 & \cellcolor{blue!12}0.660 $\pm$ 0.035 & \cellcolor{blue!12}0.662 $\pm$ 0.035 & 0.679 $\pm$ 0.055 & 628K & 0.09 & 0.81 $\pm$ 0.03 \\
\hspace{0.8em}SAGN & \cellcolor{blue!12}0.656 $\pm$ 0.034 & \cellcolor{blue!12}0.674 $\pm$ 0.032 & \cellcolor{blue!12}0.674 $\pm$ 0.032 & \cellcolor{blue!12}0.709 $\pm$ 0.047 & 1.4M & 0.75 & 0.88 $\pm$ 0.03 \\
\hspace{0.8em}ChebNet & \textbf{0.673 $\pm$ 0.035} & \textbf{0.691 $\pm$ 0.035} & \textbf{0.692 $\pm$ 0.037} & \textbf{0.740 $\pm$ 0.035} & 599K & 0.44 & 1.69 $\pm$ 0.10 \\
\hspace{0.8em}MLA-GNN & 0.610 $\pm$ 0.073 & 0.629 $\pm$ 0.066 & 0.631 $\pm$ 0.064 & 0.667 $\pm$ 0.108 & 268K & 15.64 & 10.14 $\pm$ 0.55 \\
\midrule
\multicolumn{8}{l}{\textbf{BRCA}}\\
\hspace{0.8em}SVM & 0.756 $\pm$ 0.038 & 0.799 $\pm$ 0.025 & \cellcolor{blue!12}0.812 $\pm$ 0.022 & 0.943 $\pm$ 0.010 & -- & -- & -- \\
\hspace{0.8em}Elastic Net & \textbf{0.812 $\pm$ 0.026} & \textbf{0.827 $\pm$ 0.024} & \textbf{0.827 $\pm$ 0.024} & \textbf{0.952 $\pm$ 0.009} & -- & -- & -- \\
\cmidrule(lr){1-8}
\hspace{0.8em}MLP & \cellcolor{blue!12}0.800 $\pm$ 0.031 & \cellcolor{blue!12}0.816 $\pm$ 0.017 & \cellcolor{blue!12}0.816 $\pm$ 0.018 & 0.931 $\pm$ 0.010 & 28K & 0.07 & \textbf{0.39 $\pm$ 0.02} \\
\hspace{0.8em}GIN & 0.676 $\pm$ 0.091 & 0.740 $\pm$ 0.081 & 0.752 $\pm$ 0.064 & 0.901 $\pm$ 0.016 & \textbf{22K} & 0.06 & 1.10 $\pm$ 0.04 \\
\hspace{0.8em}GCN & 0.688 $\pm$ 0.034 & 0.729 $\pm$ 0.037 & 0.725 $\pm$ 0.046 & 0.898 $\pm$ 0.022 & 271K & 0.22 & 1.74 $\pm$ 0.29 \\
\hspace{0.8em}GATv2 & 0.716 $\pm$ 0.021 & 0.769 $\pm$ 0.025 & 0.770 $\pm$ 0.025 & 0.920 $\pm$ 0.011 & 137K & 0.14 & 0.71 $\pm$ 0.06 \\
\hspace{0.8em}SAGE & 0.773 $\pm$ 0.048 & 0.802 $\pm$ 0.028 & \cellcolor{blue!12}0.803 $\pm$ 0.028 & 0.941 $\pm$ 0.016 & 272K & \textbf{0.03} & 0.58 $\pm$ 0.04 \\
\hspace{0.8em}GPS & 0.780 $\pm$ 0.039 & \cellcolor{blue!12}0.812 $\pm$ 0.030 & \cellcolor{blue!12}0.809 $\pm$ 0.031 & 0.901 $\pm$ 0.017 & 279K & 0.08 & 2.74 $\pm$ 0.09 \\
\hspace{0.8em}SAGN & 0.737 $\pm$ 0.044 & 0.768 $\pm$ 0.026 & 0.789 $\pm$ 0.020 & 0.925 $\pm$ 0.014 & 2.8M & 1.53 & 1.19 $\pm$ 0.05 \\
\hspace{0.8em}ChebNet & \cellcolor{blue!12}0.801 $\pm$ 0.025 & \cellcolor{blue!12}0.824 $\pm$ 0.008 & \cellcolor{blue!12}0.825 $\pm$ 0.008 & 0.941 $\pm$ 0.015 & 348K & 0.16 & 0.97 $\pm$ 0.09 \\
\hspace{0.8em}MLA-GNN & \cellcolor{blue!12}0.795 $\pm$ 0.037 & 0.803 $\pm$ 0.025 & 0.797 $\pm$ 0.025 & 0.941 $\pm$ 0.013 & 267K & 2.75 & 2.87 $\pm$ 0.13 \\
\bottomrule
\end{tabular*}
\end{table*}

\clearpage
\section{Effect of Edge Construction Methods and Model Readout}
\label{sec:readout_and_edge_construction}

This appendix reports per-backbone results for the two ablations that isolate how GNNs use the graph: the readout (Section~\ref{app:readout_ablation}, RQ2) and the edge source (Section~\ref{app:edge_con}, RQ4).

\subsection{Readout Ablation} \label{app:readout_ablation}

\begin{figure}[H]
    \centering
    \includegraphics[width=\linewidth]{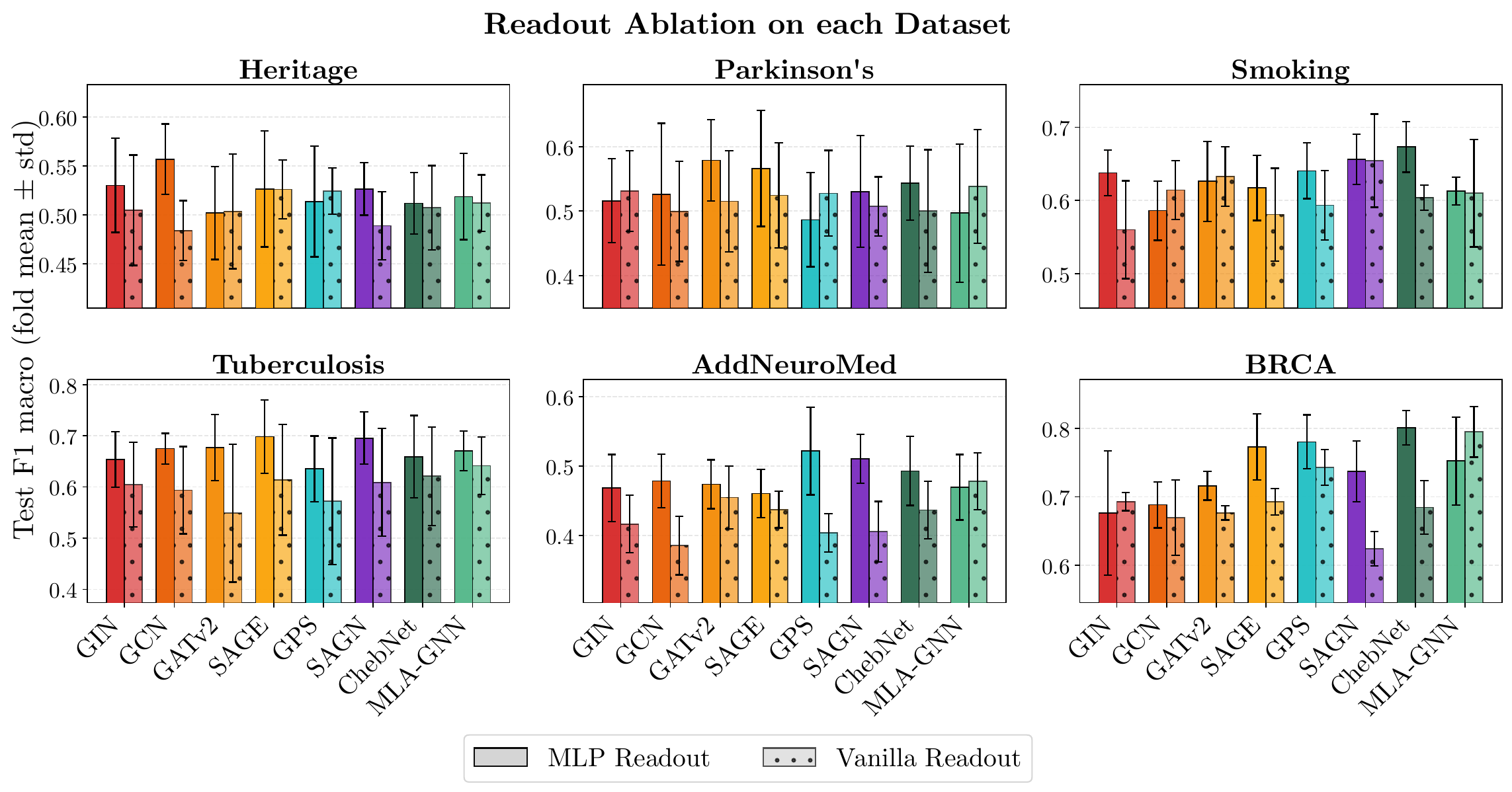}
    \caption{Test $F_{\mathrm{macro}}$ by readout type for each GNN backbone, sweeping over node selection method, sampling ratio, and model hyperparameters. For each model and readout, the configuration with the highest mean validation $F_{\mathrm{macro}}$ across folds is reported (mean $\pm$ std). Replacing the MLP readout with mean pooling reduces performance for most backbones on Tuberculosis, AddNeuroMed, and BRCA, while on Heritage, Parkinson's, and Smoking the effect is smaller and inconsistent in direction. MLA-GNN, which flattens node features before the readout, is largely unaffected by mean pooling on the three datasets where other backbones degrade most.}
    \label{fig:readout_effect}
\end{figure}

Figure~\ref{fig:readout_effect} and Tables~\ref{tab:motrpac-readout-vs-noreadout}--\ref{tab:brca-readout-vs-noreadout} report per-backbone results for the readout ablation discussed in RQ2.

\subsection{Edge Construction Ablation} \label{app:edge_con}

\begin{figure}[H]
    \centering
    \includegraphics[width=\linewidth]{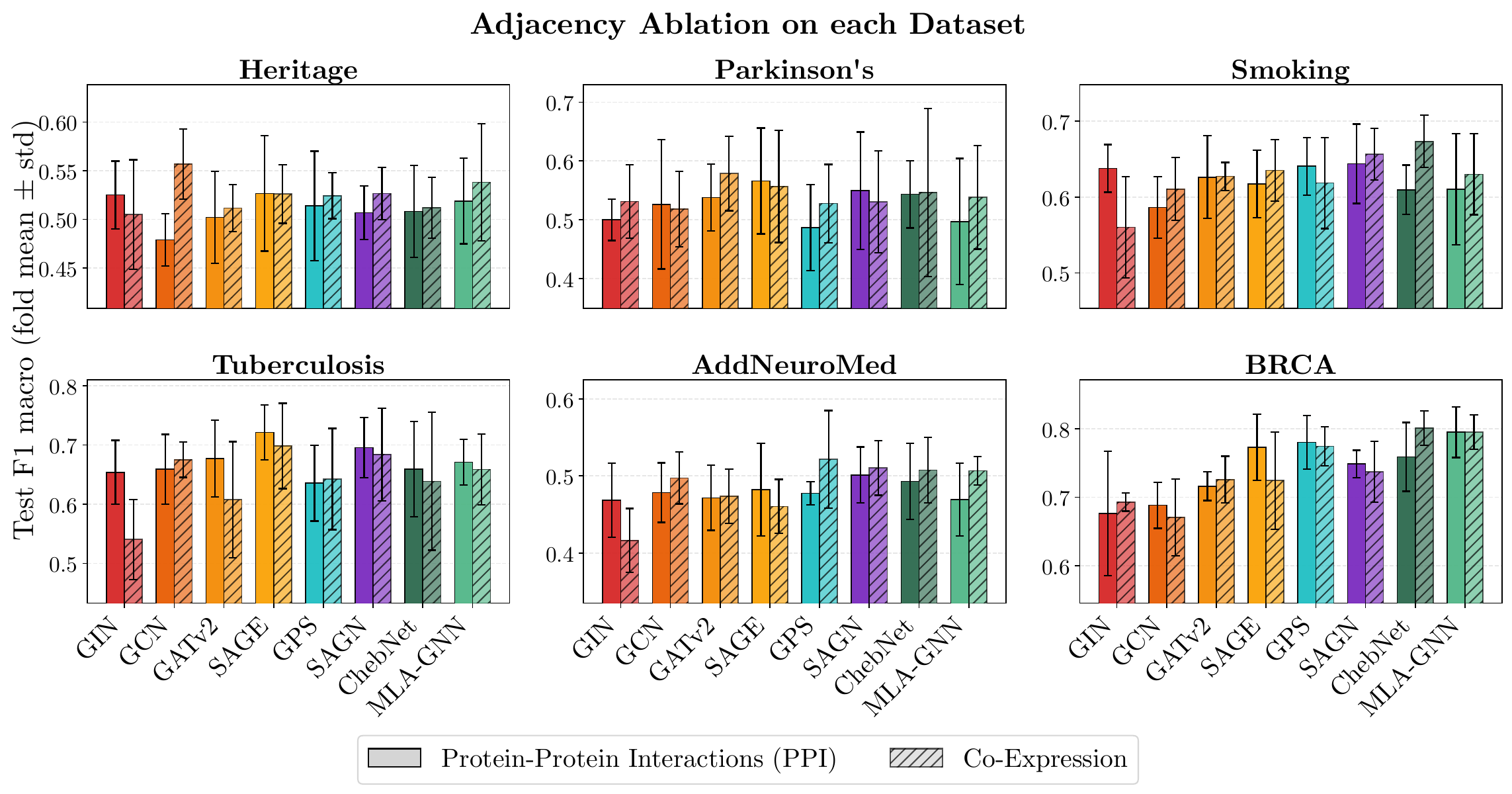}
    \caption{Test $F_{\mathrm{macro}}$ by edge construction method for each GNN backbone, sweeping over node selection method, sampling ratio, and model hyperparameters. For each model and edge source, the configuration with the highest mean validation $F_{\mathrm{macro}}$ across folds is reported (mean $\pm$ std). Most differences fall within fold-to-fold variance, and the better edge source varies across backbones within every dataset. Tuberculosis, the only dataset with a pathogen rather than human PPI network, is the one dataset where most backbones perform better with PPI edges.}
    \label{fig:edge_construct_effect}
\end{figure}

Figure~\ref{fig:edge_construct_effect} and Tables~\ref{tab:motrpac-adjacency-ppi-vs-coexpression}--\ref{tab:brca-adjacency-ppi-vs-coexpression} report per-backbone results for the edge construction ablation discussed in RQ4, comparing STRING-derived PPI networks (biological prior) with co-expression graphs estimated from each training split (data-driven).

\clearpage
\section{Ablation Tables}
\label{sec:ablation_tables}

Tables~\ref{tab:motrpac-readout-vs-noreadout}--\ref{tab:brca-readout-vs-noreadout} report all four test metrics for the readout ablation summarized in Figure~\ref{fig:readout_effect}, and Tables~\ref{tab:motrpac-adjacency-ppi-vs-coexpression}--\ref{tab:brca-adjacency-ppi-vs-coexpression} report them for the edge construction ablation summarized in Figure~\ref{fig:edge_construct_effect}. For each model and setting, the configuration with the highest mean validation $F_{\mathrm{macro}}$ is reported.

\begin{table*}[!h]
\centering
\small
\setlength{\tabcolsep}{5pt}
\renewcommand{\arraystretch}{1.15}
\caption{Readout ablation on \textit{Heritage}. For each (model, readout), the highest val $F_{macro}$ configuration is selected; test metrics (mean $\pm$ std) are reported.}
\begin{tabularx}{\textwidth}{>{\raggedright\arraybackslash}p{1cm} l l Y Y Y Y}
\toprule
Dataset & Model & Readout & $F_{macro}$ & $F_{weighted}$ & Accuracy & AUROC \\
\midrule
\multicolumn{7}{@{}l}{\textbf{Heritage}}\\[-0.6ex]
 & MLP & OmicsReadOut & --- & --- & --- & --- \\
 &  & NoReadOut & \textbf{0.512 $\pm$ 0.057} & \textbf{0.517 $\pm$ 0.057} & \textbf{0.515 $\pm$ 0.058} & \textbf{0.547 $\pm$ 0.054} \\
\midrule
 & GIN & OmicsReadOut & \textbf{0.530 $\pm$ 0.048} & \textbf{0.546 $\pm$ 0.048} & \textbf{0.563 $\pm$ 0.052} & \textbf{0.527 $\pm$ 0.049} \\
 &  & NoReadOut & \withinstd 0.505 $\pm$ 0.056 & \withinstd 0.520 $\pm$ 0.054 & \withinstd 0.535 $\pm$ 0.061 & \withinstd 0.524 $\pm$ 0.065 \\
\midrule
 & GCN & OmicsReadOut & \textbf{0.557 $\pm$ 0.036} & \textbf{0.566 $\pm$ 0.037} & \textbf{0.567 $\pm$ 0.038} & \textbf{0.580 $\pm$ 0.021} \\
 &  & NoReadOut & 0.484 $\pm$ 0.031 & 0.499 $\pm$ 0.030 & 0.508 $\pm$ 0.032 & 0.500 $\pm$ 0.028 \\
\midrule
 & GATv2 & OmicsReadOut & \withinstd 0.502 $\pm$ 0.047 & \textbf{0.516 $\pm$ 0.049} & \textbf{0.523 $\pm$ 0.052} & \textbf{0.536 $\pm$ 0.065} \\
 &  & NoReadOut & \textbf{0.503 $\pm$ 0.058} & \withinstd 0.512 $\pm$ 0.061 & \withinstd 0.522 $\pm$ 0.055 & \withinstd 0.524 $\pm$ 0.059 \\
\midrule
 & SAGE & OmicsReadOut & \textbf{0.527 $\pm$ 0.059} & \withinstd 0.535 $\pm$ 0.055 & \withinstd 0.543 $\pm$ 0.049 & \textbf{0.540 $\pm$ 0.063} \\
 &  & NoReadOut & \withinstd 0.526 $\pm$ 0.030 & \textbf{0.538 $\pm$ 0.026} & \textbf{0.544 $\pm$ 0.022} & \withinstd 0.530 $\pm$ 0.040 \\
\midrule
 & GPS & OmicsReadOut & \withinstd 0.514 $\pm$ 0.056 & \withinstd 0.526 $\pm$ 0.053 & \textbf{0.531 $\pm$ 0.050} & \withinstd 0.530 $\pm$ 0.058 \\
 &  & NoReadOut & \textbf{0.524 $\pm$ 0.024} & \textbf{0.531 $\pm$ 0.023} & \withinstd 0.531 $\pm$ 0.023 & \textbf{0.548 $\pm$ 0.058} \\
\midrule
 & SAGN & OmicsReadOut & \textbf{0.527 $\pm$ 0.027} & \textbf{0.535 $\pm$ 0.024} & \textbf{0.535 $\pm$ 0.022} & \textbf{0.525 $\pm$ 0.035} \\
 &  & NoReadOut & 0.489 $\pm$ 0.035 & 0.504 $\pm$ 0.038 & 0.512 $\pm$ 0.046 & 0.485 $\pm$ 0.056 \\
\midrule
 & ChebNet & OmicsReadOut & \textbf{0.512 $\pm$ 0.031} & \textbf{0.521 $\pm$ 0.032} & \textbf{0.526 $\pm$ 0.036} & \withinstd 0.533 $\pm$ 0.047 \\
 &  & NoReadOut & \withinstd 0.507 $\pm$ 0.043 & \withinstd 0.517 $\pm$ 0.047 & \withinstd 0.520 $\pm$ 0.048 & \textbf{0.534 $\pm$ 0.050} \\
\midrule
 & MLA-GNN & OmicsReadOut & \textbf{0.519 $\pm$ 0.044} & \textbf{0.531 $\pm$ 0.039} & \textbf{0.535 $\pm$ 0.033} & \textbf{0.531 $\pm$ 0.058} \\
 &  & NoReadOut & \withinstd 0.512 $\pm$ 0.029 & \withinstd 0.526 $\pm$ 0.027 & \withinstd 0.532 $\pm$ 0.026 & \withinstd 0.522 $\pm$ 0.056 \\
\bottomrule
\end{tabularx}
\label{tab:motrpac-readout-vs-noreadout}
\end{table*}

\newpage

\begin{table*}[!h]
\centering
\small
\setlength{\tabcolsep}{5pt}
\renewcommand{\arraystretch}{1.15}
\caption{Readout ablation on \textit{Tuberculosis}. For each (model, readout), the highest val $F_{macro}$ configuration is selected; test metrics (mean $\pm$ std) are reported.}
\begin{tabularx}{\textwidth}{>{\raggedright\arraybackslash}p{1cm} l l Y Y Y Y}
\toprule
Dataset & Model & Readout & $F_{macro}$ & $F_{weighted}$ & Accuracy & AUROC \\
\midrule
\multicolumn{7}{@{}l}{\textbf{Tuberculosis}}\\[-0.6ex]
 & MLP & OmicsReadOut & --- & --- & --- & --- \\
 &  & NoReadOut & \textbf{0.642 $\pm$ 0.137} & \textbf{0.660 $\pm$ 0.150} & \textbf{0.667 $\pm$ 0.137} & \textbf{0.729 $\pm$ 0.119} \\
\midrule
 & GIN & OmicsReadOut & \textbf{0.654 $\pm$ 0.054} & \textbf{0.677 $\pm$ 0.051} & \textbf{0.670 $\pm$ 0.053} & \textbf{0.731 $\pm$ 0.053} \\
 &  & NoReadOut & \withinstd 0.605 $\pm$ 0.083 & \withinstd 0.631 $\pm$ 0.100 & \withinstd 0.631 $\pm$ 0.092 & 0.651 $\pm$ 0.109 \\
\midrule
 & GCN & OmicsReadOut & \textbf{0.675 $\pm$ 0.030} & \textbf{0.705 $\pm$ 0.028} & \textbf{0.709 $\pm$ 0.040} & \textbf{0.731 $\pm$ 0.048} \\
 &  & NoReadOut & 0.594 $\pm$ 0.085 & 0.620 $\pm$ 0.101 & 0.620 $\pm$ 0.090 & 0.644 $\pm$ 0.076 \\
\midrule
 & GATv2 & OmicsReadOut & \textbf{0.677 $\pm$ 0.065} & \textbf{0.702 $\pm$ 0.078} & \textbf{0.703 $\pm$ 0.078} & \textbf{0.741 $\pm$ 0.053} \\
 &  & NoReadOut & 0.549 $\pm$ 0.135 & 0.581 $\pm$ 0.139 & 0.596 $\pm$ 0.129 & 0.626 $\pm$ 0.135 \\
\midrule
 & SAGE & OmicsReadOut & \textbf{0.698 $\pm$ 0.072} & \textbf{0.724 $\pm$ 0.080} & \textbf{0.724 $\pm$ 0.078} & \textbf{0.756 $\pm$ 0.076} \\
 &  & NoReadOut & 0.614 $\pm$ 0.108 & 0.642 $\pm$ 0.121 & \withinstd 0.650 $\pm$ 0.109 & 0.664 $\pm$ 0.109 \\
\midrule
 & GPS & OmicsReadOut & \textbf{0.635 $\pm$ 0.064} & \textbf{0.668 $\pm$ 0.071} & \textbf{0.670 $\pm$ 0.073} & \textbf{0.702 $\pm$ 0.055} \\
 &  & NoReadOut & \withinstd 0.572 $\pm$ 0.124 & 0.587 $\pm$ 0.138 & \withinstd 0.600 $\pm$ 0.109 & \withinstd 0.692 $\pm$ 0.065 \\
\midrule
 & SAGN & OmicsReadOut & \textbf{0.695 $\pm$ 0.051} & \textbf{0.723 $\pm$ 0.046} & \textbf{0.724 $\pm$ 0.052} & \textbf{0.734 $\pm$ 0.057} \\
 &  & NoReadOut & 0.609 $\pm$ 0.105 & 0.639 $\pm$ 0.109 & 0.639 $\pm$ 0.109 & \withinstd 0.684 $\pm$ 0.097 \\
\midrule
 & ChebNet & OmicsReadOut & \textbf{0.659 $\pm$ 0.080} & \textbf{0.679 $\pm$ 0.086} & \textbf{0.678 $\pm$ 0.086} & \textbf{0.738 $\pm$ 0.062} \\
 &  & NoReadOut & \withinstd 0.621 $\pm$ 0.096 & \withinstd 0.647 $\pm$ 0.102 & \withinstd 0.655 $\pm$ 0.088 & \withinstd 0.691 $\pm$ 0.050 \\
\midrule
 & MLA-GNN & OmicsReadOut & \textbf{0.671 $\pm$ 0.039} & \textbf{0.699 $\pm$ 0.027} & \textbf{0.696 $\pm$ 0.029} & \textbf{0.720 $\pm$ 0.019} \\
 &  & NoReadOut & \withinstd 0.641 $\pm$ 0.056 & 0.670 $\pm$ 0.065 & 0.666 $\pm$ 0.062 & \withinstd 0.710 $\pm$ 0.049 \\
\bottomrule
\end{tabularx}
\label{tab:tuberculosis-readout-vs-noreadout}
\end{table*}

\newpage

\begin{table*}[!h]
\centering
\small
\setlength{\tabcolsep}{5pt}
\renewcommand{\arraystretch}{1.15}
\caption{Readout ablation on \textit{Parkinson's}. For each (model, readout), the highest val $F_{macro}$ configuration is selected; test metrics (mean $\pm$ std) are reported.}
\begin{tabularx}{\textwidth}{>{\raggedright\arraybackslash}p{1cm} l l Y Y Y Y}
\toprule
Dataset & Model & Readout & $F_{macro}$ & $F_{weighted}$ & Accuracy & AUROC \\
\midrule
\multicolumn{7}{@{}l}{\textbf{Parkinson's}}\\[-0.6ex]
 & MLP & OmicsReadOut & --- & --- & --- & --- \\
 &  & NoReadOut & \textbf{0.553 $\pm$ 0.071} & \textbf{0.579 $\pm$ 0.071} & \textbf{0.587 $\pm$ 0.058} & \textbf{0.600 $\pm$ 0.075} \\
\midrule
 & GIN & OmicsReadOut & \withinstd 0.516 $\pm$ 0.065 & \textbf{0.561 $\pm$ 0.047} & \textbf{0.599 $\pm$ 0.052} & \textbf{0.559 $\pm$ 0.053} \\
 &  & NoReadOut & \textbf{0.531 $\pm$ 0.062} & \withinstd 0.559 $\pm$ 0.055 & \withinstd 0.562 $\pm$ 0.051 & \withinstd 0.536 $\pm$ 0.067 \\
\midrule
 & GCN & OmicsReadOut & \textbf{0.526 $\pm$ 0.110} & \textbf{0.551 $\pm$ 0.108} & \textbf{0.575 $\pm$ 0.059} & \textbf{0.580 $\pm$ 0.083} \\
 &  & NoReadOut & \withinstd 0.500 $\pm$ 0.077 & \withinstd 0.518 $\pm$ 0.078 & \withinstd 0.520 $\pm$ 0.083 & \withinstd 0.504 $\pm$ 0.101 \\
\midrule
 & GATv2 & OmicsReadOut & \textbf{0.579 $\pm$ 0.063} & \textbf{0.603 $\pm$ 0.046} & \textbf{0.606 $\pm$ 0.047} & \textbf{0.606 $\pm$ 0.098} \\
 &  & NoReadOut & 0.515 $\pm$ 0.079 & 0.546 $\pm$ 0.071 & \withinstd 0.561 $\pm$ 0.064 & \withinstd 0.566 $\pm$ 0.124 \\
\midrule
 & SAGE & OmicsReadOut & \textbf{0.566 $\pm$ 0.090} & \textbf{0.594 $\pm$ 0.081} & \textbf{0.602 $\pm$ 0.063} & \textbf{0.596 $\pm$ 0.115} \\
 &  & NoReadOut & \withinstd 0.524 $\pm$ 0.081 & \withinstd 0.552 $\pm$ 0.078 & \withinstd 0.560 $\pm$ 0.066 & \withinstd 0.567 $\pm$ 0.096 \\
\midrule
 & GPS & OmicsReadOut & \withinstd 0.487 $\pm$ 0.073 & 0.508 $\pm$ 0.079 & 0.531 $\pm$ 0.064 & \withinstd 0.565 $\pm$ 0.067 \\
 &  & NoReadOut & \textbf{0.528 $\pm$ 0.067} & \textbf{0.581 $\pm$ 0.038} & \textbf{0.623 $\pm$ 0.037} & \textbf{0.596 $\pm$ 0.046} \\
\midrule
 & SAGN & OmicsReadOut & \textbf{0.530 $\pm$ 0.087} & \textbf{0.551 $\pm$ 0.081} & \textbf{0.564 $\pm$ 0.060} & \textbf{0.570 $\pm$ 0.113} \\
 &  & NoReadOut & \withinstd 0.507 $\pm$ 0.046 & \withinstd 0.535 $\pm$ 0.030 & \withinstd 0.533 $\pm$ 0.038 & \withinstd 0.517 $\pm$ 0.047 \\
\midrule
 & ChebNet & OmicsReadOut & \textbf{0.543 $\pm$ 0.057} & \textbf{0.577 $\pm$ 0.030} & \textbf{0.598 $\pm$ 0.021} & \textbf{0.612 $\pm$ 0.069} \\
 &  & NoReadOut & \withinstd 0.500 $\pm$ 0.095 & 0.523 $\pm$ 0.082 & 0.536 $\pm$ 0.059 & 0.540 $\pm$ 0.114 \\
\midrule
 & MLA-GNN & OmicsReadOut & \withinstd 0.497 $\pm$ 0.107 & \withinstd 0.545 $\pm$ 0.105 & \textbf{0.589 $\pm$ 0.054} & \withinstd 0.550 $\pm$ 0.096 \\
 &  & NoReadOut & \textbf{0.538 $\pm$ 0.088} & \textbf{0.563 $\pm$ 0.078} & \withinstd 0.583 $\pm$ 0.041 & \textbf{0.593 $\pm$ 0.102} \\
\bottomrule
\end{tabularx}
\label{tab:parkinsons-readout-vs-noreadout}
\end{table*}

\newpage

\begin{table*}[!h]
\centering
\small
\setlength{\tabcolsep}{5pt}
\renewcommand{\arraystretch}{1.15}
\caption{Readout ablation on \textit{AddNeuroMed}. For each (model, readout), the highest val $F_{macro}$ configuration is selected; test metrics (mean $\pm$ std) are reported.}
\begin{tabularx}{\textwidth}{>{\raggedright\arraybackslash}p{1cm} l l Y Y Y Y}
\toprule
Dataset & Model & Readout & $F_{macro}$ & $F_{weighted}$ & Accuracy & AUROC \\
\midrule
\multicolumn{7}{@{}l}{\textbf{AddNeuroMed}}\\[-0.6ex]
 & MLP & OmicsReadOut & --- & --- & --- & --- \\
 &  & NoReadOut & \textbf{0.516 $\pm$ 0.039} & \textbf{0.527 $\pm$ 0.042} & \textbf{0.532 $\pm$ 0.042} & \textbf{0.716 $\pm$ 0.025} \\
\midrule
 & GIN & OmicsReadOut & \textbf{0.468 $\pm$ 0.048} & \textbf{0.482 $\pm$ 0.047} & \textbf{0.488 $\pm$ 0.040} & \textbf{0.658 $\pm$ 0.035} \\
 &  & NoReadOut & 0.416 $\pm$ 0.041 & 0.424 $\pm$ 0.043 & 0.425 $\pm$ 0.048 & 0.604 $\pm$ 0.050 \\
\midrule
 & GCN & OmicsReadOut & \textbf{0.478 $\pm$ 0.039} & \textbf{0.499 $\pm$ 0.038} & \textbf{0.508 $\pm$ 0.036} & \textbf{0.683 $\pm$ 0.042} \\
 &  & NoReadOut & 0.385 $\pm$ 0.042 & 0.390 $\pm$ 0.045 & 0.402 $\pm$ 0.040 & 0.608 $\pm$ 0.026 \\
\midrule
 & GATv2 & OmicsReadOut & \textbf{0.474 $\pm$ 0.035} & \textbf{0.492 $\pm$ 0.031} & \textbf{0.502 $\pm$ 0.031} & \textbf{0.680 $\pm$ 0.017} \\
 &  & NoReadOut & \withinstd 0.455 $\pm$ 0.046 & \withinstd 0.464 $\pm$ 0.045 & 0.468 $\pm$ 0.046 & 0.647 $\pm$ 0.043 \\
\midrule
 & SAGE & OmicsReadOut & \textbf{0.460 $\pm$ 0.035} & \textbf{0.484 $\pm$ 0.034} & \textbf{0.511 $\pm$ 0.022} & \textbf{0.687 $\pm$ 0.013} \\
 &  & NoReadOut & \withinstd 0.437 $\pm$ 0.026 & 0.443 $\pm$ 0.021 & 0.446 $\pm$ 0.025 & 0.644 $\pm$ 0.023 \\
\midrule
 & GPS & OmicsReadOut & \textbf{0.522 $\pm$ 0.063} & \textbf{0.533 $\pm$ 0.064} & \textbf{0.533 $\pm$ 0.066} & \textbf{0.709 $\pm$ 0.035} \\
 &  & NoReadOut & 0.404 $\pm$ 0.027 & 0.416 $\pm$ 0.031 & 0.444 $\pm$ 0.031 & 0.638 $\pm$ 0.030 \\
\midrule
 & SAGN & OmicsReadOut & \textbf{0.510 $\pm$ 0.035} & \textbf{0.525 $\pm$ 0.035} & \textbf{0.530 $\pm$ 0.032} & \textbf{0.684 $\pm$ 0.042} \\
 &  & NoReadOut & 0.406 $\pm$ 0.044 & 0.412 $\pm$ 0.046 & 0.413 $\pm$ 0.052 & 0.641 $\pm$ 0.027 \\
\midrule
 & ChebNet & OmicsReadOut & \textbf{0.493 $\pm$ 0.050} & \textbf{0.509 $\pm$ 0.046} & \textbf{0.511 $\pm$ 0.042} & \textbf{0.687 $\pm$ 0.051} \\
 &  & NoReadOut & 0.437 $\pm$ 0.042 & 0.445 $\pm$ 0.041 & 0.453 $\pm$ 0.044 & \withinstd 0.650 $\pm$ 0.040 \\
\midrule
 & MLA-GNN & OmicsReadOut & \withinstd 0.469 $\pm$ 0.047 & \withinstd 0.484 $\pm$ 0.040 & \textbf{0.492 $\pm$ 0.030} & 0.658 $\pm$ 0.038 \\
 &  & NoReadOut & \textbf{0.478 $\pm$ 0.041} & \textbf{0.488 $\pm$ 0.043} & \withinstd 0.491 $\pm$ 0.041 & \textbf{0.694 $\pm$ 0.034} \\
\bottomrule
\end{tabularx}
\label{tab:addneuromed-readout-vs-noreadout}
\end{table*}

\newpage

\begin{table*}[!h]
\centering
\small
\setlength{\tabcolsep}{5pt}
\renewcommand{\arraystretch}{1.15}
\caption{Readout ablation on \textit{Smoking}. For each (model, readout), the highest val $F_{macro}$ configuration is selected; test metrics (mean $\pm$ std) are reported.}
\begin{tabularx}{\textwidth}{>{\raggedright\arraybackslash}p{1cm} l l Y Y Y Y}
\toprule
Dataset & Model & Readout & $F_{macro}$ & $F_{weighted}$ & Accuracy & AUROC \\
\midrule
\multicolumn{7}{@{}l}{\textbf{Smoking}}\\[-0.6ex]
 & MLP & OmicsReadOut & --- & --- & --- & --- \\
 &  & NoReadOut & \textbf{0.636 $\pm$ 0.035} & \textbf{0.655 $\pm$ 0.031} & \textbf{0.655 $\pm$ 0.030} & \textbf{0.680 $\pm$ 0.034} \\
\midrule
 & GIN & OmicsReadOut & \textbf{0.638 $\pm$ 0.031} & \textbf{0.649 $\pm$ 0.028} & \textbf{0.644 $\pm$ 0.029} & \textbf{0.659 $\pm$ 0.032} \\
 &  & NoReadOut & 0.560 $\pm$ 0.067 & 0.579 $\pm$ 0.068 & 0.578 $\pm$ 0.072 & 0.574 $\pm$ 0.062 \\
\midrule
 & GCN & OmicsReadOut & \withinstd 0.586 $\pm$ 0.040 & \withinstd 0.614 $\pm$ 0.037 & \withinstd 0.623 $\pm$ 0.034 & \textbf{0.663 $\pm$ 0.070} \\
 &  & NoReadOut & \textbf{0.614 $\pm$ 0.040} & \textbf{0.632 $\pm$ 0.041} & \textbf{0.631 $\pm$ 0.043} & \withinstd 0.652 $\pm$ 0.052 \\
\midrule
 & GATv2 & OmicsReadOut & \withinstd 0.626 $\pm$ 0.055 & \textbf{0.650 $\pm$ 0.049} & \textbf{0.655 $\pm$ 0.045} & \textbf{0.701 $\pm$ 0.029} \\
 &  & NoReadOut & \textbf{0.633 $\pm$ 0.040} & \withinstd 0.649 $\pm$ 0.042 & \withinstd 0.651 $\pm$ 0.042 & 0.653 $\pm$ 0.029 \\
\midrule
 & SAGE & OmicsReadOut & \textbf{0.617 $\pm$ 0.044} & \textbf{0.639 $\pm$ 0.038} & \textbf{0.642 $\pm$ 0.033} & \textbf{0.719 $\pm$ 0.049} \\
 &  & NoReadOut & \withinstd 0.581 $\pm$ 0.063 & 0.601 $\pm$ 0.069 & 0.604 $\pm$ 0.074 & 0.539 $\pm$ 0.056 \\
\midrule
 & GPS & OmicsReadOut & \textbf{0.640 $\pm$ 0.038} & \textbf{0.660 $\pm$ 0.035} & \textbf{0.662 $\pm$ 0.035} & \textbf{0.679 $\pm$ 0.055} \\
 &  & NoReadOut & 0.594 $\pm$ 0.047 & 0.614 $\pm$ 0.044 & 0.616 $\pm$ 0.045 & \withinstd 0.649 $\pm$ 0.064 \\
\midrule
 & SAGN & OmicsReadOut & \textbf{0.656 $\pm$ 0.034} & \textbf{0.674 $\pm$ 0.032} & \textbf{0.674 $\pm$ 0.032} & \withinstd 0.709 $\pm$ 0.047 \\
 &  & NoReadOut & \withinstd 0.654 $\pm$ 0.064 & \withinstd 0.671 $\pm$ 0.059 & \withinstd 0.670 $\pm$ 0.058 & \textbf{0.713 $\pm$ 0.070} \\
\midrule
 & ChebNet & OmicsReadOut & \textbf{0.673 $\pm$ 0.035} & \textbf{0.691 $\pm$ 0.035} & \textbf{0.692 $\pm$ 0.037} & \textbf{0.740 $\pm$ 0.035} \\
 &  & NoReadOut & 0.604 $\pm$ 0.017 & 0.623 $\pm$ 0.015 & 0.625 $\pm$ 0.014 & 0.623 $\pm$ 0.033 \\
\midrule
 & MLA-GNN & OmicsReadOut & \textbf{0.613 $\pm$ 0.019} & \textbf{0.631 $\pm$ 0.017} & \withinstd 0.629 $\pm$ 0.018 & \withinstd 0.648 $\pm$ 0.033 \\
 &  & NoReadOut & \withinstd 0.610 $\pm$ 0.073 & \withinstd 0.629 $\pm$ 0.066 & \textbf{0.631 $\pm$ 0.064} & \textbf{0.667 $\pm$ 0.108} \\
\bottomrule
\end{tabularx}
\label{tab:smoking-readout-vs-noreadout}
\end{table*}

\newpage

\begin{table*}[!h]
\centering
\small
\setlength{\tabcolsep}{5pt}
\renewcommand{\arraystretch}{1.15}
\caption{Readout ablation on \textit{BRCA}. For each (model, readout), the highest val $F_{macro}$ configuration is selected; test metrics (mean $\pm$ std) are reported.}
\begin{tabularx}{\textwidth}{>{\raggedright\arraybackslash}p{2.4cm} l l Y Y Y Y}
\toprule
Dataset & Model & Readout & $F_{macro}$ & $F_{weighted}$ & Accuracy & AUROC \\
\midrule
\multicolumn{7}{@{}l}{\textbf{BRCA}}\\[-0.6ex]
 & MLP & OmicsReadOut & --- & --- & --- & --- \\
 &  & NoReadOut & \textbf{0.800 $\pm$ 0.031} & \textbf{0.816 $\pm$ 0.017} & \textbf{0.816 $\pm$ 0.018} & \textbf{0.931 $\pm$ 0.010} \\
\midrule
 & GIN & OmicsReadOut & 0.676 $\pm$ 0.091 & \withinstd 0.740 $\pm$ 0.081 & \textbf{0.752 $\pm$ 0.064} & \textbf{0.901 $\pm$ 0.016} \\
 &  & NoReadOut & \textbf{0.693 $\pm$ 0.013} & \textbf{0.744 $\pm$ 0.010} & \withinstd 0.741 $\pm$ 0.013 & \withinstd 0.896 $\pm$ 0.016 \\
\midrule
 & GCN & OmicsReadOut & \textbf{0.688 $\pm$ 0.034} & \textbf{0.729 $\pm$ 0.037} & \textbf{0.725 $\pm$ 0.046} & \textbf{0.898 $\pm$ 0.022} \\
 &  & NoReadOut & \withinstd 0.670 $\pm$ 0.055 & \withinstd 0.697 $\pm$ 0.028 & \withinstd 0.689 $\pm$ 0.023 & \withinstd 0.886 $\pm$ 0.007 \\
\midrule
 & GATv2 & OmicsReadOut & \textbf{0.716 $\pm$ 0.021} & \textbf{0.769 $\pm$ 0.025} & \textbf{0.770 $\pm$ 0.025} & \textbf{0.920 $\pm$ 0.011} \\
 &  & NoReadOut & 0.676 $\pm$ 0.010 & 0.714 $\pm$ 0.024 & 0.706 $\pm$ 0.029 & 0.891 $\pm$ 0.016 \\
\midrule
 & SAGE & OmicsReadOut & \textbf{0.773 $\pm$ 0.048} & \textbf{0.802 $\pm$ 0.028} & \textbf{0.803 $\pm$ 0.028} & \textbf{0.941 $\pm$ 0.016} \\
 &  & NoReadOut & 0.693 $\pm$ 0.019 & 0.733 $\pm$ 0.027 & 0.725 $\pm$ 0.029 & 0.892 $\pm$ 0.007 \\
\midrule
 & GPS & OmicsReadOut & \textbf{0.780 $\pm$ 0.039} & \textbf{0.812 $\pm$ 0.030} & \textbf{0.809 $\pm$ 0.031} & \withinstd 0.901 $\pm$ 0.017 \\
 &  & NoReadOut & \withinstd 0.743 $\pm$ 0.026 & 0.763 $\pm$ 0.022 & 0.759 $\pm$ 0.019 & \textbf{0.909 $\pm$ 0.022} \\
\midrule
 & SAGN & OmicsReadOut & \textbf{0.737 $\pm$ 0.044} & \textbf{0.768 $\pm$ 0.026} & \textbf{0.789 $\pm$ 0.020} & \textbf{0.925 $\pm$ 0.014} \\
 &  & NoReadOut & 0.624 $\pm$ 0.025 & 0.688 $\pm$ 0.019 & 0.684 $\pm$ 0.025 & 0.867 $\pm$ 0.013 \\
\midrule
 & ChebNet & OmicsReadOut & \textbf{0.801 $\pm$ 0.025} & \textbf{0.824 $\pm$ 0.008} & \textbf{0.825 $\pm$ 0.008} & \textbf{0.941 $\pm$ 0.015} \\
 &  & NoReadOut & 0.685 $\pm$ 0.039 & 0.726 $\pm$ 0.035 & 0.717 $\pm$ 0.041 & 0.888 $\pm$ 0.014 \\
\midrule
 & MLA-GNN & OmicsReadOut & 0.752 $\pm$ 0.064 & \withinstd 0.795 $\pm$ 0.034 & \withinstd 0.795 $\pm$ 0.036 & 0.913 $\pm$ 0.017 \\
 &  & NoReadOut & \textbf{0.795 $\pm$ 0.037} & \textbf{0.803 $\pm$ 0.025} & \textbf{0.797 $\pm$ 0.025} & \textbf{0.941 $\pm$ 0.013} \\
\bottomrule
\end{tabularx}
\label{tab:brca-readout-vs-noreadout}
\end{table*}

\newpage

\begin{table*}[!h]
\centering
\small
\setlength{\tabcolsep}{5pt}
\renewcommand{\arraystretch}{1.15}
\caption{Adjacency ablation on \textit{Heritage}. For each (model, adjacency), the highest val $F_{macro}$ configuration is selected; test metrics (mean $\pm$ std) are reported.}
\begin{tabularx}{\textwidth}{>{\raggedright\arraybackslash}p{1.5cm} l l Y Y Y Y}
\toprule
Dataset & Model & Edges & $F_{macro}$ & $F_{weighted}$ & Accuracy & AUROC \\
\midrule
\multicolumn{7}{@{}l}{\textbf{Heritage}}\\[-0.6ex]
 & MLP & PPI & \textbf{0.534 $\pm$ 0.064} & \textbf{0.539 $\pm$ 0.064} & \textbf{0.537 $\pm$ 0.064} & \textbf{0.566 $\pm$ 0.059} \\
 &  & Co-expression & \withinstd 0.512 $\pm$ 0.057 & \withinstd 0.517 $\pm$ 0.057 & \withinstd 0.515 $\pm$ 0.058 & \withinstd 0.547 $\pm$ 0.054 \\
\midrule
 & GIN & PPI & \textbf{0.525 $\pm$ 0.035} & \textbf{0.532 $\pm$ 0.034} & \withinstd 0.531 $\pm$ 0.034 & \withinstd 0.518 $\pm$ 0.033 \\
 &  & Co-expression & \withinstd 0.505 $\pm$ 0.056 & \withinstd 0.520 $\pm$ 0.054 & \textbf{0.535 $\pm$ 0.061} & \textbf{0.524 $\pm$ 0.065} \\
\midrule
 & GCN & PPI & 0.479 $\pm$ 0.027 & 0.491 $\pm$ 0.021 & 0.497 $\pm$ 0.016 & 0.512 $\pm$ 0.057 \\
 &  & Co-expression & \textbf{0.557 $\pm$ 0.036} & \textbf{0.566 $\pm$ 0.037} & \textbf{0.567 $\pm$ 0.038} & \textbf{0.580 $\pm$ 0.021} \\
\midrule
 & GATv2 & PPI & \withinstd 0.502 $\pm$ 0.047 & \withinstd 0.516 $\pm$ 0.049 & \withinstd 0.523 $\pm$ 0.052 & \textbf{0.536 $\pm$ 0.065} \\
 &  & Co-expression & \textbf{0.512 $\pm$ 0.024} & \textbf{0.521 $\pm$ 0.019} & \textbf{0.524 $\pm$ 0.015} & \withinstd 0.510 $\pm$ 0.038 \\
\midrule
 & SAGE & PPI & \textbf{0.527 $\pm$ 0.059} & \withinstd 0.535 $\pm$ 0.055 & \withinstd 0.543 $\pm$ 0.049 & \textbf{0.540 $\pm$ 0.063} \\
 &  & Co-expression & \withinstd 0.526 $\pm$ 0.030 & \textbf{0.538 $\pm$ 0.026} & \textbf{0.544 $\pm$ 0.022} & \withinstd 0.530 $\pm$ 0.040 \\
\midrule
 & GPS & PPI & \withinstd 0.514 $\pm$ 0.056 & \withinstd 0.526 $\pm$ 0.053 & \textbf{0.531 $\pm$ 0.050} & \withinstd 0.530 $\pm$ 0.058 \\
 &  & Co-expression & \textbf{0.524 $\pm$ 0.024} & \textbf{0.531 $\pm$ 0.023} & \withinstd 0.531 $\pm$ 0.023 & \textbf{0.548 $\pm$ 0.058} \\
\midrule
 & SAGN & PPI & \withinstd 0.507 $\pm$ 0.027 & 0.508 $\pm$ 0.025 & 0.509 $\pm$ 0.027 & \textbf{0.532 $\pm$ 0.051} \\
 &  & Co-expression & \textbf{0.527 $\pm$ 0.027} & \textbf{0.535 $\pm$ 0.024} & \textbf{0.535 $\pm$ 0.022} & \withinstd 0.525 $\pm$ 0.035 \\
\midrule
 & ChebNet & PPI & \withinstd 0.508 $\pm$ 0.047 & \withinstd 0.515 $\pm$ 0.041 & \withinstd 0.523 $\pm$ 0.034 & \textbf{0.541 $\pm$ 0.064} \\
 &  & Co-expression & \textbf{0.512 $\pm$ 0.031} & \textbf{0.521 $\pm$ 0.032} & \textbf{0.526 $\pm$ 0.036} & \withinstd 0.533 $\pm$ 0.047 \\
\midrule
 & MLA-GNN & PPI & \withinstd 0.519 $\pm$ 0.044 & \withinstd 0.531 $\pm$ 0.039 & \withinstd 0.535 $\pm$ 0.033 & \withinstd 0.531 $\pm$ 0.058 \\
 &  & Co-expression & \textbf{0.538 $\pm$ 0.060} & \textbf{0.551 $\pm$ 0.056} & \textbf{0.558 $\pm$ 0.051} & \textbf{0.540 $\pm$ 0.066} \\
\bottomrule
\end{tabularx}
\label{tab:motrpac-adjacency-ppi-vs-coexpression}
\end{table*}

\newpage

\begin{table*}[!h]
\centering
\small
\setlength{\tabcolsep}{5pt}
\renewcommand{\arraystretch}{1.15}
\caption{Adjacency ablation on \textit{Tuberculosis}. For each (model, adjacency), the highest val $F_{macro}$ configuration is selected; test metrics (mean $\pm$ std) are reported.}
\begin{tabularx}{\textwidth}{>{\raggedright\arraybackslash}p{1.5cm} l l Y Y Y Y}
\toprule
Dataset & Model & Edges & $F_{macro}$ & $F_{weighted}$ & Accuracy & AUROC \\
\midrule
\multicolumn{7}{@{}l}{\textbf{Tuberculosis}}\\[-0.6ex]
 & MLP & PPI & \withinstd 0.642 $\pm$ 0.137 & \withinstd 0.660 $\pm$ 0.150 & \withinstd 0.667 $\pm$ 0.137 & \withinstd 0.729 $\pm$ 0.119 \\
 &  & Co-expression & \textbf{0.652 $\pm$ 0.104} & \textbf{0.676 $\pm$ 0.111} & \textbf{0.684 $\pm$ 0.104} & \textbf{0.753 $\pm$ 0.082} \\
\midrule
 & GIN & PPI & \textbf{0.654 $\pm$ 0.054} & \textbf{0.677 $\pm$ 0.051} & \textbf{0.670 $\pm$ 0.053} & \textbf{0.731 $\pm$ 0.053} \\
 &  & Co-expression & 0.540 $\pm$ 0.068 & 0.568 $\pm$ 0.111 & 0.583 $\pm$ 0.090 & 0.602 $\pm$ 0.062 \\
\midrule
 & GCN & PPI & \withinstd 0.659 $\pm$ 0.059 & \withinstd 0.680 $\pm$ 0.069 & \withinstd 0.675 $\pm$ 0.064 & \textbf{0.753 $\pm$ 0.066} \\
 &  & Co-expression & \textbf{0.675 $\pm$ 0.030} & \textbf{0.705 $\pm$ 0.028} & \textbf{0.709 $\pm$ 0.040} & \withinstd 0.731 $\pm$ 0.048 \\
\midrule
 & GATv2 & PPI & \textbf{0.677 $\pm$ 0.065} & \textbf{0.702 $\pm$ 0.078} & \textbf{0.703 $\pm$ 0.078} & \textbf{0.741 $\pm$ 0.053} \\
 &  & Co-expression & 0.608 $\pm$ 0.098 & \withinstd 0.633 $\pm$ 0.092 & \withinstd 0.630 $\pm$ 0.096 & 0.667 $\pm$ 0.104 \\
\midrule
 & SAGE & PPI & \textbf{0.721 $\pm$ 0.046} & \textbf{0.748 $\pm$ 0.038} & \textbf{0.746 $\pm$ 0.037} & \textbf{0.767 $\pm$ 0.053} \\
 &  & Co-expression & \withinstd 0.698 $\pm$ 0.072 & \withinstd 0.724 $\pm$ 0.080 & \withinstd 0.724 $\pm$ 0.078 & \withinstd 0.756 $\pm$ 0.076 \\
\midrule
 & GPS & PPI & \withinstd 0.635 $\pm$ 0.064 & \withinstd 0.668 $\pm$ 0.071 & \withinstd 0.670 $\pm$ 0.073 & \withinstd 0.702 $\pm$ 0.055 \\
 &  & Co-expression & \textbf{0.642 $\pm$ 0.086} & \textbf{0.670 $\pm$ 0.106} & \textbf{0.680 $\pm$ 0.090} & \textbf{0.716 $\pm$ 0.058} \\
\midrule
 & SAGN & PPI & \textbf{0.695 $\pm$ 0.051} & \textbf{0.723 $\pm$ 0.046} & \textbf{0.724 $\pm$ 0.052} & \withinstd 0.734 $\pm$ 0.057 \\
 &  & Co-expression & \withinstd 0.684 $\pm$ 0.078 & \withinstd 0.709 $\pm$ 0.087 & \withinstd 0.710 $\pm$ 0.085 & \textbf{0.754 $\pm$ 0.044} \\
\midrule
 & ChebNet & PPI & \textbf{0.659 $\pm$ 0.080} & \textbf{0.679 $\pm$ 0.086} & \textbf{0.678 $\pm$ 0.086} & \textbf{0.738 $\pm$ 0.062} \\
 &  & Co-expression & \withinstd 0.639 $\pm$ 0.116 & \withinstd 0.659 $\pm$ 0.134 & \withinstd 0.663 $\pm$ 0.122 & \withinstd 0.694 $\pm$ 0.113 \\
\midrule
 & MLA-GNN & PPI & \textbf{0.671 $\pm$ 0.039} & \textbf{0.699 $\pm$ 0.027} & \textbf{0.696 $\pm$ 0.029} & \textbf{0.720 $\pm$ 0.019} \\
 &  & Co-expression & \withinstd 0.659 $\pm$ 0.060 & \withinstd 0.686 $\pm$ 0.064 & \withinstd 0.683 $\pm$ 0.068 & 0.697 $\pm$ 0.056 \\
\bottomrule
\end{tabularx}
\label{tab:tuberculosis-adjacency-ppi-vs-coexpression}
\end{table*}

\newpage

\begin{table*}[!h]
\centering
\small
\setlength{\tabcolsep}{5pt}
\renewcommand{\arraystretch}{1.15}
\caption{Adjacency ablation on \textit{Parkinson's}. For each (model, adjacency), the highest val $F_{macro}$ configuration is selected; test metrics (mean $\pm$ std) are reported.}
\begin{tabularx}{\textwidth}{>{\raggedright\arraybackslash}p{1.5cm} l l Y Y Y Y}
\toprule
Dataset & Model & Edges & $F_{macro}$ & $F_{weighted}$ & Accuracy & AUROC \\
\midrule
\multicolumn{7}{@{}l}{\textbf{Parkinson's}}\\[-0.6ex]
 & MLP & PPI & \textbf{0.553 $\pm$ 0.071} & \textbf{0.579 $\pm$ 0.071} & \textbf{0.587 $\pm$ 0.058} & \textbf{0.600 $\pm$ 0.075} \\
 &  & Co-expression & \textbf{0.553 $\pm$ 0.071} & \textbf{0.579 $\pm$ 0.071} & \textbf{0.587 $\pm$ 0.058} & \textbf{0.600 $\pm$ 0.075} \\
\midrule
 & GIN & PPI & \withinstd 0.500 $\pm$ 0.035 & \withinstd 0.527 $\pm$ 0.038 & \withinstd 0.528 $\pm$ 0.029 & \withinstd 0.504 $\pm$ 0.056 \\
 &  & Co-expression & \textbf{0.531 $\pm$ 0.062} & \textbf{0.559 $\pm$ 0.055} & \textbf{0.562 $\pm$ 0.051} & \textbf{0.536 $\pm$ 0.067} \\
\midrule
 & GCN & PPI & \textbf{0.526 $\pm$ 0.110} & \textbf{0.551 $\pm$ 0.108} & \textbf{0.575 $\pm$ 0.059} & \textbf{0.580 $\pm$ 0.083} \\
 &  & Co-expression & \withinstd 0.519 $\pm$ 0.064 & \withinstd 0.546 $\pm$ 0.068 & \withinstd 0.563 $\pm$ 0.054 & \withinstd 0.560 $\pm$ 0.065 \\
\midrule
 & GATv2 & PPI & \withinstd 0.538 $\pm$ 0.057 & \withinstd 0.561 $\pm$ 0.057 & \withinstd 0.586 $\pm$ 0.035 & \textbf{0.612 $\pm$ 0.121} \\
 &  & Co-expression & \textbf{0.579 $\pm$ 0.063} & \textbf{0.603 $\pm$ 0.046} & \textbf{0.606 $\pm$ 0.047} & \withinstd 0.606 $\pm$ 0.098 \\
\midrule
 & SAGE & PPI & \textbf{0.566 $\pm$ 0.090} & \textbf{0.594 $\pm$ 0.081} & \textbf{0.602 $\pm$ 0.063} & \textbf{0.596 $\pm$ 0.115} \\
 &  & Co-expression & \withinstd 0.557 $\pm$ 0.095 & \withinstd 0.585 $\pm$ 0.087 & \withinstd 0.598 $\pm$ 0.087 & \withinstd 0.571 $\pm$ 0.126 \\
\midrule
 & GPS & PPI & \withinstd 0.487 $\pm$ 0.073 & 0.508 $\pm$ 0.079 & 0.531 $\pm$ 0.064 & \withinstd 0.565 $\pm$ 0.067 \\
 &  & Co-expression & \textbf{0.528 $\pm$ 0.067} & \textbf{0.581 $\pm$ 0.038} & \textbf{0.623 $\pm$ 0.037} & \textbf{0.596 $\pm$ 0.046} \\
\midrule
 & SAGN & PPI & \textbf{0.550 $\pm$ 0.100} & \textbf{0.582 $\pm$ 0.084} & \textbf{0.599 $\pm$ 0.061} & \textbf{0.600 $\pm$ 0.107} \\
 &  & Co-expression & \withinstd 0.530 $\pm$ 0.087 & \withinstd 0.551 $\pm$ 0.081 & \withinstd 0.564 $\pm$ 0.060 & \withinstd 0.570 $\pm$ 0.113 \\
\midrule
 & ChebNet & PPI & \withinstd 0.543 $\pm$ 0.057 & \textbf{0.577 $\pm$ 0.030} & \withinstd 0.598 $\pm$ 0.021 & \withinstd 0.612 $\pm$ 0.069 \\
 &  & Co-expression & \textbf{0.546 $\pm$ 0.143} & \withinstd 0.577 $\pm$ 0.130 & \textbf{0.607 $\pm$ 0.089} & \textbf{0.619 $\pm$ 0.101} \\
\midrule
 & MLA-GNN & PPI & \withinstd 0.497 $\pm$ 0.107 & \withinstd 0.545 $\pm$ 0.105 & \textbf{0.589 $\pm$ 0.054} & \withinstd 0.550 $\pm$ 0.096 \\
 &  & Co-expression & \textbf{0.538 $\pm$ 0.088} & \textbf{0.563 $\pm$ 0.078} & \withinstd 0.583 $\pm$ 0.041 & \textbf{0.593 $\pm$ 0.102} \\
\bottomrule
\end{tabularx}
\label{tab:parkinsons-adjacency-ppi-vs-coexpression}
\end{table*}

\newpage

\begin{table*}[!h]
\centering
\small
\setlength{\tabcolsep}{5pt}
\renewcommand{\arraystretch}{1.15}
\caption{Adjacency ablation on \textit{AddNeuroMed}. For each (model, adjacency), the highest val $F_{macro}$ configuration is selected; test metrics (mean $\pm$ std) are reported.}
\begin{tabularx}{\textwidth}{>{\raggedright\arraybackslash}p{1.5cm} l l Y Y Y Y}
\toprule
Dataset & Model & Edges & $F_{macro}$ & $F_{weighted}$ & Accuracy & AUROC \\
\midrule
\multicolumn{7}{@{}l}{\textbf{AddNeuroMed}}\\[-0.6ex]
 & MLP & PPI & \textbf{0.516 $\pm$ 0.039} & \textbf{0.527 $\pm$ 0.042} & \textbf{0.532 $\pm$ 0.042} & \textbf{0.716 $\pm$ 0.025} \\
 &  & Co-expression & \textbf{0.516 $\pm$ 0.039} & \textbf{0.527 $\pm$ 0.042} & \textbf{0.532 $\pm$ 0.042} & \textbf{0.716 $\pm$ 0.025} \\
\midrule
 & GIN & PPI & \textbf{0.468 $\pm$ 0.048} & \textbf{0.482 $\pm$ 0.047} & \textbf{0.488 $\pm$ 0.040} & \textbf{0.658 $\pm$ 0.035} \\
 &  & Co-expression & 0.416 $\pm$ 0.041 & 0.424 $\pm$ 0.043 & 0.425 $\pm$ 0.048 & 0.604 $\pm$ 0.050 \\
\midrule
 & GCN & PPI & \withinstd 0.478 $\pm$ 0.039 & \withinstd 0.499 $\pm$ 0.038 & \withinstd 0.508 $\pm$ 0.036 & \withinstd 0.683 $\pm$ 0.042 \\
 &  & Co-expression & \textbf{0.497 $\pm$ 0.033} & \textbf{0.513 $\pm$ 0.034} & \textbf{0.530 $\pm$ 0.030} & \textbf{0.687 $\pm$ 0.033} \\
\midrule
 & GATv2 & PPI & \withinstd 0.472 $\pm$ 0.042 & \withinstd 0.483 $\pm$ 0.037 & \withinstd 0.496 $\pm$ 0.035 & \withinstd 0.676 $\pm$ 0.043 \\
 &  & Co-expression & \textbf{0.474 $\pm$ 0.035} & \textbf{0.492 $\pm$ 0.031} & \textbf{0.502 $\pm$ 0.031} & \textbf{0.680 $\pm$ 0.017} \\
\midrule
 & SAGE & PPI & \textbf{0.482 $\pm$ 0.060} & \textbf{0.499 $\pm$ 0.057} & \textbf{0.516 $\pm$ 0.049} & \textbf{0.693 $\pm$ 0.031} \\
 &  & Co-expression & \withinstd 0.460 $\pm$ 0.035 & \withinstd 0.484 $\pm$ 0.034 & \withinstd 0.511 $\pm$ 0.022 & \withinstd 0.687 $\pm$ 0.013 \\
\midrule
 & GPS & PPI & \withinstd 0.477 $\pm$ 0.015 & \withinstd 0.490 $\pm$ 0.016 & \withinstd 0.494 $\pm$ 0.024 & 0.671 $\pm$ 0.040 \\
 &  & Co-expression & \textbf{0.522 $\pm$ 0.063} & \textbf{0.533 $\pm$ 0.064} & \textbf{0.533 $\pm$ 0.066} & \textbf{0.709 $\pm$ 0.035} \\
\midrule
 & SAGN & PPI & \withinstd 0.501 $\pm$ 0.036 & \withinstd 0.518 $\pm$ 0.034 & \textbf{0.533 $\pm$ 0.032} & \withinstd 0.679 $\pm$ 0.037 \\
 &  & Co-expression & \textbf{0.510 $\pm$ 0.035} & \textbf{0.525 $\pm$ 0.035} & \withinstd 0.530 $\pm$ 0.032 & \textbf{0.684 $\pm$ 0.042} \\
\midrule
 & ChebNet & PPI & \withinstd 0.493 $\pm$ 0.050 & \withinstd 0.509 $\pm$ 0.046 & \withinstd 0.511 $\pm$ 0.042 & \withinstd 0.687 $\pm$ 0.051 \\
 &  & Co-expression & \textbf{0.508 $\pm$ 0.042} & \textbf{0.522 $\pm$ 0.047} & \textbf{0.527 $\pm$ 0.044} & \textbf{0.709 $\pm$ 0.022} \\
\midrule
 & MLA-GNN & PPI & 0.469 $\pm$ 0.047 & 0.484 $\pm$ 0.040 & 0.492 $\pm$ 0.030 & 0.658 $\pm$ 0.038 \\
 &  & Co-expression & \textbf{0.507 $\pm$ 0.019} & \textbf{0.518 $\pm$ 0.018} & \textbf{0.518 $\pm$ 0.018} & \textbf{0.686 $\pm$ 0.010} \\
\bottomrule
\end{tabularx}
\label{tab:addneuromed-adjacency-ppi-vs-coexpression}
\end{table*}

\newpage

\begin{table*}[!h]
\centering
\small
\setlength{\tabcolsep}{5pt}
\renewcommand{\arraystretch}{1.15}
\caption{Adjacency ablation on \textit{Smoking}. For each (model, adjacency), the highest val $F_{macro}$ configuration is selected; test metrics (mean $\pm$ std) are reported.}
\begin{tabularx}{\textwidth}{>{\raggedright\arraybackslash}p{1.5cm} l l Y Y Y Y}
\toprule
Dataset & Model & Edges & $F_{macro}$ & $F_{weighted}$ & Accuracy & AUROC \\
\midrule
\multicolumn{7}{@{}l}{\textbf{Smoking}}\\[-0.6ex]
 & MLP & PPI & \textbf{0.636 $\pm$ 0.035} & \textbf{0.655 $\pm$ 0.031} & \textbf{0.655 $\pm$ 0.030} & \textbf{0.680 $\pm$ 0.034} \\
 &  & Co-expression & \textbf{0.636 $\pm$ 0.035} & \textbf{0.655 $\pm$ 0.031} & \textbf{0.655 $\pm$ 0.030} & \textbf{0.680 $\pm$ 0.034} \\
\midrule
 & GIN & PPI & \textbf{0.638 $\pm$ 0.031} & \textbf{0.649 $\pm$ 0.028} & \textbf{0.644 $\pm$ 0.029} & \textbf{0.659 $\pm$ 0.032} \\
 &  & Co-expression & 0.560 $\pm$ 0.067 & 0.579 $\pm$ 0.068 & 0.578 $\pm$ 0.072 & 0.574 $\pm$ 0.062 \\
\midrule
 & GCN & PPI & \withinstd 0.586 $\pm$ 0.040 & \withinstd 0.614 $\pm$ 0.037 & \withinstd 0.623 $\pm$ 0.034 & \withinstd 0.663 $\pm$ 0.070 \\
 &  & Co-expression & \textbf{0.610 $\pm$ 0.041} & \textbf{0.633 $\pm$ 0.042} & \textbf{0.638 $\pm$ 0.045} & \textbf{0.682 $\pm$ 0.039} \\
\midrule
 & GATv2 & PPI & \withinstd 0.626 $\pm$ 0.055 & \textbf{0.650 $\pm$ 0.049} & \textbf{0.655 $\pm$ 0.045} & \textbf{0.701 $\pm$ 0.029} \\
 &  & Co-expression & \textbf{0.627 $\pm$ 0.018} & \withinstd 0.643 $\pm$ 0.019 & \withinstd 0.640 $\pm$ 0.020 & \withinstd 0.689 $\pm$ 0.042 \\
\midrule
 & SAGE & PPI & \withinstd 0.617 $\pm$ 0.044 & \withinstd 0.639 $\pm$ 0.038 & \withinstd 0.642 $\pm$ 0.033 & \textbf{0.719 $\pm$ 0.049} \\
 &  & Co-expression & \textbf{0.635 $\pm$ 0.041} & \textbf{0.653 $\pm$ 0.038} & \textbf{0.655 $\pm$ 0.041} & \withinstd 0.688 $\pm$ 0.047 \\
\midrule
 & GPS & PPI & \textbf{0.640 $\pm$ 0.038} & \textbf{0.660 $\pm$ 0.035} & \textbf{0.662 $\pm$ 0.035} & \withinstd 0.679 $\pm$ 0.055 \\
 &  & Co-expression & \withinstd 0.618 $\pm$ 0.060 & \withinstd 0.637 $\pm$ 0.056 & \withinstd 0.636 $\pm$ 0.056 & \textbf{0.709 $\pm$ 0.048} \\
\midrule
 & SAGN & PPI & \withinstd 0.644 $\pm$ 0.052 & \withinstd 0.664 $\pm$ 0.049 & \withinstd 0.668 $\pm$ 0.049 & \textbf{0.758 $\pm$ 0.033} \\
 &  & Co-expression & \textbf{0.656 $\pm$ 0.034} & \textbf{0.674 $\pm$ 0.032} & \textbf{0.674 $\pm$ 0.032} & 0.709 $\pm$ 0.047 \\
\midrule
 & ChebNet & PPI & 0.609 $\pm$ 0.033 & 0.632 $\pm$ 0.034 & 0.638 $\pm$ 0.039 & 0.691 $\pm$ 0.045 \\
 &  & Co-expression & \textbf{0.673 $\pm$ 0.035} & \textbf{0.691 $\pm$ 0.035} & \textbf{0.692 $\pm$ 0.037} & \textbf{0.740 $\pm$ 0.035} \\
\midrule
 & MLA-GNN & PPI & \withinstd 0.610 $\pm$ 0.073 & \withinstd 0.629 $\pm$ 0.066 & \withinstd 0.631 $\pm$ 0.064 & \textbf{0.667 $\pm$ 0.108} \\
 &  & Co-expression & \textbf{0.630 $\pm$ 0.053} & \textbf{0.650 $\pm$ 0.045} & \textbf{0.653 $\pm$ 0.040} & \withinstd 0.651 $\pm$ 0.060 \\
\bottomrule
\end{tabularx}
\label{tab:smoking-adjacency-ppi-vs-coexpression}
\end{table*}

\newpage

\begin{table*}[!h]
\centering
\small
\setlength{\tabcolsep}{5pt}
\renewcommand{\arraystretch}{1.15}
\caption{Adjacency ablation on \textit{BRCA}. For each (model, adjacency), the highest val $F_{macro}$ configuration is selected; test metrics (mean $\pm$ std) are reported.}
\begin{tabularx}{\textwidth}{>{\raggedright\arraybackslash}p{1.5cm} l l Y Y Y Y}
\toprule
Dataset & Model & Edges & $F_{macro}$ & $F_{weighted}$ & Accuracy & AUROC \\
\midrule
\multicolumn{7}{@{}l}{\textbf{BRCA}}\\[-0.6ex]
 & MLP & PPI & \textbf{0.800 $\pm$ 0.031} & \textbf{0.816 $\pm$ 0.017} & \textbf{0.816 $\pm$ 0.018} & \textbf{0.931 $\pm$ 0.010} \\
 &  & Co-expression & \textbf{0.800 $\pm$ 0.031} & \textbf{0.816 $\pm$ 0.017} & \textbf{0.816 $\pm$ 0.018} & \textbf{0.931 $\pm$ 0.010} \\
\midrule
 & GIN & PPI & 0.676 $\pm$ 0.091 & \withinstd 0.740 $\pm$ 0.081 & \textbf{0.752 $\pm$ 0.064} & \textbf{0.901 $\pm$ 0.016} \\
 &  & Co-expression & \textbf{0.693 $\pm$ 0.013} & \textbf{0.744 $\pm$ 0.010} & \withinstd 0.741 $\pm$ 0.013 & \withinstd 0.896 $\pm$ 0.016 \\
\midrule
 & GCN & PPI & \textbf{0.688 $\pm$ 0.034} & \withinstd 0.729 $\pm$ 0.037 & \withinstd 0.725 $\pm$ 0.046 & \textbf{0.898 $\pm$ 0.022} \\
 &  & Co-expression & \withinstd 0.671 $\pm$ 0.056 & \textbf{0.744 $\pm$ 0.024} & \textbf{0.753 $\pm$ 0.035} & \withinstd 0.895 $\pm$ 0.035 \\
\midrule
 & GATv2 & PPI & \withinstd 0.716 $\pm$ 0.021 & \withinstd 0.769 $\pm$ 0.025 & \textbf{0.770 $\pm$ 0.025} & \textbf{0.920 $\pm$ 0.011} \\
 &  & Co-expression & \textbf{0.726 $\pm$ 0.034} & \textbf{0.772 $\pm$ 0.027} & \textbf{0.770 $\pm$ 0.024} & 0.905 $\pm$ 0.017 \\
\midrule
 & SAGE & PPI & \textbf{0.773 $\pm$ 0.048} & \textbf{0.802 $\pm$ 0.028} & \textbf{0.803 $\pm$ 0.028} & \textbf{0.941 $\pm$ 0.016} \\
 &  & Co-expression & 0.724 $\pm$ 0.071 & 0.764 $\pm$ 0.043 & 0.772 $\pm$ 0.034 & 0.924 $\pm$ 0.024 \\
\midrule
 & GPS & PPI & \textbf{0.780 $\pm$ 0.039} & \textbf{0.812 $\pm$ 0.030} & \withinstd 0.809 $\pm$ 0.031 & 0.901 $\pm$ 0.017 \\
 &  & Co-expression & \withinstd 0.774 $\pm$ 0.028 & \withinstd 0.802 $\pm$ 0.022 & \textbf{0.812 $\pm$ 0.021} & \textbf{0.922 $\pm$ 0.017} \\
\midrule
 & SAGN & PPI & \textbf{0.749 $\pm$ 0.020} & \textbf{0.788 $\pm$ 0.010} & \textbf{0.797 $\pm$ 0.010} & \textbf{0.928 $\pm$ 0.012} \\
 &  & Co-expression & \withinstd 0.737 $\pm$ 0.044 & 0.768 $\pm$ 0.026 & \withinstd 0.789 $\pm$ 0.020 & \withinstd 0.925 $\pm$ 0.014 \\
\midrule
 & ChebNet & PPI & 0.759 $\pm$ 0.050 & 0.793 $\pm$ 0.026 & 0.795 $\pm$ 0.024 & \textbf{0.942 $\pm$ 0.015} \\
 &  & Co-expression & \textbf{0.801 $\pm$ 0.025} & \textbf{0.824 $\pm$ 0.008} & \textbf{0.825 $\pm$ 0.008} & \withinstd 0.941 $\pm$ 0.015 \\
\midrule
 & MLA-GNN & PPI & \withinstd 0.795 $\pm$ 0.037 & \textbf{0.803 $\pm$ 0.025} & \textbf{0.797 $\pm$ 0.025} & \withinstd 0.941 $\pm$ 0.013 \\
 &  & Co-expression & \textbf{0.795 $\pm$ 0.025} & \withinstd 0.799 $\pm$ 0.014 & \withinstd 0.792 $\pm$ 0.017 & \textbf{0.944 $\pm$ 0.007} \\
\bottomrule
\end{tabularx}
\label{tab:brca-adjacency-ppi-vs-coexpression}
\end{table*}

\newpage

\clearpage
\section{Ensemble-Based Model Selection for Validation Set Instability}
\label{app:ensemble_selection}

\begin{figure}[!h]
\centering
\includegraphics[width=\linewidth]{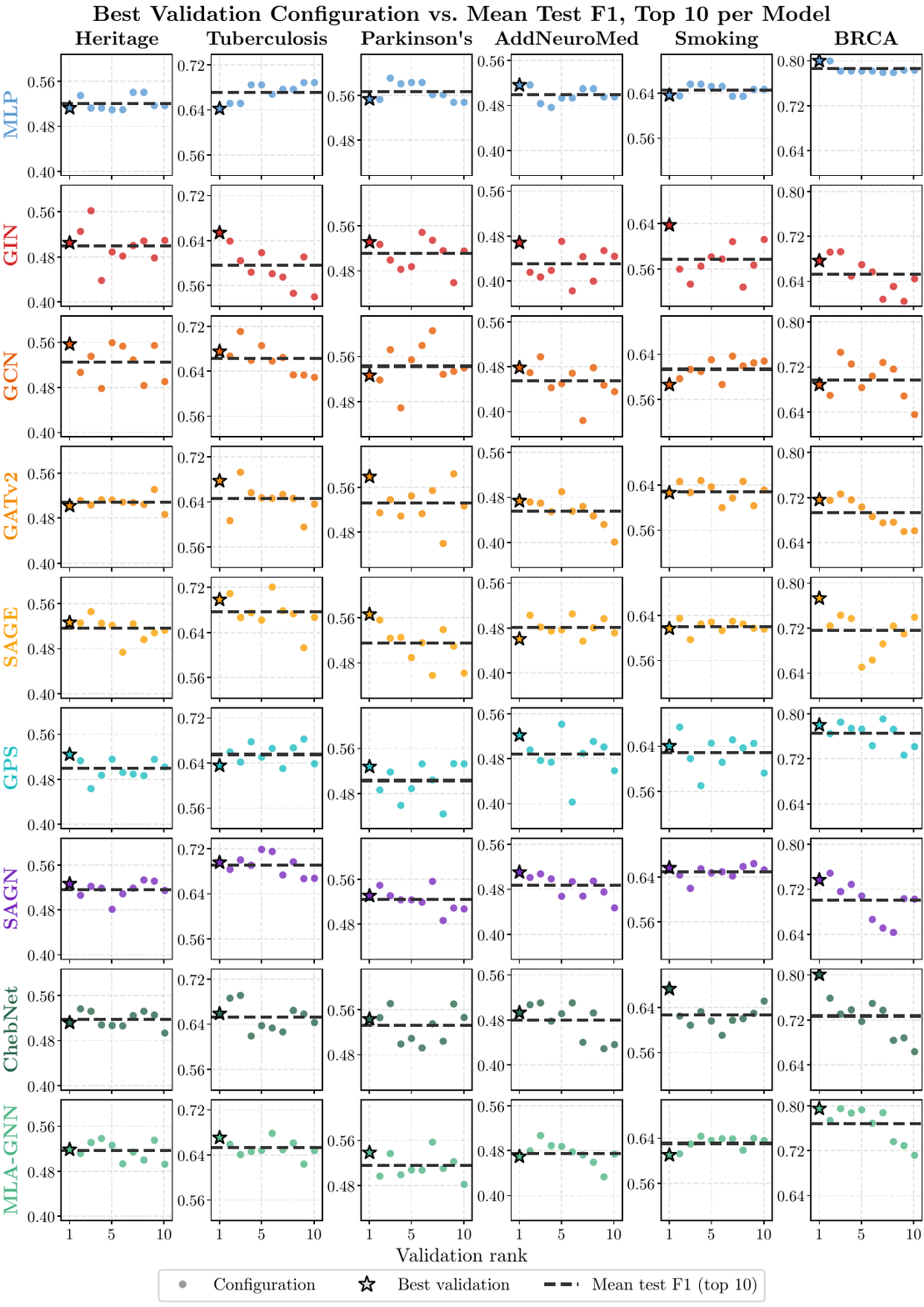}
\caption{Test $F_{\mathrm{macro}}$ of the ten highest-ranked configurations by validation $F_{\mathrm{macro}}$ for each model and dataset (star: rank 1; dashed line: mean of the top 10). Test performance varies more across top-ranked configurations for GNNs than for the MLP, so a single validation-selected configuration is a noisy estimate of model performance.}
\label{fig:val_rank_test}
\end{figure}

\subsection{Motivation}

Figure~\ref{fig:val_rank_test} shows test performance for the ten highest-ranked configurations by validation $F_{\mathrm{macro}}$. For the MLP, these configurations perform similarly on test, whereas for GNNs their test performance varies more widely, so the configuration ranked first on a small validation set may not reflect the typical performance of the model family. To test whether this affects our conclusions, we evaluate \textbf{top-$K$ ensemble aggregation}: instead of selecting the single best validation configuration, we average the predictions of the top-$K$ configurations (by mean validation F1 across folds) and report ensemble test performance. This approach mitigates overfitting to validation noise while preserving the inductive biases of well-performing model families.

\subsection{Method}
For each dataset and model (MLP and each GNN):
   \begin{enumerate}
       \item Rank all hyperparameter configurations by mean validation $F_{\mathrm{macro}}$ across the five folds.
       \item Select the top-$K$ configurations (we report $K=10$).
       \item For each fold, average the test-set class probabilities of the $K$ configurations' checkpoints for that fold and take the argmax (soft voting), then compute the ensemble's test $F_{\mathrm{macro}}$.
       \item Report the mean $\pm$ std of ensemble test $F_{\mathrm{macro}}$ across the five folds.
   \end{enumerate}

Note that folds remain independent: we ensemble within each fold's checkpoints and average performance across folds, preserving valid uncertainty quantification.

\subsection{Results}

\begin{figure}
    \centering
\includegraphics[width=\linewidth]{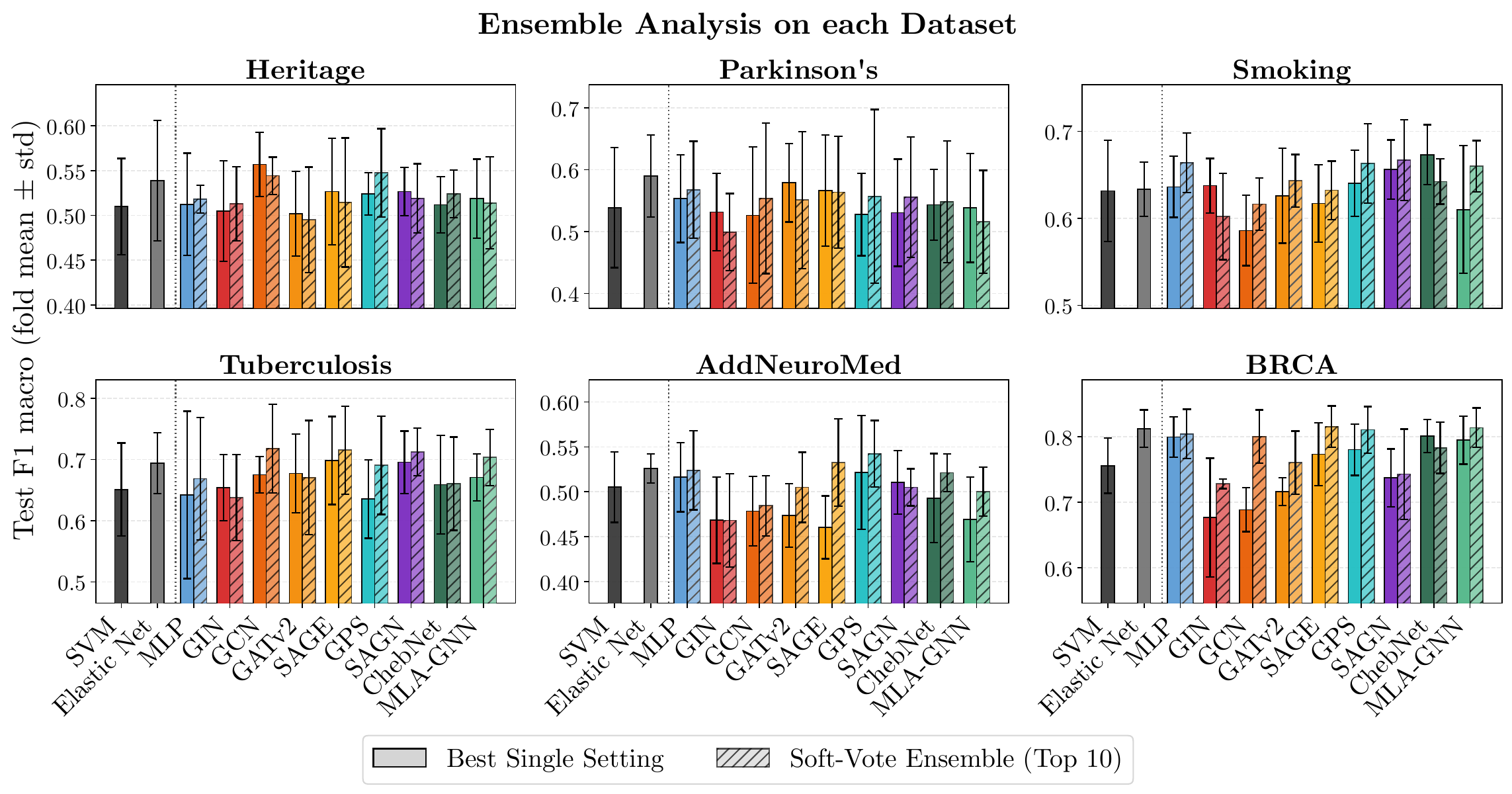}
    \caption{Test $F_{\mathrm{macro}}$ under single-best-validation selection (solid) and top-10 soft-vote ensembling (hatched). SVM and Elastic Net are shown without ensembling. Ensembling increases mean test $F_{\mathrm{macro}}$ for most models, although most gains fall within fold-to-fold variance.}
\label{fig:ensemble_comparison}
\end{figure}

Figure~\ref{fig:ensemble_comparison} compares top-10 soft-vote ensembles ($K=10$) with single-best-validation selection ($K=1$). Ensembling increases mean test $F_{\mathrm{macro}}$ for most GNN backbones, most consistently on Tuberculosis, AddNeuroMed, Smoking, and BRCA, and for the MLP on every dataset, although most gains fall within fold-to-fold variance. The largest gains occur where single-best selection performed poorly, such as GCN and GIN on BRCA. Under ensembling, a GNN achieves the highest mean $F_{\mathrm{macro}}$ on five of six datasets, compared with three of six under single-best selection, but always by margins of at most 0.03 over the best baseline, and SVM and Elastic Net were not ensembled.

\subsection{Implications for Practitioners}

\paragraph{Recommendation for $n \ll p$ settings.}
When validation sets are small relative to feature dimensionality, these results indicate that, though gains are marginal, \textbf{ensemble-based selection (K=10) could be preferred over single-best-validation selection}.

\paragraph{A Note on Computational Overhead.} We acknowledge that ensemble-based selection multiplies inference cost by K. Practitioners should weigh this against the reported gains.
\clearpage
\section{Detailed Node Selection Strategy vs. Ratio Ablation}
\label{sec:ratiovsablation}

\begin{figure}[!h]
    \centering
    \includegraphics[width=0.95\linewidth]{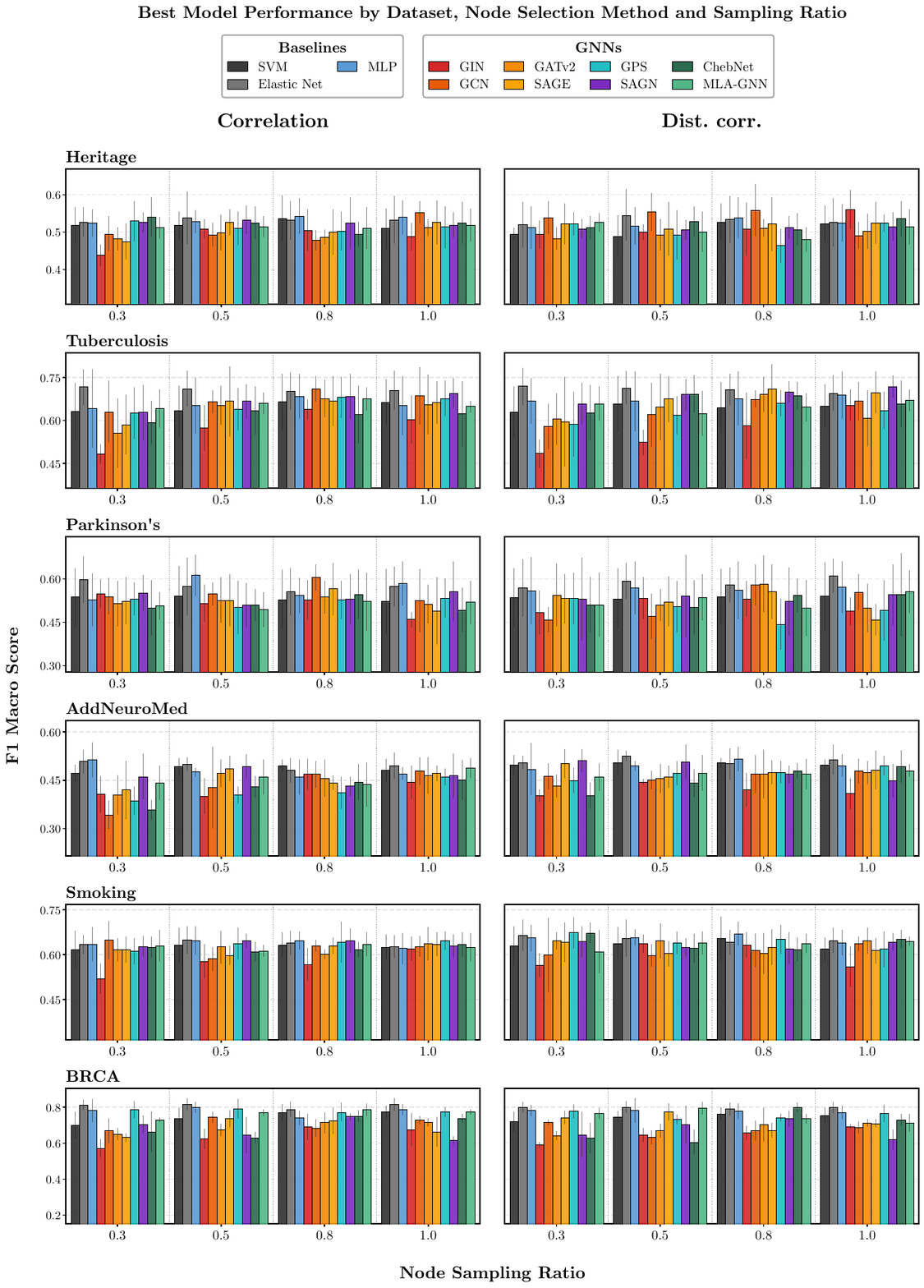}
    \caption{{\textbf{Test $F_{\mathrm{macro}}$ for all models under target-informed node selection.} Correlation-based (left) and distance-correlation-based (right) node selection across sampling ratios, for all baselines and GNN backbones. The sampling ratio is the number of samples relative to the number of selected nodes, so higher ratios retain fewer nodes. For each model, selection method, and ratio, the configuration with the highest mean validation $F_{\mathrm{macro}}$ is reported (mean $\pm$ std over five folds). Results for random and variance-based selection are shown in Figure~\ref{fig:full_pt2}.}}
    \label{fig:full_pt1}
\end{figure}

\begin{figure}
    \centering
    \includegraphics[width=\linewidth]{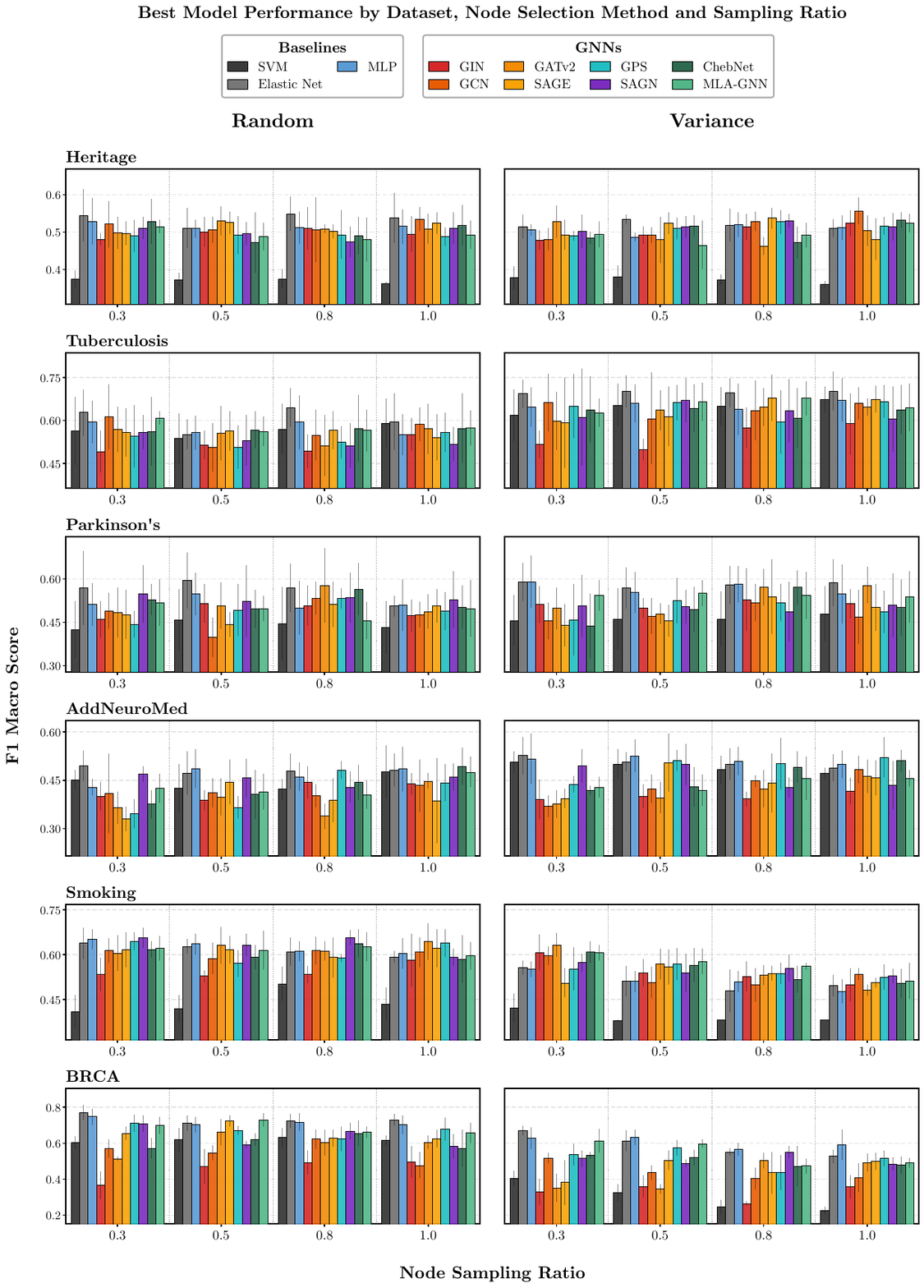}
    \caption{\textbf{Test $F_{\mathrm{macro}}$ for all models under unsupervised node selection.} Random (left) and variance-based (right) node selection across sampling ratios, for all baselines and GNN backbones. The sampling ratio is the number of samples relative to the number of selected nodes, so higher ratios retain fewer nodes. For each model, selection method, and ratio, the configuration with the highest mean validation $F_{\mathrm{macro}}$ is reported (mean $\pm$ std over five folds). Results for correlation and distance-correlation selection are shown in Figure~\ref{fig:full_pt1}.}
    \label{fig:full_pt2}
\end{figure}

\begin{figure}
    \centering
    \includegraphics[width=\linewidth]{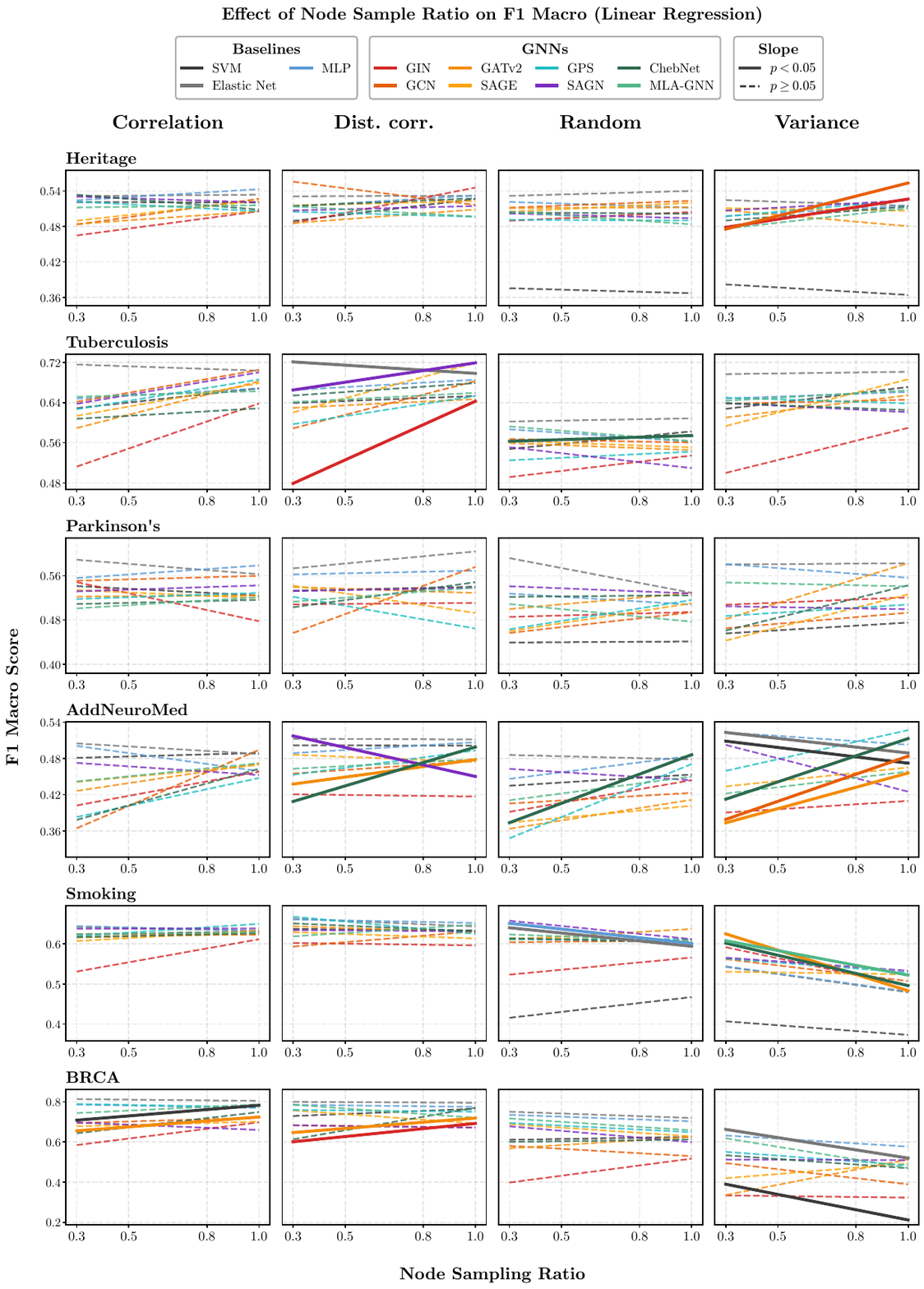}
    \caption{\textbf{Linear fit of test $F_{\mathrm{macro}}$ against sampling ratio for all models and selection methods.} Each line is a least-squares fit across sampling ratios to the mean test $F_{\mathrm{macro}}$ of the best validation configuration at each ratio; solid lines indicate slopes significant at $p<0.05$. The sampling ratio is the number of samples relative to the number of selected nodes, so higher ratios retain fewer nodes. Under target-informed selection, most GNN backbones on Tuberculosis and AddNeuroMed improve as fewer nodes are retained, whereas baselines are flat or decline. Under variance-based selection, most models decline on the epigenomic datasets.}
    \label{fig:ratio_regression}
\end{figure}

This appendix extends Figure~\ref{fig:ratio_method} to all models. Figures~\ref{fig:full_pt1} and~\ref{fig:full_pt2} show test $F_{\mathrm{macro}}$ for every baseline and GNN backbone under each node selection method and sampling ratio, and Figure~\ref{fig:ratio_regression} summarizes the effect of the sampling ratio with per-model linear fits. Together, they support the RQ3 findings that target-informed selection generally performs best and that GNNs and baselines respond differently to the number of retained nodes.

\clearpage
\section{Graph Structural Property Analysis}
\label{sec:struct_propt}

We test whether structural properties of the STRING PPI graphs explain when GNNs outperform the MLP. We restrict this analysis to STRING graphs because co-expression graphs are fixed in density (0.1) by construction, so their average degree is set by the number of nodes, and their edges are selected by the same feature correlations that homophily measures. Properties are computed on the graphs used for training (combined score $\geq 0.4$, self-loops excluded) and averaged over the five folds of each construction (Table~\ref{tab:graph_props_values}). Feature homophily is the mean cosine similarity, over edges, between the standardized expression profiles of the two endpoint nodes across training samples. Diameter is the length of the longest shortest path, counted in hops, within the graph's largest connected component. Modularity is Newman's modularity~\citep{doi:10.1073/pnas.0601602103} of the partition found by Louvain community detection~\citep{blondel2008fast}, both computed on the largest connected component.
We correlate each property (Spearman) with the gap $\Delta F_1 = F_{\text{macro}}^{\text{GNN}} - F_{\text{macro}}^{\text{MLP}}$ and with whether the GNN wins ($\Delta F_1 > 0$).

Across all $N=96$ constructions, only homophily correlates with the GNN--MLP gap ($\rho=+0.22$, $p=0.028$) and with GNN wins ($\rho=+0.25$, $p=0.013$; Table~\ref{tab:graph_props_pooled}). The effect is weak: it does not survive Bonferroni correction for the 12 tests, and although it is positive in five of six datasets, it is significant in none (Table~\ref{tab:graph_props_ds}). Modularity, diameter, clustering, average degree, and density show no pooled association ($|\rho|\le0.09$). Within datasets ($N=16$ each), correlations are mostly weak and disagree in sign. The significant ones are confined to Heritage (modularity, clustering, average degree) and Parkinson's (density). Graph structure therefore does not consistently predict whether a GNN improves over the MLP.

\begin{table}[!ht]
\centering
\small
\setlength{\tabcolsep}{4pt}
\caption{Structural properties of the STRING PPI graphs used for training (combined score $\geq 0.4$, self-loops excluded), averaged over the four node selection methods and the five folds. Avg.\ deg.: average degree. Dens.: density. Homoph.: feature homophily, the mean cosine similarity over edges of the endpoint nodes' standardized expression profiles across training samples. Mod.: modularity. Clust.: clustering coefficient. Diam.: diameter of the largest connected component, in hops. LCC: fraction of nodes in the largest connected component.}
\label{tab:graph_props_values}
\begin{tabular}{llrrrrrrrr}
\toprule
Dataset & $n_{\text{train}}/p$ & Nodes & Avg.\ deg. & Dens.\ (\%) & Homoph. & Mod. & Clust. & Diam. & LCC (\%) \\
\midrule
\multirow{4}{*}{Heritage} & 1.0 & 392 & 6.9 & 1.76 & 0.18 & 0.50 & 0.30 & 8.9 & 81.5 \\
 & 0.8 & 490 & 8.8 & 1.79 & 0.18 & 0.48 & 0.31 & 8.6 & 85.2 \\
 & 0.5 & 785 & 13.6 & 1.74 & 0.16 & 0.45 & 0.32 & 8.6 & 90.0 \\
 & 0.3 & 1{,}308 & 21.9 & 1.68 & 0.14 & 0.43 & 0.33 & 8.4 & 93.7 \\
\midrule
\multirow{4}{*}{Tuberculosis} & 1.0 & 337 & 5.1 & 1.51 & 0.75 & 0.57 & 0.26 & 10.2 & 80.3 \\
 & 0.8 & 420 & 6.3 & 1.50 & 0.76 & 0.55 & 0.27 & 10.3 & 84.5 \\
 & 0.5 & 673 & 10.1 & 1.51 & 0.78 & 0.52 & 0.29 & 8.9 & 91.4 \\
 & 0.3 & 1{,}122 & 16.1 & 1.43 & 0.81 & 0.49 & 0.29 & 8.7 & 95.6 \\
\midrule
\multirow{4}{*}{Parkinson's} & 1.0 & 321 & 2.6 & 0.81 & 0.31 & 0.66 & 0.25 & 10.6 & 49.4 \\
 & 0.8 & 401 & 3.0 & 0.76 & 0.31 & 0.66 & 0.24 & 10.4 & 54.8 \\
 & 0.5 & 642 & 4.4 & 0.68 & 0.30 & 0.62 & 0.25 & 10.6 & 62.1 \\
 & 0.3 & 1{,}070 & 6.8 & 0.64 & 0.28 & 0.57 & 0.26 & 10.8 & 70.2 \\
\midrule
\multirow{4}{*}{AddNeuroMed} & 1.0 & 427 & 8.0 & 1.87 & 0.30 & 0.56 & 0.31 & 10.4 & 65.2 \\
 & 0.8 & 533 & 8.7 & 1.64 & 0.28 & 0.55 & 0.31 & 9.4 & 68.3 \\
 & 0.5 & 853 & 11.3 & 1.33 & 0.23 & 0.53 & 0.30 & 9.2 & 74.7 \\
 & 0.3 & 1{,}422 & 15.5 & 1.09 & 0.18 & 0.51 & 0.29 & 9.2 & 80.8 \\
\midrule
\multirow{4}{*}{Smoking} & 1.0 & 278 & 1.2 & 0.43 & 0.18 & 0.61 & 0.21 & 10.8 & 29.5 \\
 & 0.8 & 347 & 1.5 & 0.43 & 0.19 & 0.63 & 0.21 & 12.0 & 37.7 \\
 & 0.5 & 557 & 2.4 & 0.44 & 0.20 & 0.61 & 0.23 & 12.8 & 51.3 \\
 & 0.3 & 928 & 3.7 & 0.40 & 0.16 & 0.59 & 0.23 & 12.4 & 61.5 \\
\midrule
\multirow{4}{*}{BRCA} & 1.0 & 384 & 1.8 & 0.46 & 0.27 & 0.69 & 0.18 & 14.0 & 48.7 \\
 & 0.8 & 480 & 2.2 & 0.46 & 0.26 & 0.67 & 0.19 & 14.1 & 57.1 \\
 & 0.5 & 768 & 3.8 & 0.49 & 0.26 & 0.61 & 0.20 & 12.6 & 70.4 \\
 & 0.3 & 1{,}280 & 6.4 & 0.50 & 0.27 & 0.56 & 0.23 & 12.2 & 81.0 \\
\bottomrule
\end{tabular}
\end{table}

\begin{table}[!ht]
\centering
\caption{Spearman correlations between STRING PPI graph properties and the GNN--MLP performance gap \textbf{across all datasets} (test $F_{\text{macro}}$; GNN = best GNN on validation). Each point is a graph construction (dataset $\times$ node sampling ratio $\times$ node selection method; $N=96$). $\Delta F_1 = F_{\text{macro}}^{\text{GNN}} - F_{\text{macro}}^{\text{MLP}}$; GNN wins indicates $\Delta F_1 > 0$. $^{*}$ indicates $p<0.05$. $p$-values are uncorrected. Under Bonferroni correction for the 12 tests in this table (threshold $p < 0.004$), no correlation is significant.}
\label{tab:graph_props_pooled}
\begin{tabular}{lrrrr}
\toprule
 & \multicolumn{2}{c}{$\rho$ vs.\ $\Delta F_1$} & \multicolumn{2}{c}{$\rho$ vs.\ GNN wins} \\
\cmidrule(lr){2-3} \cmidrule(lr){4-5}
Property & $\rho$ & $p$ & $\rho$ & $p$ \\
\midrule
Homophily & +0.22$^{*}$ & 0.028 & +0.25$^{*}$ & 0.013 \\
Modularity & $-$0.05 & 0.625 & $-$0.00 & 0.985 \\
Diameter & +0.05 & 0.638 & +0.03 & 0.742 \\
Clustering coefficient & $-$0.07 & 0.503 & $-$0.09 & 0.390 \\
Average degree & $-$0.00 & 0.979 & $-$0.04 & 0.681 \\
Density (\%) & $-$0.05 & 0.650 & $-$0.08 & 0.452 \\
\bottomrule
\end{tabular}
\end{table}

\begin{table}[!ht]
\centering
\small
\caption{Spearman correlations between STRING PPI graph properties and the GNN--MLP performance gap $\Delta F_1$ (test $F_{\text{macro}}$; GNN = best GNN on validation) \textbf{per dataset}, computed over each dataset's graph constructions (node sampling ratio $\times$ node selection method; $N=16$ per dataset). $^{*}$ indicates $p<0.05$.}
\label{tab:graph_props_ds}
\begin{tabular}{lrrrrrr}
\toprule
Property & Heritage & Tuberculosis & Parkinson's & AddNeuroMed & Smoking & BRCA \\
\midrule
Homophily & +0.05 & +0.27 & $-$0.46 & +0.29 & +0.30 & +0.31 \\
Modularity & $-$0.79$^{*}$ & +0.18 & +0.17 & $-$0.09 & +0.44 & $-$0.30 \\
Diameter & $-$0.49 & $-$0.10 & +0.21 & $-$0.01 & +0.03 & +0.26 \\
Clustering coefficient & +0.72$^{*}$ & $-$0.03 & $-$0.44 & +0.06 & +0.14 & $-$0.15 \\
Average degree & +0.70$^{*}$ & $-$0.11 & $-$0.26 & +0.13 & $-$0.09 & $-$0.04 \\
Density (\%) & +0.16 & +0.31 & $-$0.53$^{*}$ & +0.26 & $-$0.09 & $-$0.48 \\
\bottomrule
\end{tabular}
\end{table}
\clearpage

\newpage
\section*{NeurIPS Paper Checklist}

\begin{enumerate}

\item {\bf Claims}
    \item[] Question: Do the main claims made in the abstract and introduction accurately reflect the paper's contributions and scope?
    \item[] Answer: \answerYes{} % Replace by \answerYes{}, \answerNo{}, or \answerNA{}.
    \item[] Justification: In this paper, we provide the promised benchmark suite and framework (Section \ref{sec:benchmark-design}), as well as a reference benchmark result (Section \ref{sec:experiments}).

\item {\bf Limitations}
    \item[] Question: Does the paper discuss the limitations of the work performed by the authors?
    \item[] Answer: \answerYes{} % Replace by \answerYes{}, \answerNo{}, or \answerNA{}.
    \item[] Justification: Limitations provided in Section \ref{sec:Limitations}.

\item {\bf Theory assumptions and proofs}
    \item[] Question: For each theoretical result, does the paper provide the full set of assumptions and a complete (and correct) proof?
    \item[] Answer: \answerNA{} % Replace by \answerYes{}, \answerNo{}, or \answerNA{}.
    \item[] Justification: \answerNA{}

    \item {\bf Experimental result reproducibility}
    \item[] Question: Does the paper fully disclose all the information needed to reproduce the main experimental results of the paper to the extent that it affects the main claims and/or conclusions of the paper (regardless of whether the code and data are provided or not)?
    \item[] Answer: \answerYes{} % Replace by \answerYes{}, \answerNo{}, or \answerNA{}.
    \item[] Justification: All code is provided and experiments are thoroughly detailed in Appendix \ref{sec:setup_details}.

\item {\bf Open access to data and code}
    \item[] Question: Does the paper provide open access to the data and code, with sufficient instructions to faithfully reproduce the main experimental results, as described in supplemental material?
    \item[] Answer: \answerYes{} % Replace by \answerYes{}, \answerNo{}, or \answerNA{}.
    \item[] Justification: All code is provided.

\item {\bf Experimental setting/details}
    \item[] Question: Does the paper specify all the training and test details (e.g., data splits, hyperparameters, how they were chosen, type of optimizer) necessary to understand the results?
    \item[] Answer: \answerYes{} % Replace by \answerYes{}, \answerNo{}, or \answerNA{}.
    \item[] Justification: Data splits are released and training details are all in Appendix \ref{sec:setup_details}.

\item {\bf Experiment statistical significance}
    \item[] Question: Does the paper report error bars suitably and correctly defined or other appropriate information about the statistical significance of the experiments?
    \item[] Answer: \answerYes{} % Replace by \answerYes{}, \answerNo{}, or \answerNA{}.
    \item[] Justification: All performance results are reported as mean ± std over five cross-validation folds with a single training seed.

\item {\bf Experiments compute resources}
    \item[] Question: For each experiment, does the paper provide sufficient information on the computer resources (type of compute workers, memory, time of execution) needed to reproduce the experiments?
    \item[] Answer: \answerYes{} % Replace by \answerYes{}, \answerNo{}, or \answerNA{}.
    \item[] Justification: Computer resources are detailed in Appendix \ref{sec:setup_details}
    
\item {\bf Code of ethics}
    \item[] Question: Does the research conducted in the paper conform, in every respect, with the NeurIPS Code of Ethics \url{https://neurips.cc/public/EthicsGuidelines}?
    \item[] Answer: \answerYes{} % Replace by \answerYes{}, \answerNo{}, or \answerNA{}.
    \item[] Justification: This work involves no new human subjects research. All datasets used are publicly available, anonymized omics expression profiles collected under the original studies' institutional protocols. Code and data are released openly to support reproducibility. We do not foresee negative societal consequences beyond standard omics data privacy considerations.

\item {\bf Broader impacts}
    \item[] Question: Does the paper discuss both potential positive societal impacts and negative societal impacts of the work performed?
    \item[] Answer: \answerNA{}
    \item[] Justification: This work is foundational benchmarking research with no direct deployment path. All datasets are publicly available and anonymized; the benchmark itself introduces no new data collection, generative models, or surveillance capabilities, and poses no foreseeable risk of misuse.
    
\item {\bf Safeguards}
    \item[] Question: Does the paper describe safeguards that have been put in place for responsible release of data or models that have a high risk for misuse (e.g., pre-trained language models, image generators, or scraped datasets)?
    \item[] Answer: \answerNA{}
    \item[] Justification: This work releases a benchmarking framework and preprocessed omics datasets, not generative models or scraped data. The released artifacts pose no foreseeable risk of misuse.

\item {\bf Licenses for existing assets}
    \item[] Question: Are the creators or original owners of assets (e.g., code, data, models), used in the paper, properly credited and are the license and terms of use explicitly mentioned and properly respected?
    \item[] Answer: \answerYes{}
    \item[] Justification: All datasets are cited in the paper. 
    AddNeuroMed \citep{addneuromed}, Parkinson's 
    \citep{shamir2017parkinsons}, Tuberculosis (GSE19433) \citep{kunnath2010dynamic}, and Smoking (GSE50660) \citep{tsaprouni2014cigarette} are publicly available on NCBI GEO under GEO's public data access policy. The BRCA epigenomics data is sourced from MLOmics \citep{mlomics}, which is explicitly released under CC-BY 4.0 (available on Figshare and Hugging Face). The HERITAGE proteomics dataset \citep{robbins2021human} is available via the MoTrPAC Data Hub (\url{motrpac-data.org}) under CC-BY 4.0. All datasets are cited with their original sources. License details are documented in the released code repository.

\item {\bf New assets}
    \item[] Question: Are new assets introduced in the paper well documented and is the documentation provided alongside the assets?
    \item[] Answer: \answerYes{} % Replace by \answerYes{}, \answerNo{}, or \answerNA{}.
    \item[] Justification: All assets are detailed in Section \ref{sec:benchmark-design} of the paper.

\item {\bf Crowdsourcing and research with human subjects}
    \item[] Question: For crowdsourcing experiments and research with human subjects, does the paper include the full text of instructions given to participants and screenshots, if applicable, as well as details about compensation (if any)? 
    \item[] Answer: \answerNA{}
    \item[] Justification: This paper involves no crowdsourcing or direct 
    research with human subjects. All datasets are existing, publicly 
    available cohorts collected by their original authors.

\item {\bf Institutional review board (IRB) approvals or equivalent for research with human subjects}
    \item[] Question: Does the paper describe potential risks incurred by study participants, whether such risks were disclosed to the subjects, and whether Institutional Review Board (IRB) approvals (or an equivalent approval/review based on the requirements of your country or institution) were obtained?
    \item[] Answer: \answerNA{}
    \item[] Justification: This paper involves no new human subjects research. All datasets used are existing public cohorts governed by the original studies' institutional protocols.

\item {\bf Declaration of LLM usage}
    \item[] Question: Does the paper describe the usage of LLMs if it is an important, original, or non-standard component of the core methods in this research? Note that if the LLM is used only for writing, editing, or formatting purposes and does \emph{not} impact the core methodology, scientific rigor, or originality of the research, declaration is not required.
    %this research? 
    \item[] Answer:\answerNA{} % Replace by \answerYes{}, \answerNo{}, or \answerNA{}.
    \item[] Justification: LLM use does not impact the core methodology. 

\end{enumerate}

\end{document}